%% file: acl_latex.tex
\pdfoutput=1
\documentclass[11pt]{article}

\usepackage[final]{acl}

\usepackage{times}
\usepackage{latexsym}

\usepackage[T1]{fontenc}
\usepackage[utf8]{inputenc}

\usepackage{microtype}

\usepackage{inconsolata}

\usepackage{graphicx}

\usepackage{booktabs}
\usepackage{multirow}
\usepackage{subfigure}
\usepackage{amsmath,amssymb}
\newcommand{\MODELNAME}{\textsc{RECT}}

\title{Model Editing Harms General Abilities of Large Language Models: Regularization to the Rescue}

\author{
Jia-Chen Gu$^1$\thanks{\hspace{0.5mm}Equal contribution.}, Hao-Xiang Xu$^2$\footnotemark[1], Jun-Yu Ma$^2$, Pan Lu$^1$,   \\
{\bf Zhen-Hua Ling$^2$, Kai-Wei Chang$^1$, Nanyun Peng$^1$} \\
  $^1$University of California, Los Angeles \\
  $^2$University of Science and Technology of China\\
{\tt \{gujc,panbruin\}@ucla.edu}, {\tt \{nh2001620,mjy1999\}@mail.ustc.edu.cn}\\
{\tt {\tt zhling@ustc.edu.cn}, \{kwchang,violetpeng\}@cs.ucla.edu}
}

\begin{document}
\maketitle
\begin{abstract}
  Model editing is a technique that edits the large language models (LLMs) with updated knowledge to alleviate hallucinations without resource-intensive retraining. 
  While current model editing methods can effectively modify a model's behavior within a specific area of interest, they often overlook the potential unintended side effects on the general abilities of LLMs such as reasoning, natural language inference, and question answering. 
  In this paper, we raise concerns that model editing's improvements on factuality may come at the cost of a significant degradation of the model's general abilities.
  We systematically analyze the side effects by evaluating four popular editing methods on three LLMs across eight representative tasks.
  Our extensive empirical experiments show that it is challenging for current editing methods to simultaneously improve factuality of LLMs and maintain their general abilities. 
  Our analysis reveals that the side effects are caused by model editing altering the original model weights excessively, leading to \emph{overfitting} to the edited facts. 
  To mitigate this, a method named \MODELNAME{} is proposed to regularize the edit update weights by imposing constraints on their complexity based on the \textbf{RE}lative \textbf{C}hange in weigh\textbf{T}.
  Evaluation results show that \MODELNAME{} can significantly mitigate the side effects of editing while still maintaining over 94\% editing performance\footnote{\href{https://github.com/JasonForJoy/Model-Editing-Hurt}{https://github.com/JasonForJoy/Model-Editing-Hurt}}.
\end{abstract}

\input{1-introduction}

\input{2-related}

\input{3-preliminary}
\input{4-evaluation}
\input{5-regularization}
\input{6-conclusion}

\section*{Acknowledgement}
This research is based upon work supported by an Amazon AGI foundation research award, a google research scholar grant, CISCO sponsored research award, and NSF \#2331966.
We thank Tanmay Parekh, Po-Nien Kung, Sidi Lu, Fabrice Harel-Canada, UCLA NLP group members and anonymous reviewers for their valuable feedback.

\bibliography{custom}

\clearpage
\appendix
\onecolumn
\input{7-appendix}

\end{document}

%% file: 1-introduction.tex
\section{Introduction}

\begin{figure}[t]
\centering
\includegraphics[width=0.48\textwidth]{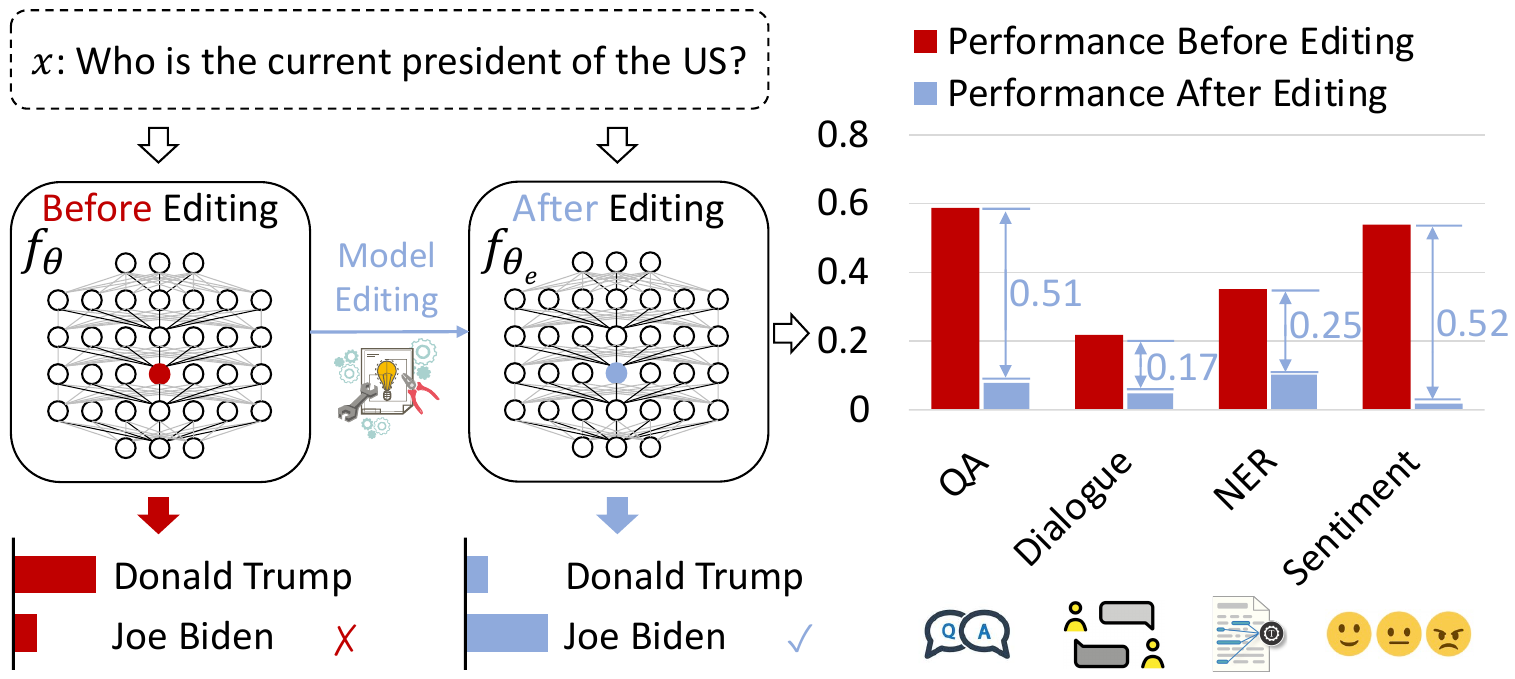}
\caption{
Demonstration of model editing and its impact on the general abilities of LLMs.
Although the factuality of the model has been improved, the general abilities of LLMs, such as question answering, dialogue, named entity recognition, sentiment analysis, have been substantially impaired after editing.
$f_{\theta}$ / $f_{\theta_{e}}$ denotes the models before / after editing.
} 
\label{fig-demo}
\end{figure}

  As real-world knowledge is dynamically increasing and updating, existing large language models (LLMs) need to constantly incorporate the inherit knowledge and up-to-date information for life-long learning. 
  Despite continual training, LLMs inevitably manifest hallucinations caused by missing, false or outdated knowledge embedded in their parameters~\cite{DBLP:journals/corr/abs-2309-01219, DBLP:journals/corr/abs-2302-12813, DBLP:journals/csur/JiLFYSXIBMF23}. 
  Due to the intensive computational cost of retraining LLMs, researchers have increasingly focused on \emph{model editing} (a.k.a., \emph{knowledge editing})~\cite{DBLP:conf/iclr/SinitsinPPPB20,DBLP:conf/emnlp/CaoAT21,DBLP:conf/acl/DaiDHSCW22,DBLP:conf/icml/MitchellLBMF22,DBLP:conf/nips/MengBAB22,DBLP:conf/iclr/MengSABB23,DBLP:conf/emnlp/YaoWT0LDC023,DBLP:conf/emnlp/ZhongWMPC23,DBLP:journals/corr/abs-2310-10322,DBLP:journals/corr/abs-2401-01286}. 
  This task is to efficiently modify a model's behavior within a specific area of interest through targeted interventions without resource-intensive model retraining. 
  
  At present, the assessment of editing methods typically involves evaluation along three critical dimensions~\cite{DBLP:conf/emnlp/YaoWT0LDC023}. 
  First, \emph{reliability} ensures the edited model can accurately recall the specific edited fact. 
  Second, \emph{generalization} validates the adaptability of the edited model by assessing the model's ability to recall the fact under diverse paraphrase prompts. 
  Finally, \emph{locality} checks if the edited model's output for unrelated inputs remains consistent after editing.
  These multi-faceted criteria collectively contribute to a nuanced understanding of the effectiveness and robustness of editing methods.

  In this paper, we put forward a critical concern regarding the overall robustness and adaptability of edited models.
  As shown in Figure~\ref{fig-demo}, while model editing methods have demonstrated improved factuality, it may come at the significant cost of the general abilities of LLMs such as summarization, question answering (QA), natural language inference.
  We argue that improving model factuality must be balanced with the need to maintain effectiveness across a range of abilities.

  In light of the above issues, we systematically study if model editing hurts the general abilities of LLMs. 
  This work studies model editing in the \emph{single}- versus \emph{sequential}-editing and \emph{instance}- versus \emph{batch}-editing settings.
  The edited models are evaluated on a variety of downstream tasks to see if there are any side effects on performance \emph{before} and \emph{after} editing.
  Extensive empirical experiments are conducted on \textbf{four popular editing methods}: 
  KN~\cite{DBLP:conf/acl/DaiDHSCW22}, 
  MEND~\cite{DBLP:conf/iclr/MitchellLBFM22},
  ROME~\cite{DBLP:conf/nips/MengBAB22}, and
  MEMIT~\cite{DBLP:conf/iclr/MengSABB23}
  applied to \textbf{three representative LLMs}: 
  GPT-2 XL (1.5B)~\cite{radford2019language}, 
  LLaMA-1 (7B)~\cite{DBLP:journals/corr/abs-2302-13971}, and 
  LLaMA-2 (7B)~\cite{DBLP:journals/corr/abs-2307-09288}.
  \textbf{Eight representative tasks} including 
  reasoning~\cite{DBLP:journals/corr/abs-2110-14168}, 
  natural language inference~\cite{DBLP:conf/mlcw/DaganGM05}, 
  open-domain QA~\cite{DBLP:journals/tacl/KwiatkowskiPRCP19}, 
  closed-domain QA~\cite{DBLP:conf/naacl/ClarkLCK0T19}, 
  dialogue~\cite{DBLP:conf/acl/CuiWLZZ20}, 
  summarization~\cite{gliwa-etal-2019-samsum}, 
  named entity recognition~\cite{DBLP:conf/conll/SangM03}, and 
  sentiment analysis~\cite{DBLP:conf/emnlp/SocherPWCMNP13}
  are employed to understand the impact of model editing on the general abilities of LLMs.
  
  Experimental results show that existing LLMs are not robust to weight perturbations, and editing even a few parameters can significantly affect their general abilities.
  Strikingly, with a single pass of editing involving less than 1\% parameters, LLaMA-1 (7B) exhibited a drastic performance degradation to nearly 0 on all the tasks we tried.
  These results demonstrate that current editing algorithms struggle to work effectively in tandem with LLMs to simultaneously improve model factuality and maintain general abilities.

  Furthermore, our analysis of the causes of side effects reveals that current model editing methods change the original model weights too much, resulting in \emph{overfitting} to new editing facts.
  The accumulation of overfitting across multiple edits can amplify the negative impact on the general abilities of LLMs.
  As a result, the edited model can recall new editing facts well but fails to generalize to various downstream tasks. 
  To this end, we design a regularization method named \MODELNAME{} (\textbf{RE}lative \textbf{C}hange in weigh\textbf{T}) to prevent overfitting.
  Basically, this regularization discourages overly complex editing updates that are more likely to overfit.
  Specifically, the top-\emph{k}\% elements in an edit update weight that change the most according to relative change in weights are considered as the principal editing information and keep their original values.
  While for the remaining elements in this edit update weight, they are treated as minor contributions to editing and set to zero for regularization.
  Evaluation results show that the edited models regularized by \MODELNAME{} can effectively mitigate the side effects of editing while still maintaining over 94\% editing performance 

  In summary, we demonstrate that although model editing is effective in updating parametric knowledge in a resource-efficient and target-specific way, current methods still have significant flaws in preserving the general abilities of LLMs.
  Existing research on model editing excessively pursued altering a model's behavior under specific knowledge while overlooked the premise of not compromising general abilities. 
  This paper points out the urgent shortcomings in model editing and proposes a regularization method to prevent overfitting across multiple edits to rescue the general abilities, calling for follow-up research efforts on trustworthy and robust model editing.

%% file: 2-related.tex
\section{Related Work}
Many studies have investigated model editing, including memory-based, meta-learning, and locate-then-edit \cite{DBLP:journals/corr/abs-2308-07269,DBLP:conf/emnlp/YaoWT0LDC023}.
Memory-based methods do not modify model weights but store the editing facts with an external memory~\cite{DBLP:conf/icml/MitchellLBMF22,DBLP:conf/emnlp/ZhongWMPC23}.
\citet{DBLP:conf/icml/MitchellLBMF22} stored edits in a base model and learned to reason over them to adjust its predictions as needed.
The latter two classes of methods are developed to directly modify the internal parameters of models, which is the focus of this paper.
On the one hand, meta-learning methods train a hypernetwork to get gradient changes to update model parameters~\cite{DBLP:conf/emnlp/CaoAT21,DBLP:conf/iclr/MitchellLBFM22}.
\citet{DBLP:conf/emnlp/CaoAT21} utilized a hypernetwork to predict parameter shift at test time.
\citet{DBLP:conf/iclr/MitchellLBFM22} learned to transform the fine-tuning gradient into a low-rank decomposition of the gradient.
On the other hand, locate-then-edit methods first locate knowledge neurons in LLMs that exhibit a positive correlation with a knowledge expression, and then modify them accordingly~\cite{DBLP:conf/acl/DaiDHSCW22,DBLP:conf/nips/MengBAB22,DBLP:conf/iclr/MengSABB23}.
In particular, \citet{DBLP:conf/acl/DaiDHSCW22} computed the contribution of each neurons to a certain knowledge, then updated or erased knowledge by modifying these neurons with the embedding vectors of facts.
\citet{DBLP:conf/nips/MengBAB22} located multi-layer perceptron (MLP) storing factual knowledge, and then edited such knowledge by injecting new key-value pair in the MLP module.
Besides, some works investigate the evaluation paradigm for model editing~\cite{DBLP:conf/emnlp/ZhongWMPC23, DBLP:journals/tacl/CohenBYGG24, DBLP:journals/corr/abs-2310-10322, DBLP:conf/iclr/Li0YW0C24, DBLP:conf/nips/HaseBKG23, DBLP:journals/corr/abs-2308-09954, DBLP:conf/iccv/GandikotaMFB23, DBLP:journals/corr/abs-2401-10471}.
For example, \citet{DBLP:journals/tacl/CohenBYGG24} introduced the ripple effects of editing, suggesting that editing a particular fact implies that many other facts need to be updated.
Additionally, recent works have also applied editing in various domains, such as changing model personality~\cite{DBLP:journals/corr/abs-2310-02168}, editing multimodal models~\cite{DBLP:conf/emnlp/0008TL0WC023}, protecting users privacy~\cite{DBLP:conf/emnlp/WuLXDW0X23}, etc.

A main difference between this work and previous related studies should be highlighted. 
These approaches target at designing editing algorithms to improve or evaluation paradigms to assess the editing performance.
In contrast, this study rethinks model editing and explores if current editing methods inadvertently cause the potential side effects on the underlying general abilities of LLMs.
The contemporaneous work \cite{DBLP:conf/acl/GuptaRA24} presents a similar finding that model editing at scale leads to catastrophic forgetting, but no mitigation method is proposed.
To the best of our knowledge, this paper makes the first call for attention to side effects on a variety of tasks beyond editing performance by presenting a systematical evaluation of four editing methods on three LLMs covering eight tasks. 
Besides, we also analyze the causes of side effects, and propose a regularization method to prevent editing overfitting.

%% file: 3-preliminary.tex
\section{Preliminary}
  Model editing involves modifying the memorized facts contained in LMs without retraining to better suit specific tasks or requirements.
  Various kinds of complex learned beliefs such as logical, spatial, or numerical knowledge are expected to be edited. 
  In this paper, we study editing factual knowledge in the form of (subject \emph{s}, relation \emph{r}, object \emph{o}), e.g., (\emph{s = United States, r = President of, o = Donald Trump}). 
  An LM is expected to recall a memory and predict the next token(s) representing $o$ given a natural language prompt $p(s, r)$ such as “\emph{The President of the United States is}”. 
  Editing a fact is to insert a new knowledge triple $(s, r, o^{*})$ in place of the current one $(s, r, o)$, where these two triples share the same subject and relation. 
  An editing operation is represented as $e = (s ,r, o, o^{*})$ for brevity.
  Given a set of editing facts $\mathcal{E}=\left\{e_{1}, e_{2}, \ldots\right\}$ and a model $f$, model editing involves learning a function $K$ that yields an edited LM  $f^{*}: K(f, \mathcal{E})=f^{*} $. 
  
  To evaluate the effectiveness of editing methods, previous works focus on evaluation along three dimensions~\cite{DBLP:conf/emnlp/CaoAT21,DBLP:conf/iclr/MitchellLBFM22,DBLP:conf/nips/MengBAB22,DBLP:conf/iclr/MengSABB23}. 
  First and foremost is \emph{reliability}, aiming to ascertain the ability of the edited model to accurately recall the specific editing facts. 
  The second dimension \emph{generalization} seeks to validate the adaptability of the edited model by assessing its ability to recall the editing facts under diverse paraphrase prompts. 
  The last dimension \emph{locality} (a.k.a., \emph{specificity}) is employed to verify the stability of the edited model by examining whether its output for unrelated inputs remains consistent after editing. 
  Due to limited space, readers can refer to Appendix~\ref{sec-appendix-metrics} for more detailed explanations and examples of the evaluation metrics.
  

%% file: 4-evaluation.tex
\section{Analysis of Side Effects of Editing}

\subsection{Evaluation Paradigm}

\begin{figure}[t]
\centering
\includegraphics[width=0.47\textwidth]{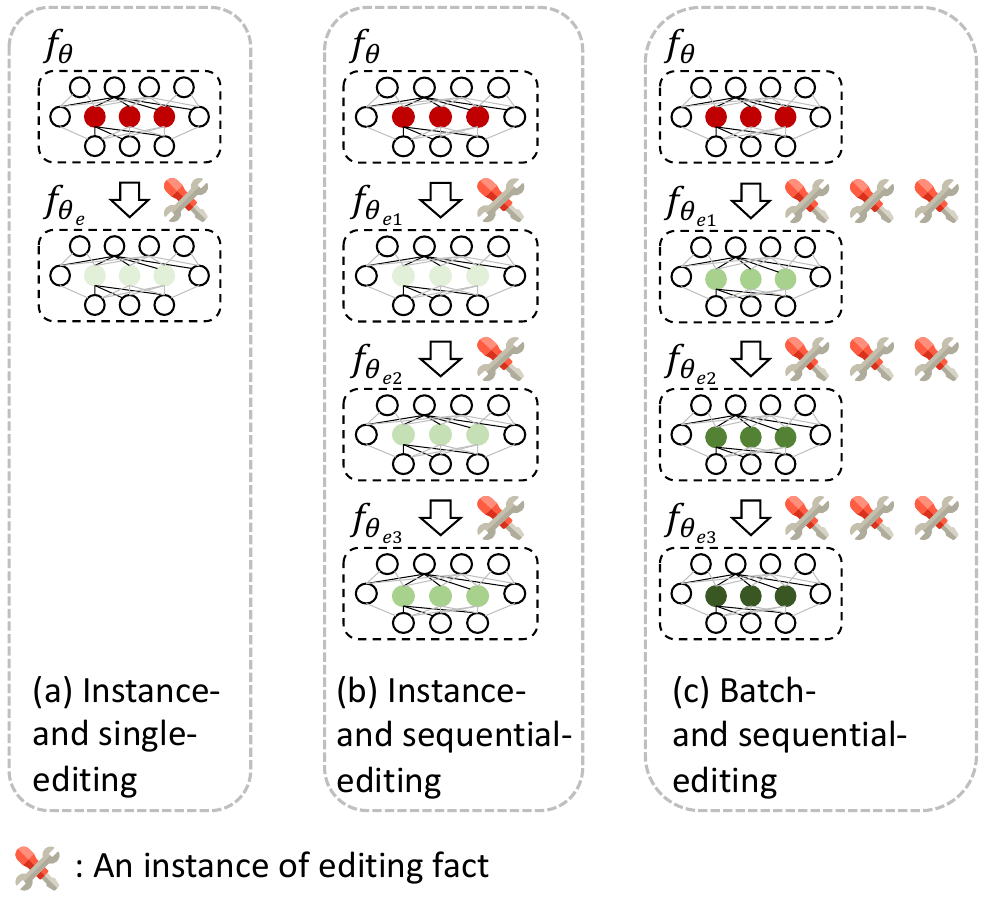}
\vspace{-1mm}
\caption{Illustration of the settings of (a) \emph{single- and instance-editing}, (b) \emph{sequential- and instance-editing}, and (c) \emph{sequential- and batch-editing}.
The darker units correspond to more edits.} 
\vspace{-2mm}
\label{fig-method}
\end{figure}

  This paper systematically studies the side effects of model editing in the \emph{single}- versus \emph{sequential}-editing and \emph{instance}- versus \emph{batch}-editing settings.
  Figure~\ref{fig-method} illustrates these experimental settings.
  The edited models are evaluated under the zero-shot setting on a variety of downstream tasks unrelated to editing facts to understand the performance \emph{before} and \emph{after} editing.

  \paragraph{Single- vs. Sequential-editing} \label{sec-single-seq}
  Single-editing involves examining the reliability and impact of making a single editing operation to a model.
  Specifically, it focuses on understanding how a model adapts to such a single alteration, and the implicit effect of such specific modifications on the overall performance. 
  It is worth noting that a single editing operation can contain either only one editing instance or multiple ones in a batch, which is further discussed later.
  In practice, there are often situations where only a particular change is needed, so it's crucial to understand how effectively the model integrates and preserves that individual edit.
  Therefore, evaluating the robustness to a single edit is crucial in determining its ability to retain the intended changes and overall performance.  
  
  In contrast to single-editing, multiple editing operations are conducted successively in sequential-editing~\cite{DBLP:conf/iclr/HuangSZZR023}.
  Similarly, each editing operation in sequential-editing can also contain either only one editing instance or multiple ones in a batch.
  Ideally, models should retain the changes from previous edits when carrying out a new one~\cite{DBLP:conf/emnlp/YaoWT0LDC023}, which is decisive for the continual learning of future LLMs. 
  Therefore, whether edited models can still maintain its general abilities after sequential editing is one of the important characteristics that should be considered. 
  For this analysis, how the performance of edited models on a variety of tasks changes as the number of edits increases will be explored.

  \paragraph{Instance- vs. Batch-editing}  \label{sec-inst-batch}
  Instance-editing refers to using only one instance per editing operation to make specific and targeted adjustments to individual pieces of knowledge within LLMs, regardless of the single- or sequential-editing settings.
  This setting is particularly valuable in situations where certain instances present unique challenges or outliers that require specialized treatment. 
  These fine-grained alterations to model behaviors over individual instances are expected to contribute to more adaptable and accurate LLMs.

  The real world is ever-changing, so there is a huge amount of knowledge that needs to be dynamically added and updated into LLMs.
  Despite the effectiveness of many instance-editing methods~\cite{DBLP:conf/acl/DaiDHSCW22,DBLP:conf/nips/MengBAB22,DBLP:journals/corr/abs-2310-10322}, ultimately at most a few dozens of pieces of knowledge can be updated~\cite{DBLP:conf/icml/MitchellLBMF22}, due to their relatively low but still non-negligible editing cost for a single instance.
  Since naive sequential applications of current state-of-the-art model editing methods fail to scale up~\cite{DBLP:conf/iclr/MengSABB23}, one may wish to update hundreds or thousands of facts simultaneously in batch-editing.
  Notably, batch-editing can also be coupled with both the single- or sequential-editing settings.

  \paragraph{Zero-shot Learning} \label{sec-zero-shot}
  Zero-shot learning aims to solve tasks without labeled training examples, and recent studies have demonstrated the superiority of LLMs for zero-shot learning~\cite{DBLP:conf/nips/BrownMRSKDNSSAA20,DBLP:conf/iclr/WeiBZGYLDDL22,DBLP:journals/jmlr/ChowdheryNDBMRBCSGSSTMRBTSPRDHPBAI23}. 
  Following these studies, we explore the zero-shot performance of unedited and edited models on a variety of tasks.
  Given a task instruction and a test problem that are concatenated as the input, the model is expected to generate a target text to address this problem. 
  The instructions and input formats of different tasks are shown in Appendix~\ref{sec-appendix-prompt}, which are taken from or inspired by~\citet{DBLP:conf/emnlp/QinZ0CYY23}.

  \subsection{Evaluation Setup}

  We briefly introduced the experimental setup regarding editing methods, editing datasets, selected LLMs, and representative tasks here.
  Readers can refer to their corresponding papers for more details.

%%%%%%%%%%%%%%%%%%%%%%%%%%%%%%%%%%%%%%%%%%%%%%%%%%%%%%%%%%%%%%%%%%%%%%%%%%%%

\begin{figure*}[t]
  \centering
  \subfigure{
  \includegraphics[width=3.8cm]{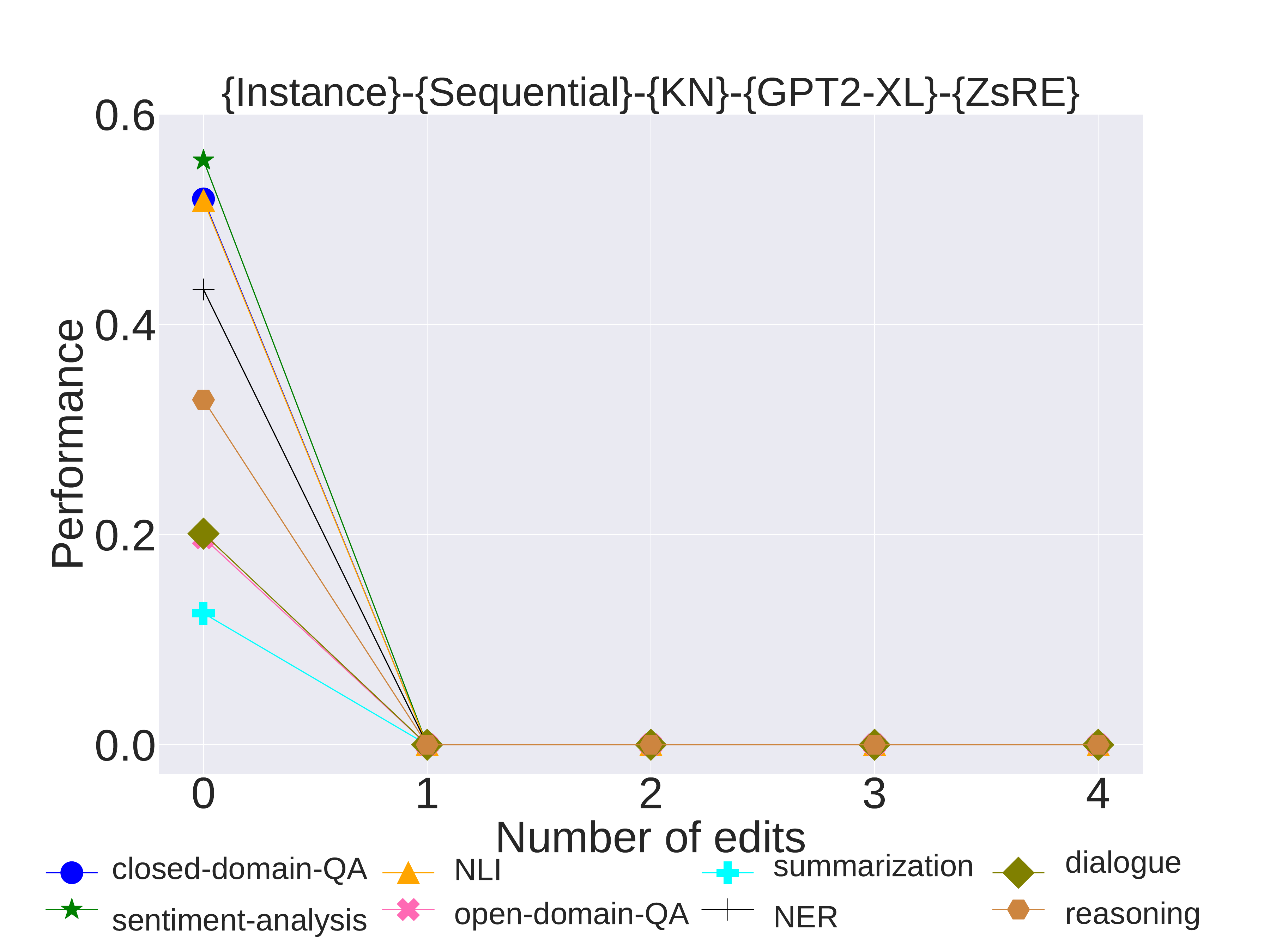}}
  \subfigure{
  \includegraphics[width=3.8cm]{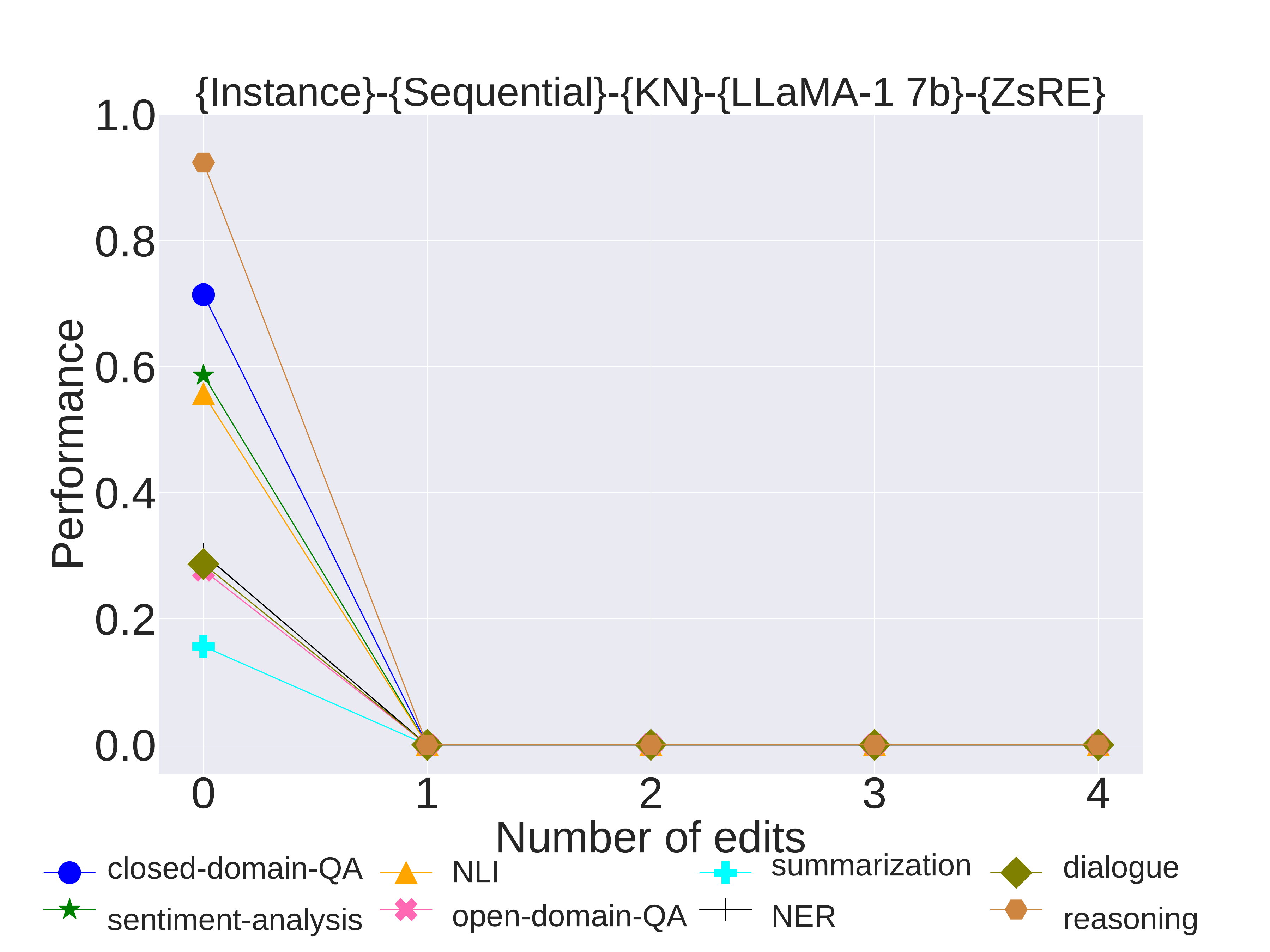}}
  \subfigure{
  \includegraphics[width=3.8cm]{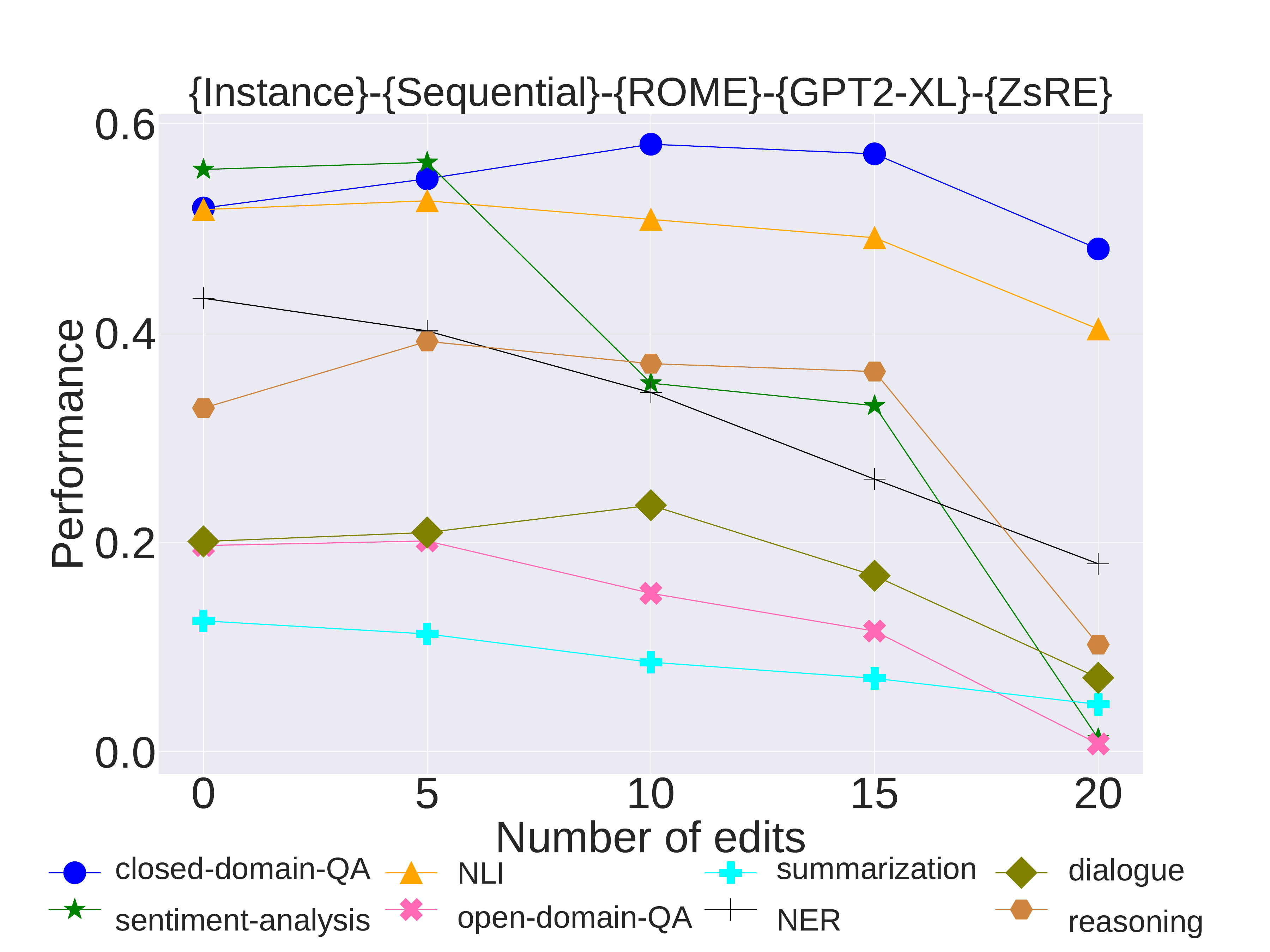}}
  \subfigure{
  \includegraphics[width=3.8cm]{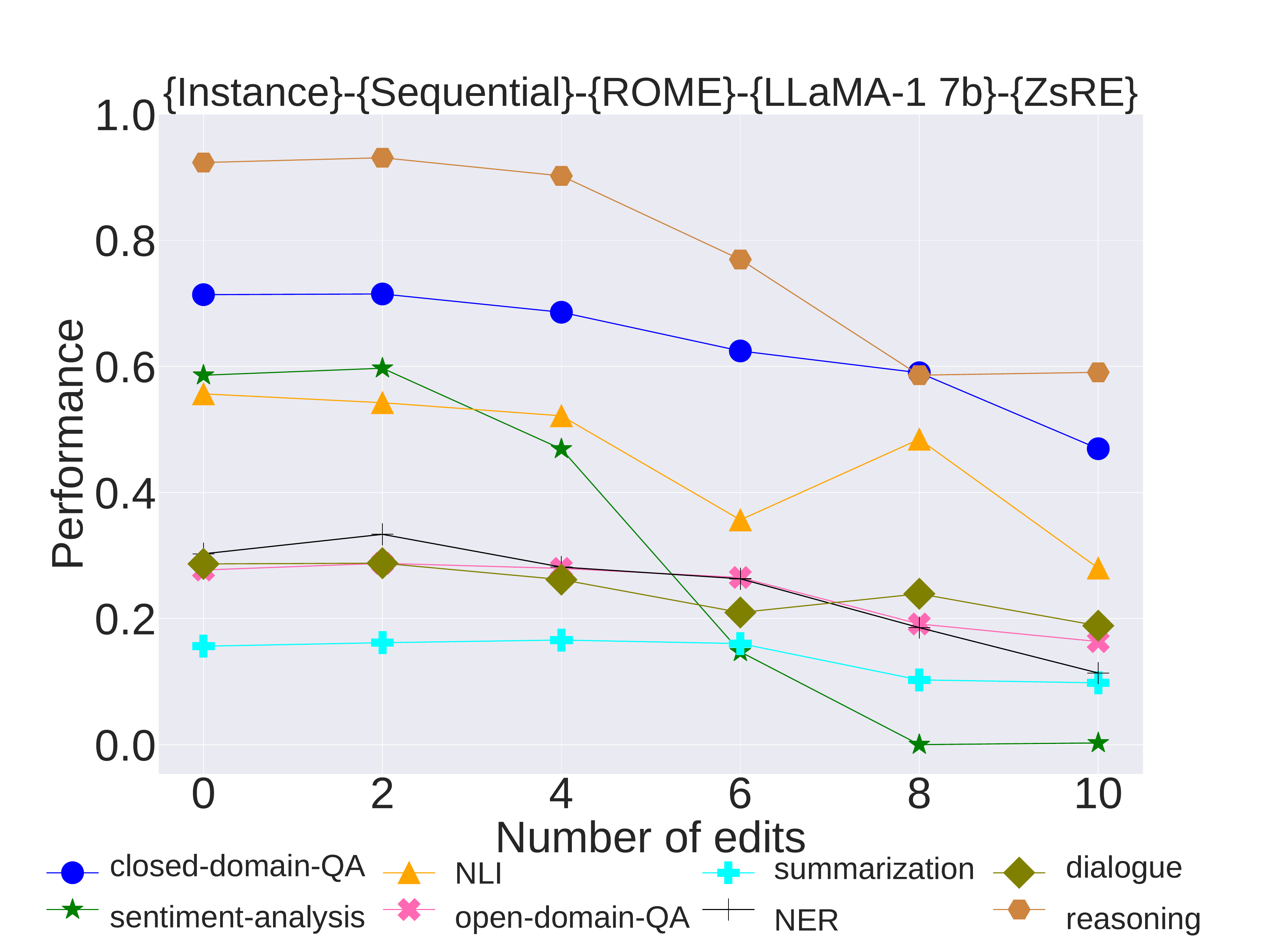}}
  \vspace{-3mm}
  \caption{Performance on general tasks of edited models using KN or ROME to edit GPT-2 XL or LLaMA-1 (7B) as the number of edits increases in \emph{instance- and sequential-editing}.}
  \vspace{-3mm}
  \label{fig-instance-sequential}
\end{figure*}

%%%%%%%%%%%%%%%%%%%%%%%%%%%%%%%%%%%%%%%%%%%%%%%%%%%%%%%%%%%%%%%%%%%%%%%%%%%%

  \paragraph{Editing Methods}
  Four popular editing methods were selected:
  (1) \emph{KN}~\cite{DBLP:conf/acl/DaiDHSCW22} involved identifying neurons linked to knowledge expression using gradient-based attributions and then enhancing the MLP layer by adding scaled embedding vectors to those specific neurons.
  (2) \emph{MEND}~\cite{DBLP:conf/iclr/MitchellLBFM22} learned a hypernetwork to produce weight updates by decomposing the fine-tuning gradients into rank-1 form. 
  (3) \emph{ROME}~\cite{DBLP:conf/nips/MengBAB22} involved localizing factual knowledge in a specific Transformer MLP layer and then updating it by directly writing new key-value pairs into the MLP module. 
  (4) \emph{MEMIT}~\cite{DBLP:conf/iclr/MengSABB23} expanded the capabilities of ROME by enabling the editing of large amounts of factual data through the updating of a sequence of MLP layers.
  It is notable that only MEND and MEMIT support batch-editing.
  All experiments were conducted using the EasyEdit tool~\cite{DBLP:journals/corr/abs-2308-07269}, ensuring standardized and reproducible evaluation.
  All editing instances were randomly sampled from the editing dataset.
  Readers can refer to Appendix~\ref{sec-edit-method} for details of these editing methods. 

  \paragraph{Editing Dataset}
  The popular model editing dataset Zero-Shot Relation Extraction \textsc{(ZsRE)}~\cite{DBLP:conf/conll/LevySCZ17} used in previous work~\cite{DBLP:conf/emnlp/CaoAT21,DBLP:conf/nips/MengBAB22,DBLP:conf/emnlp/YaoWT0LDC023} was adopted in our experiments.
  \textsc{ZsRE} is a QA dataset using question rephrasings generated by back-translation as the equivalence neighborhood.
  Each input is a question about an entity, and plausible alternative edit labels are sampled from the top-ranked predictions of a BART-base model trained on \textsc{ZsRE}.

  \paragraph{Selected LLMs}
  Experiments were conducted on three LLMs: 
  {GPT-2 XL} (1.5B)~\cite{radford2019language}, 
  {LLaMA-1} (7B)~\cite{DBLP:journals/corr/abs-2302-13971}, and 
  {LLaMA-2} (7B)~\cite{DBLP:journals/corr/abs-2307-09288}.

  \paragraph{Downstream Tasks and Metrics}
  To extensively explore whether model editing has side effects on the general abilities of LLMs, eight representative tasks were adopted:
  (1) \emph{Reasoning} on the GSM8K~\cite{DBLP:journals/corr/abs-2110-14168}, and the results were measured by solve rate.
  (2) \emph{Natural language inference (NLI)} on the RTE~\cite{DBLP:conf/mlcw/DaganGM05}, and the results were measured by accuracy of two-way classification.
  (3) \emph{Open-domain QA} on the Natural Question~\cite{DBLP:journals/tacl/KwiatkowskiPRCP19}, and the results were measured by exact match (EM) with the reference answer after minor normalization as in \citet{DBLP:conf/acl/ChenFWB17} and \citet{DBLP:conf/acl/LeeCT19}.
  (4) \emph{Closed-domain QA} on the BoolQ~\cite{DBLP:conf/naacl/ClarkLCK0T19}, and the results were also measured by EM.
  (5) \emph{Dialogue} on the MuTual~\cite{DBLP:conf/acl/CuiWLZZ20}, and the results were measured by selecting one best-matched response from four available candidates, denoted as Recall$_4$@1 as in \citet{DBLP:conf/sigdial/LowePSP15}. 
  (6) \emph{Summarization} on the SAMSum~\cite{gliwa-etal-2019-samsum}, and the results were measured by the average of ROUGE-1, ROUGE-2 and ROUGE-L as in \citet{lin-2004-rouge}.
  (7) \emph{Named entity recognition (NER)} on the CoNLL03~\cite{DBLP:conf/conll/SangM03}, and the results were measured by entity-level F1-score.
  (8) \emph{Sentiment analysis} on the SST2~\cite{DBLP:conf/emnlp/SocherPWCMNP13}, and the results were measured by accuracy of two-way classification. 

%%%%%%%%%%%%%%%%%%%%%%%%%%%%%%%%%%%%%%%%%%%%%%%%%%%%%%%%%%%%%%%%%%%%%%%%%%%%

\begin{figure*}[t]
  \centering
  \subfigure{
  \includegraphics[width=3.8cm]{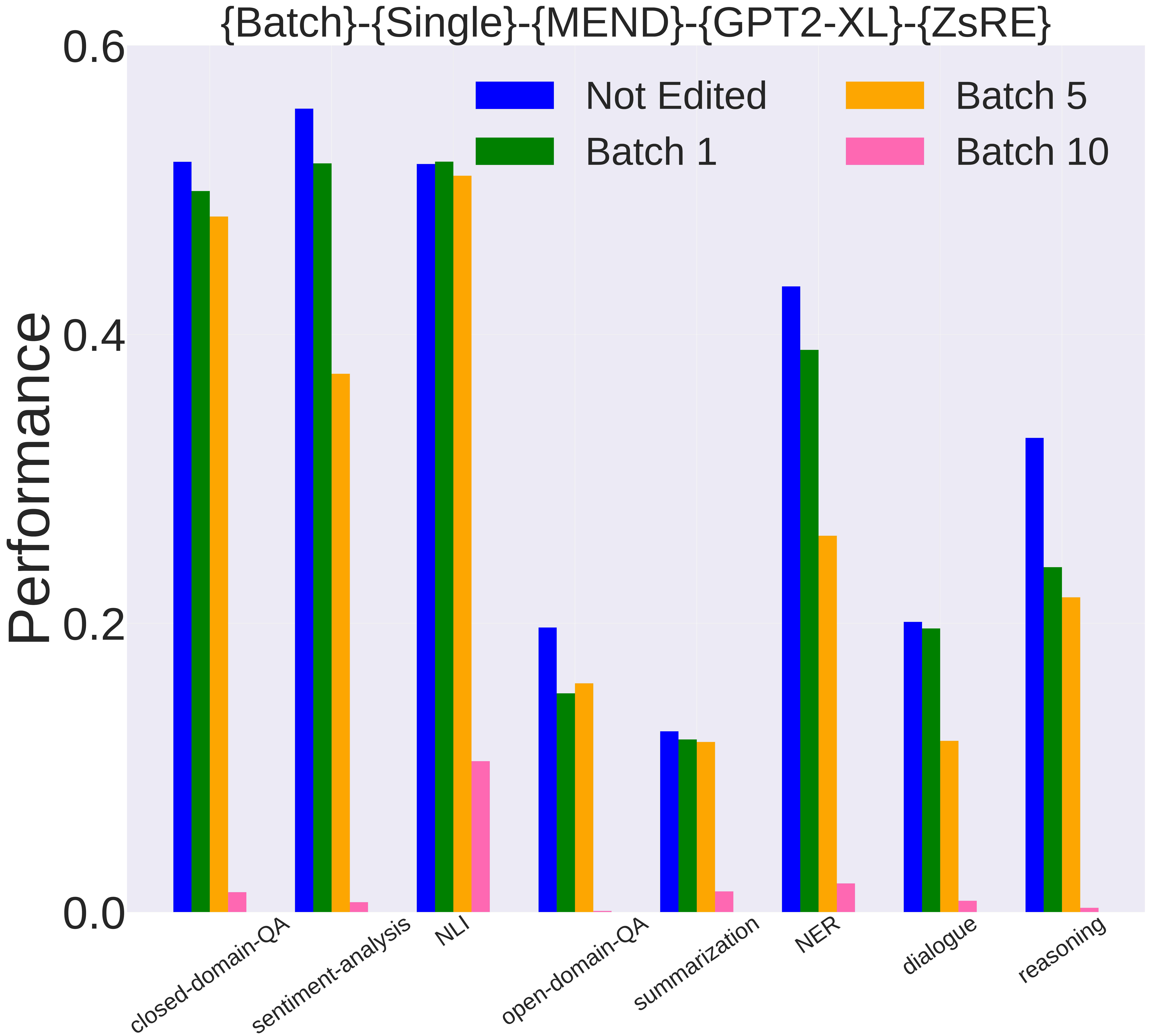}}
  \subfigure{
  \includegraphics[width=3.8cm,height=3.4cm]{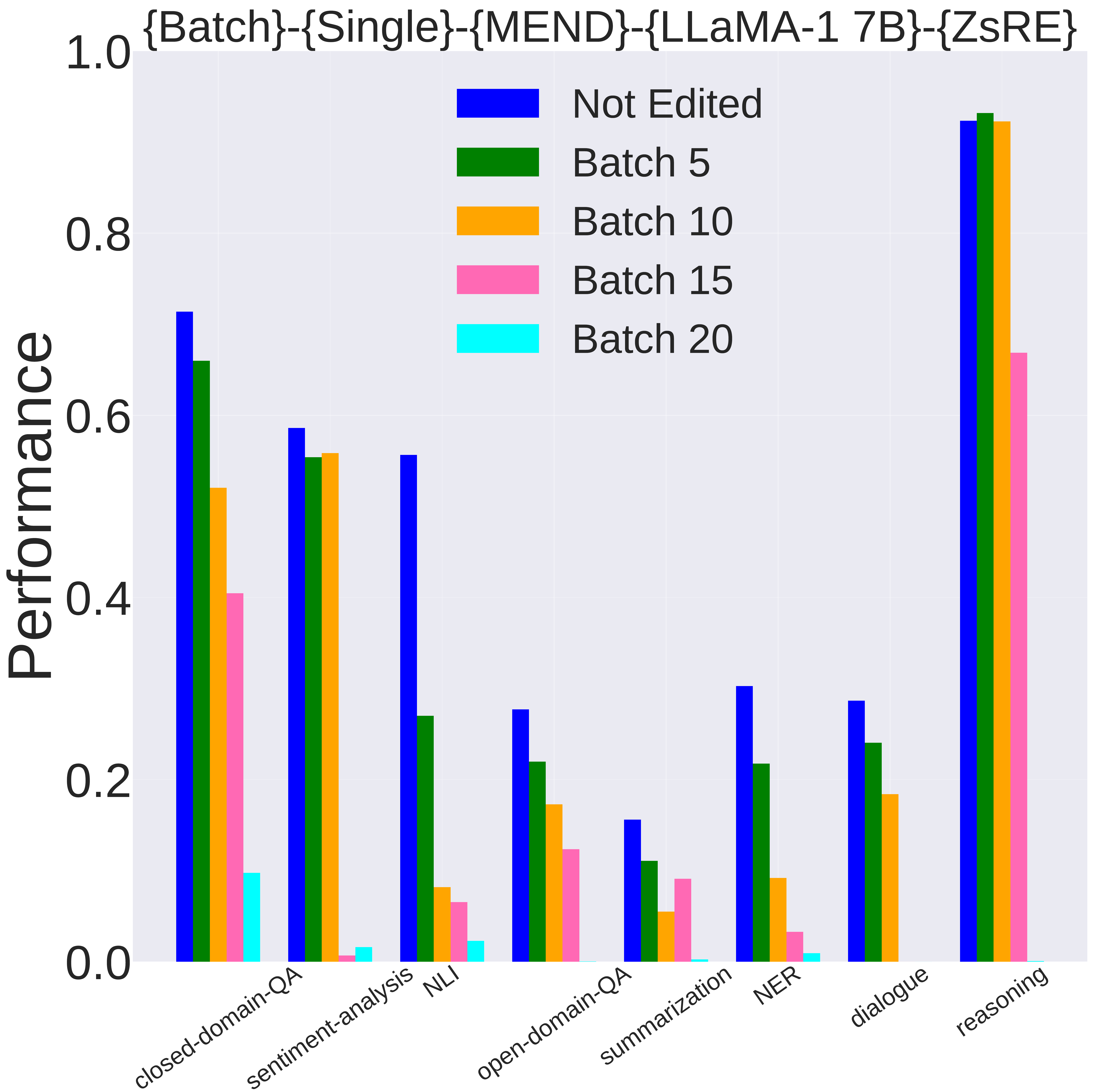}}
  \subfigure{
  \includegraphics[width=3.8cm]{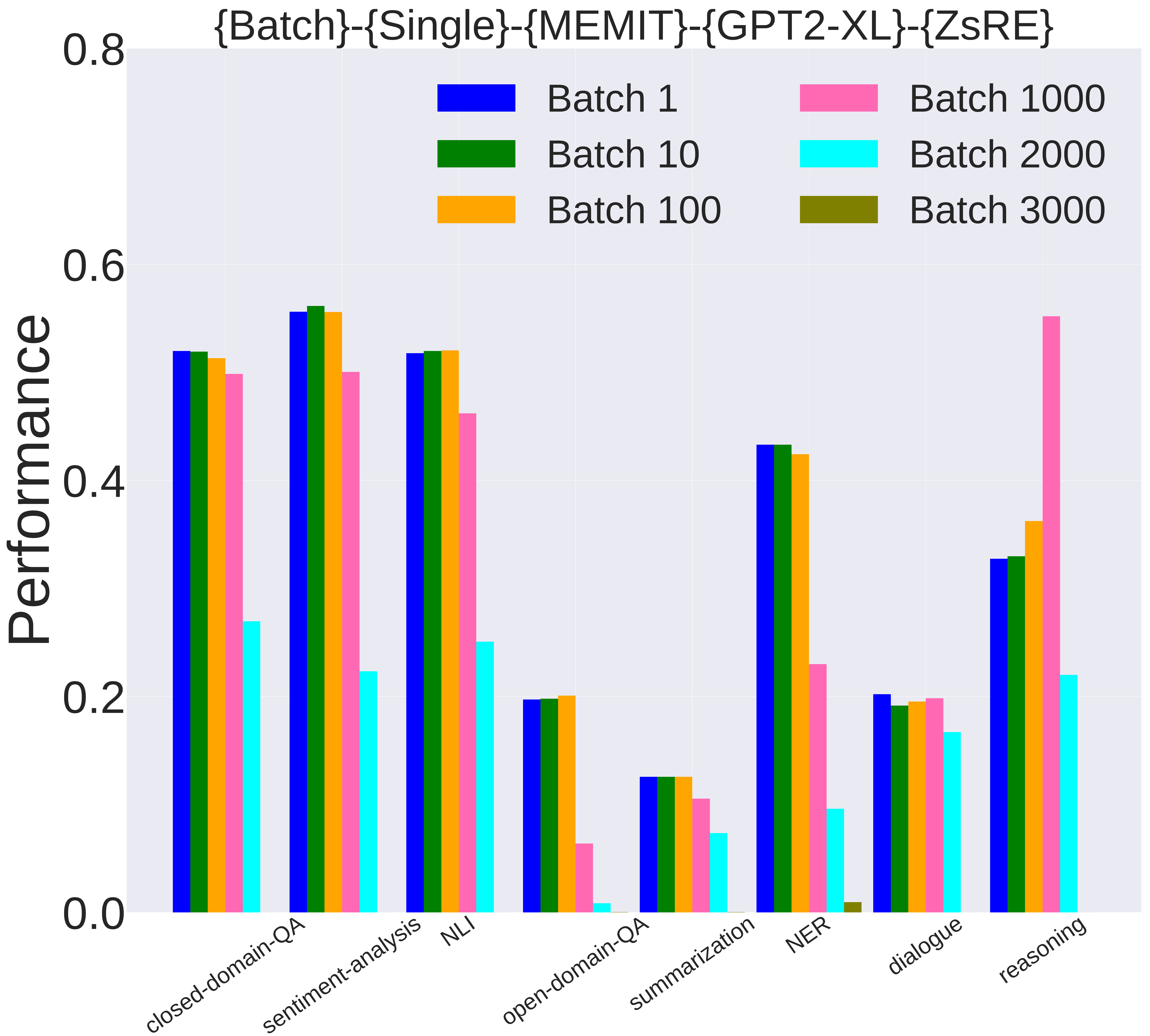}}
  \subfigure{
  \includegraphics[width=3.8cm,height=3.4cm]{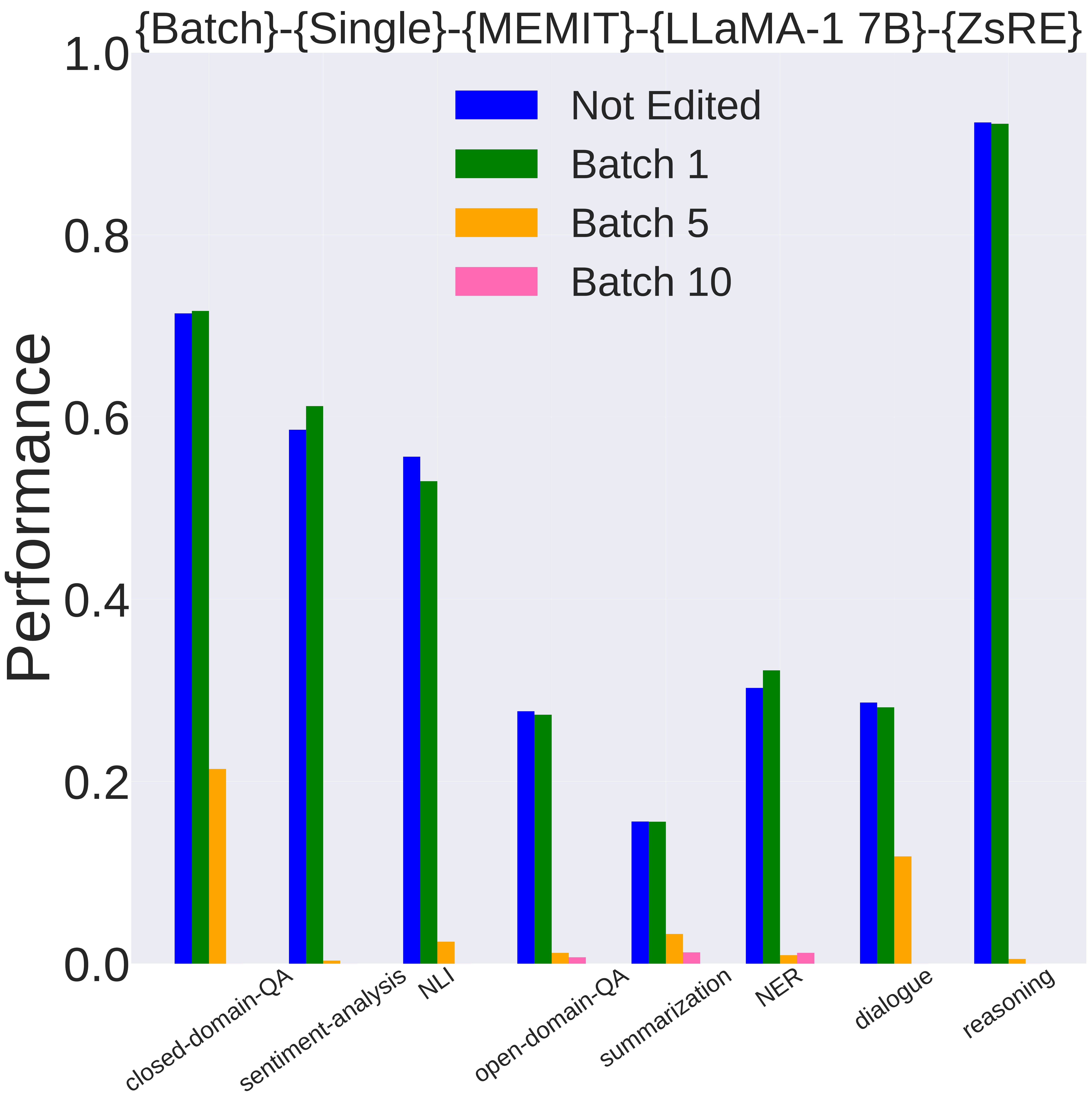}}
  \vspace{-3mm}
  \caption{Performance on general tasks of edited models using MEND or MEMIT to edit GPT-2 XL or LLaMA-1 (7B) with different batch sizes in \emph{batch- and single-editing}.}
  \vspace{-3mm}
  \label{fig-batch-single}
\end{figure*}

%%%%%%%%%%%%%%%%%%%%%%%%%%%%%%%%%%%%%%%%%%%%%%%%%%%%%%%%%%%%%%%%%%%%%%%%%%%%

  \subsection{Results}

  \paragraph{Impact of Sequential-editing} \label{sec-ins-seq}
  Since single-editing can be regarded as a special case of sequential-editing when the number of edits is 1, this subsection mainly discussed instance- and sequential-editing. 
  KN and ROME that support instance-editing but not batch-editing were adopted to facilitate this exploration.
  MEND and MEMIT that support batch- and sequential-editing will be explored later in this subsection.
  Figure~\ref{fig-instance-sequential} presents the performance on general tasks of edited models using KN or ROME to edit GPT-2 XL and LLaMA-1 (7B) as the number of edits increases.
  Due to limited space, readers can refer to Appendix~\ref{sec-appendix-result} for the results of editing LLaMA-2 (7B) which show similar trends.
  It can be seen that although there is only one instance per editing operation, the performance of edited models on various tasks fluctuates significantly and shows a downward trend as the number of edits increases.
  Strikingly, the use of KN resulted in a drastic performance degradation to nearly zero on all selected tasks with just a single edit. 
  These findings underscore two key insights.
  First, the selected LLMs are not robust to weight perturbations even if less than 1\% of the parameters are edited, whereby slight perturbations may significantly affect their general abilities. 
  Second, these outcomes also shed light on the challenging nature of effectively coupling current editing algorithms with LLMs. 
  The difficulty lies in the dual objective of improving model factuality while simultaneously maintaining their general abilities. 
  The observed trends indicate that existing editing algorithms face grand challenges in achieving this delicate balance, emphasizing the need for further research and development in the refinement of editing methodologies for LLMs.

  \paragraph{Impact of Batch-editing} \label{sec-bat-sin}
  This subsection delved into batch- and single-editing to explore the impact of batch size for scaling up the editing scope. 
  Only MEND and MEMIT that supported batch-editing were adopted to facilitate this exploration.
  Figure~\ref{fig-batch-single} presents the performance on general tasks of edited models using MEND or MEMIT to edit GPT-2 XL and LLaMA-1 (7B) with different batch sizes.
  Readers can refer to Appendix~\ref{sec-appendix-result} for the results of editing LLaMA-2 (7B).
  Remarkably, even with only one single editing operation, edited models exhibited a trend of performance degradation as the batch size increases in most cases.
  This consistent decrease in performance underlines the sensitivity of the models to increases in batch size, emphasizing the significance of carefully scaling knowledge editing for optimal updates.
  Therefore, we call for more research work on scalable editing to facilitate efficient editing of multiple instances.

  \paragraph{Impact of Batch- and Sequential-editing} \label{sec-bat-seq}
  In order to holistically take into account the interplay between batch size and sequential-editing, a joint setting of batch- and sequential-editing was explored to understand how these two factors collaboratively influence the overall performance of edited models.
  Figure~\ref{fig-batch-sequential-mend-gpt2xl} to Figure~\ref{fig-batch-sequential-memit-llama2-7b} in Appendix~\ref{sec-appendix-result} present the performance of using MEND or MEMIT to edit GPT-2 XL, LLaMA-1 (7B) and LLaMA-2 (7B) respectively as the number of edits increases.
  These results also echo our observations on sequential-editing, and those on batch-editing respectively. 
  
  \subsection{Analysis of Causes of Side Effects} \label{sec-analysis-side-effect}
  We show that the side effects of editing come from changing the original model weights too much, resulting in \emph{overfitting} to the editing facts.
  This phenomenon can be illustrated through statistics and visualization using ROME to edit GPT-2 XL.

%%%%%%%%%%%%%%%%%%%%%%%%%%%%%%%%%%%%%%%%%%%%%%%%%%%%%%%%%%%%%%%%%%%%%%%%%%%%

    \begin{table}[t]
      \vspace{2.5mm}
      \centering
      \setlength{\tabcolsep}{2.4pt}
      \resizebox{0.98\linewidth}{!}{
      \begin{tabular}{cccc}
      \toprule
        \textbf{\# edits}  & \textbf{Manhattan dist.}    &  \textbf{\% $\delta$ > 0.077} &  \textbf{\% $\delta$ > 0.171} \\
      \midrule 
          1                &           9079.2            &            20.0\%             &             10.0\%         \\
          5                &          27072.2            &            49.2\%             &             28.9\%         \\
          10               &          52245.3            &            67.4\%             &             46.2\%         \\
          15               &          63247.5            &            72.8\%             &             52.2\%         \\
      \bottomrule
      \end{tabular}
      }
      \caption{Statistics of the distinction between the final edited weight and the original unedited weight via weight change as the number of edits increases.}
      \vspace{-2mm}
      \label{tab-overfit-statistics}
    \end{table}

%%%%%%%%%%%%%%%%%%%%%%%%%%%%%%%%%%%%%%%%%%%%%%%%%%%%%%%%%%%%%%%%%%%%%%%%%%%%

  \paragraph{Statistics}
  We first show how the weights change in instance- and single-editing.
  Typically, one editing operation is to add an edit update weight $\Delta W$ to the original weight $W$, where $\Delta W$ is calculated aiming to insert new editing facts. 
  Here, we define the absolute value of the relative change in weight $ \delta = |\frac{\Delta W}{W}|$ to characterize the degree of change of each element in the update weight $\Delta W$.
  The statistics show that only 20\% of the elements in the update weight $\Delta W$ have $\delta$ greater than 0.077, while only 10\% of the elements have $\delta$ greater than 0.171. 
  These results are averaged by 100 random single edits in the ZsRE dataset. 
  It can be seen that the update weight $\Delta W$ might be quite sparse, while most elements in $\Delta W$ are minor. 
  Manhattan distance between the updated and original weights is also calculated as a measurement of distinction.
  
  Furthermore, how the weights change in instance- and sequential-editing is shown in Table~\ref{tab-overfit-statistics}. 
  As the number of edits increases, the proportion of elements whose $\delta$ is greater than a certain threshold increases significantly, and the weight is also more differentiated than the original weight.
  Therefore, the accumulation of overfitting across multiple edits can amplify changes to the original weights.

\begin{figure}[t]
  \centering
  \subfigure[1 edit]{
  \includegraphics[width=1.6cm]{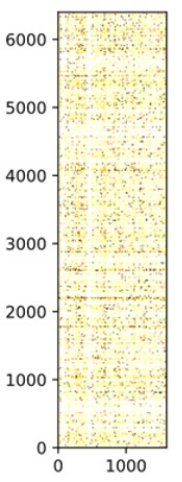}}
  \subfigure[5 edits]{
  \includegraphics[width=1.55cm]{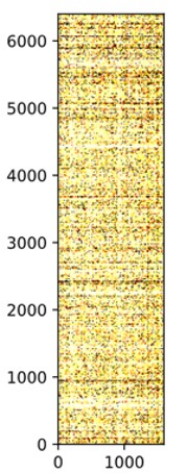}}
  \subfigure[10 edits]{
  \includegraphics[width=1.5cm]{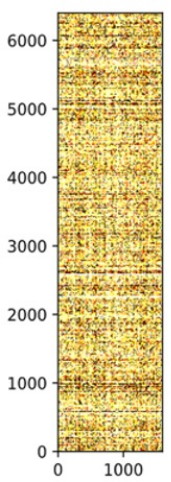}}
  \subfigure[15 edits]{
  \includegraphics[width=2.3cm,height=4.2cm]{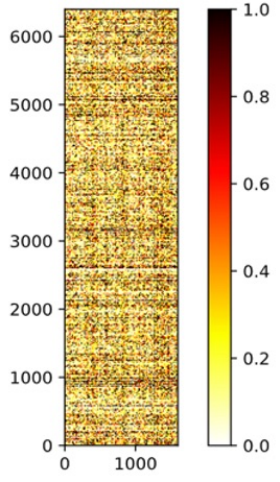}}
  \vspace{-2mm}
  \caption{Visualization of the distinction between the final edited weight and the original unedited weight via weight change $|\Delta W|$ as the number of edits increases.}
  \label{fig-overfit-visual}
\end{figure}

  \paragraph{Visualization}
  The distinction between the final edited weight and the original unedited weight is illustrated by visualizing the weight change $|\Delta W|$ as shown in Figure~\ref{fig-overfit-visual}.
  It reveals the consistent findings that the update weight $\Delta W$ might be quite sparse, while the accumulation across multiple edits can amplify changes to the original weights.

%% file: 5-regularization.tex
\section{\MODELNAME{}: \textbf{RE}lative \textbf{C}hange in weigh\textbf{T}}

  \subsection{Approach}
  We have analyzed the causes of side effects in Section~\ref{sec-analysis-side-effect} that model editing changes the original model weights too much, resulting in \emph{overfitting} to the editing facts.
  This type of editing overfitting occurs when a model learns to fit the new editing data too closely, capturing noise and outliers in the data rather than the underlying patterns. 
  Furthermore, the gradual buildup of editing-induced overfitting across sequential edits can severely impair the general abilities of LLMs. 
  Consequently, while such models may exhibit proficiency in the new editing facts, they often struggle to generalize effectively across a spectrum of downstream tasks.
  This phenomenon underscores the importance of mitigating overfitting during the editing process to ensure both the improvement of model factuality and the maintenance of their general abilities.
  
\begin{figure}[t]
\centering
\includegraphics[width=0.48\textwidth]{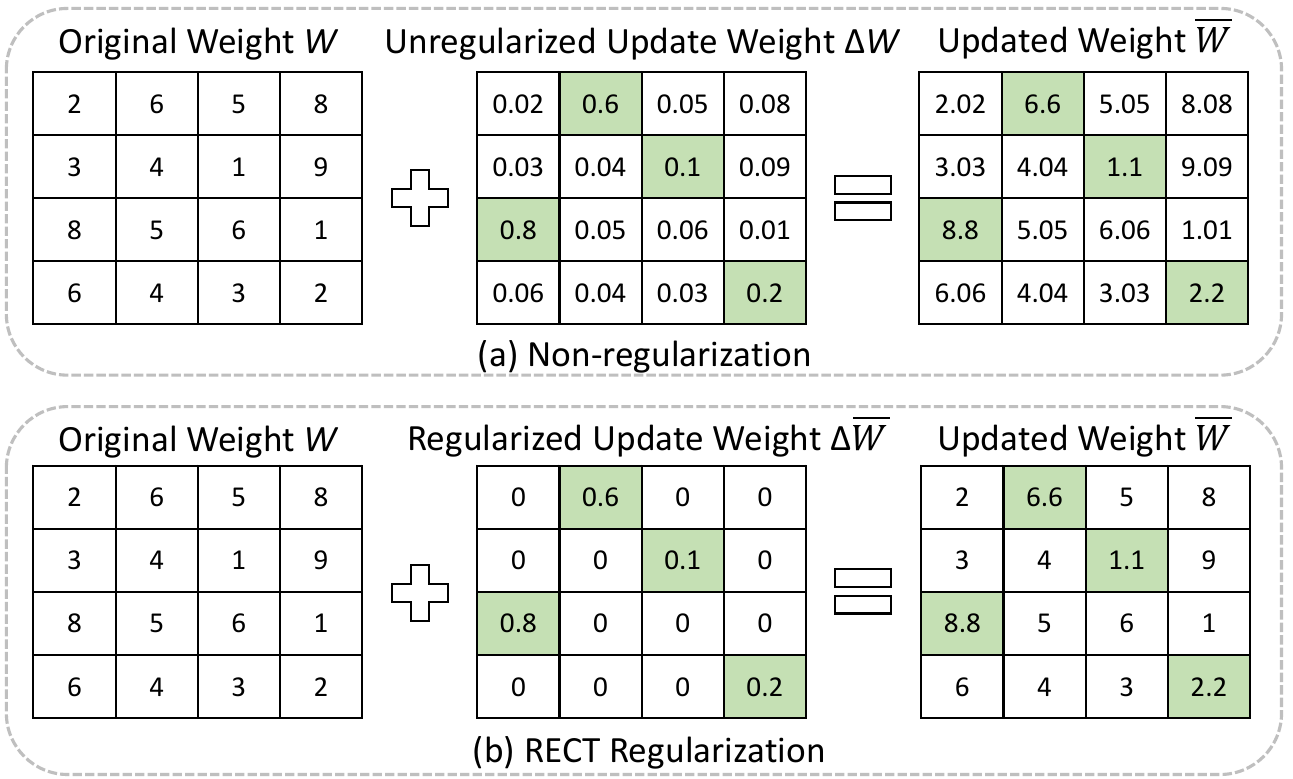}
\vspace{-2mm}
\caption{Comparison of (a) non-regularization and (b) the proposed \MODELNAME{} regularization. 
Elements in green denote the top-\emph{k}\% that change the most according to $\delta$ and are considered as the principle editing information,which should keep their original values. 
\emph{k} = 25 in this figure for illustration.} 
\vspace{-2mm}
\label{fig-rect}
\end{figure}

%%%%%%%%%%%%%%%%%%%%%%%%%%%%%%%%%%%%%%%%%%%%%%%%%%%%%%%%%%%%%%%%%%%%%%%%%%%%

\begin{figure*}[t]
  \centering
  \subfigure[ROME on GPT-2 XL]{
  \includegraphics[width=3.8cm]{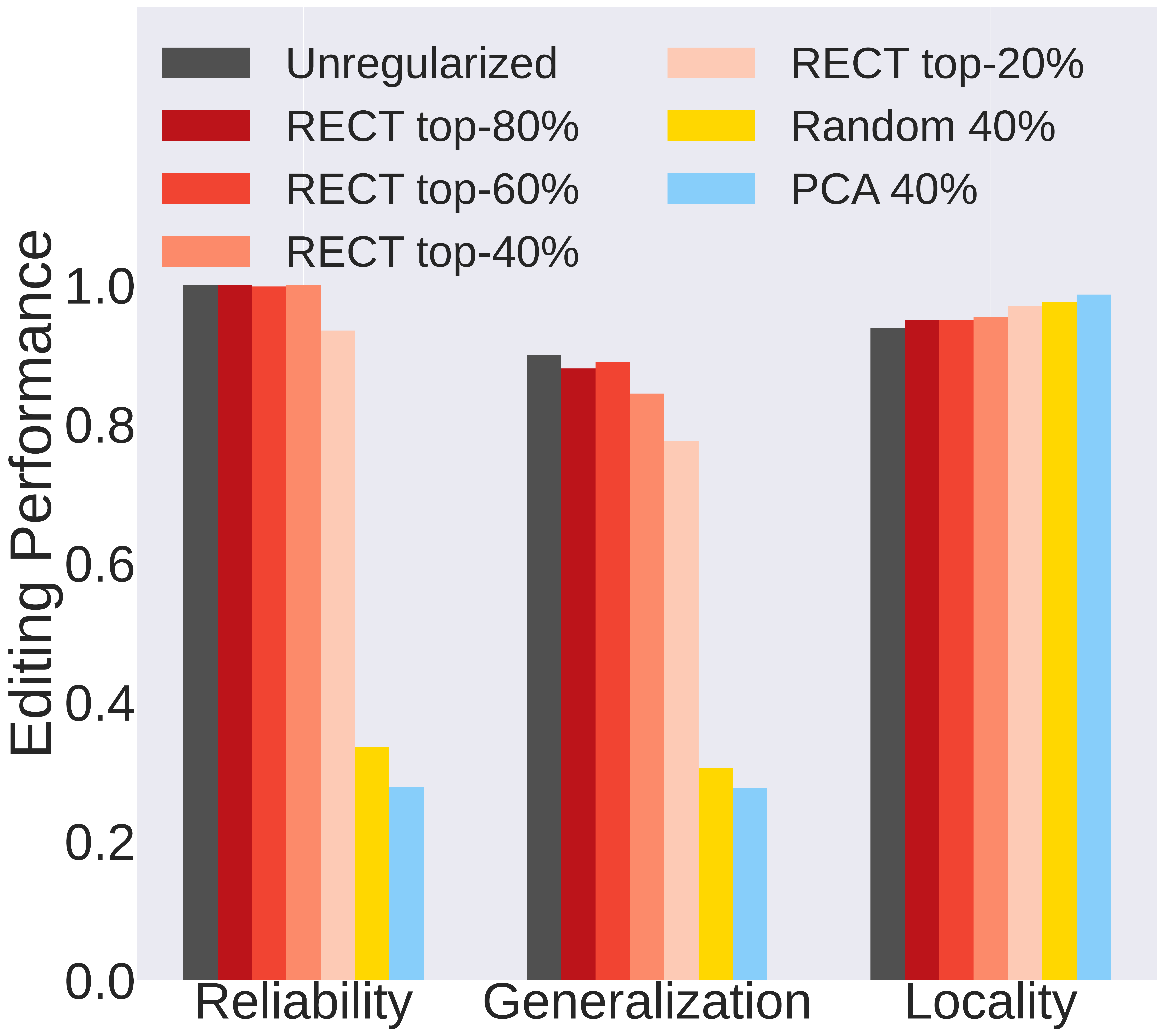}}
  \subfigure[ROME on LLaMA-1 (7B)]{
  \includegraphics[width=3.8cm]{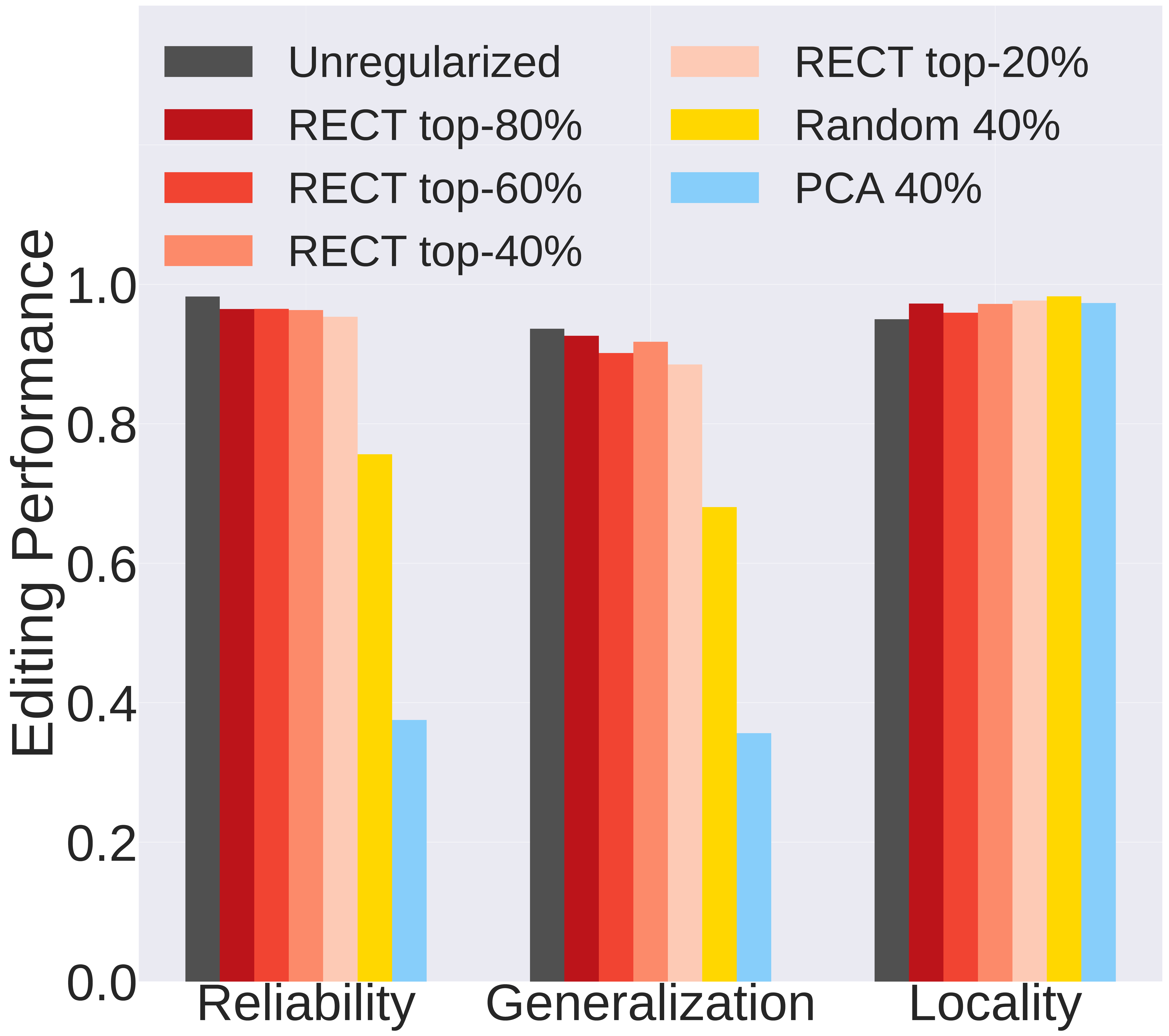}}
  \subfigure[MEMIT on GPT-2 XL]{
  \includegraphics[width=3.8cm]{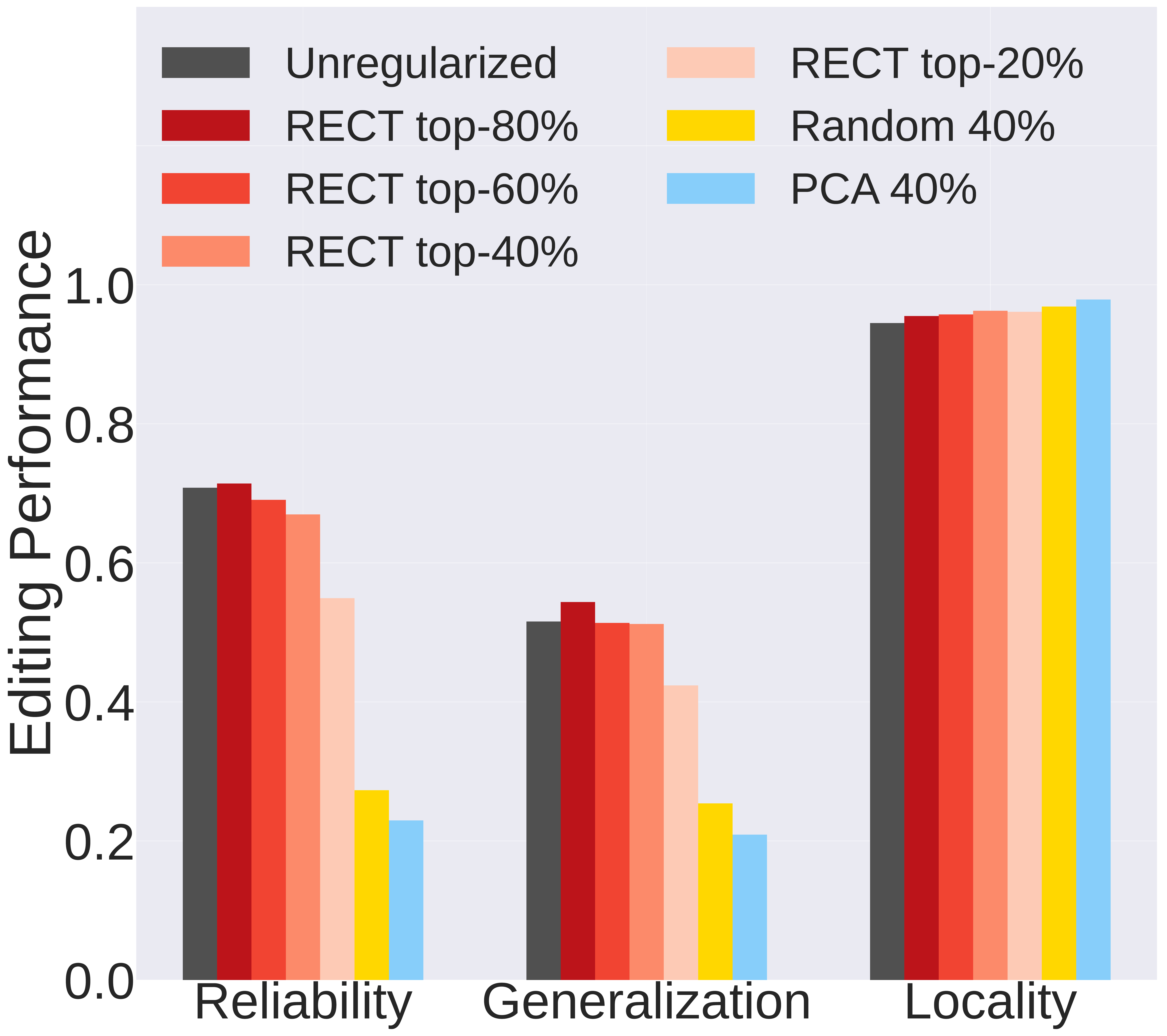}}
  \subfigure[MEMIT on LLaMA-1 (7B)]{
  \includegraphics[width=3.8cm]{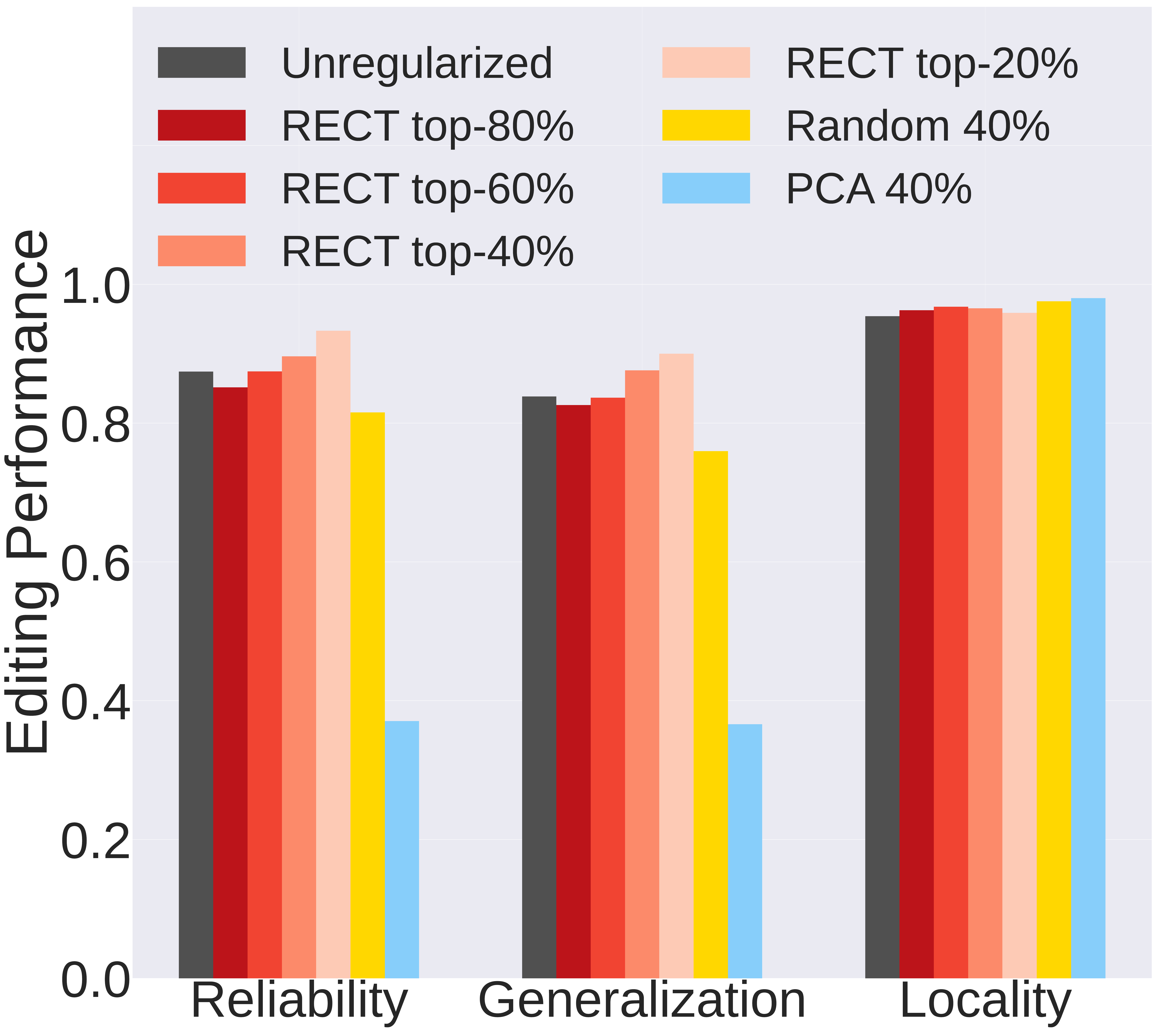}}
  \vspace{-2mm}
  \caption{Comparison of introducing various regularization methods and how the \emph{editing performance} change with respect to different top-\emph{k}\% for \MODELNAME{}.
  }
  \label{fig-rect-edit}
\end{figure*}

%%%%%%%%%%%%%%%%%%%%%%%%%%%%%%%%%%%%%%%%%%%%%%%%%%%%%%%%%%%%%%%%%%%%%%%%%%%%

\begin{figure*}[t]
  \centering
  \subfigure[Summarization]{
  \includegraphics[width=3.8cm]{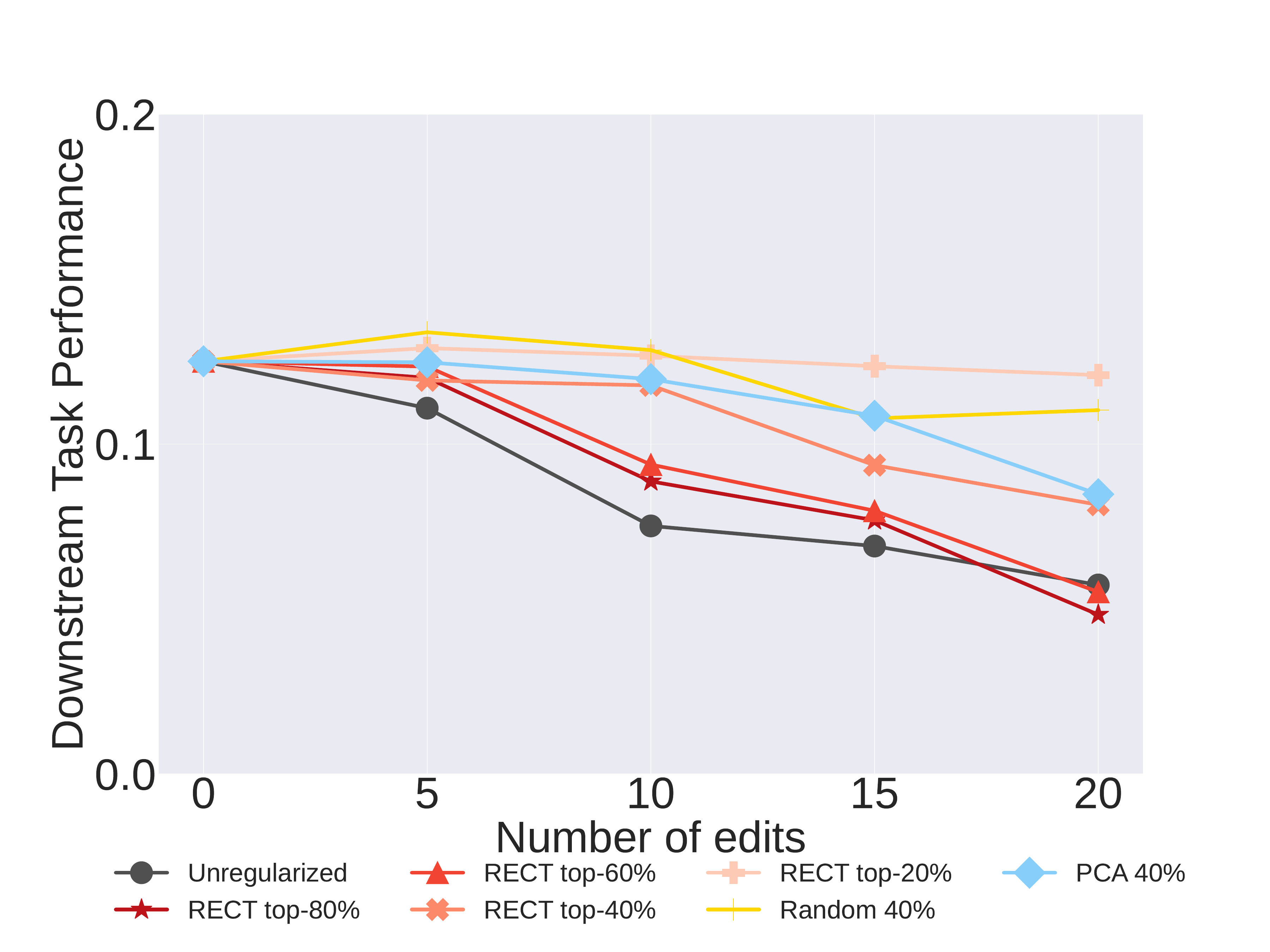}}
  \subfigure[Open-domain QA]{
  \includegraphics[width=3.8cm]{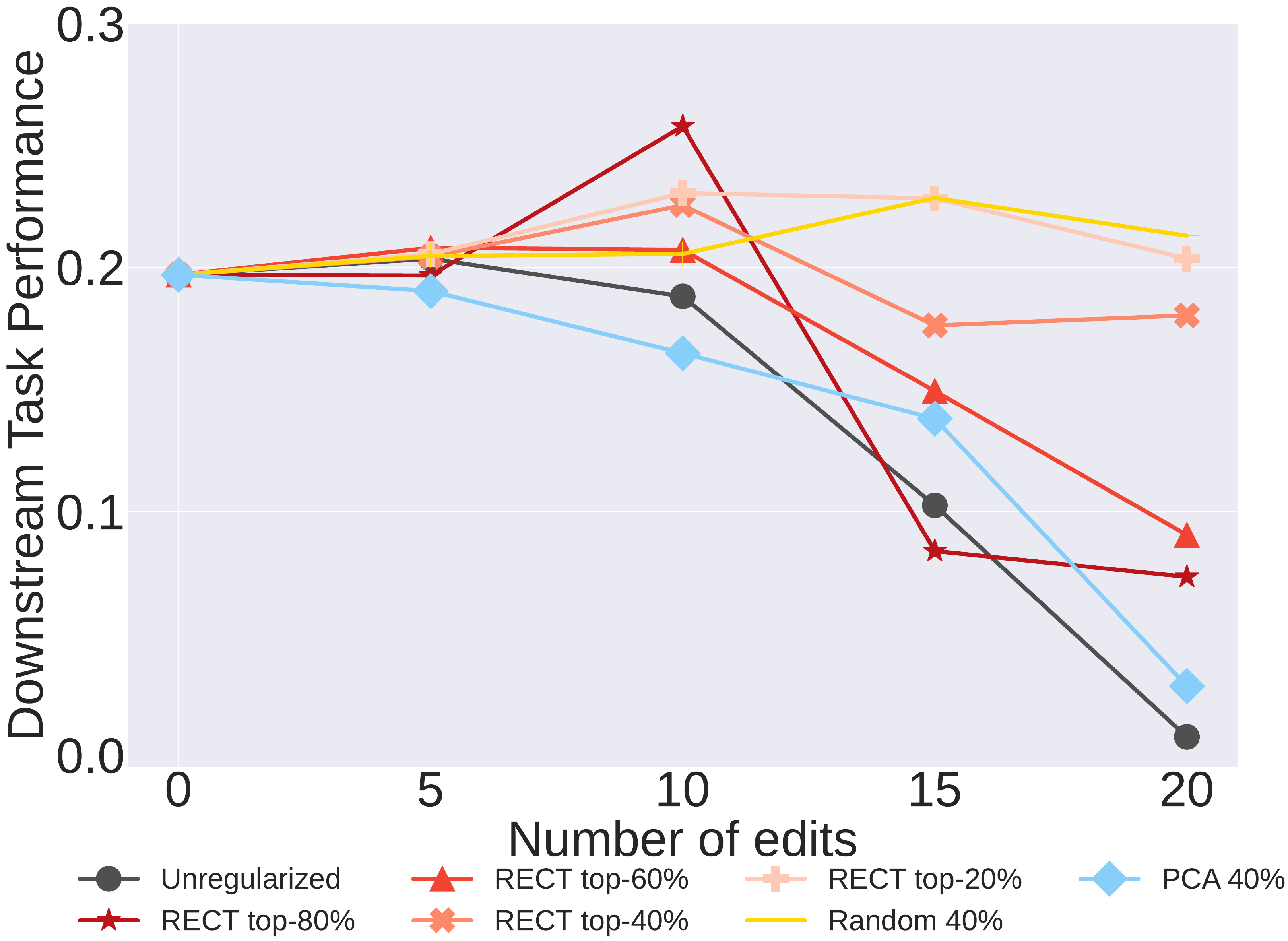}}
  \subfigure[Closed-domain QA]{
  \includegraphics[width=3.8cm]{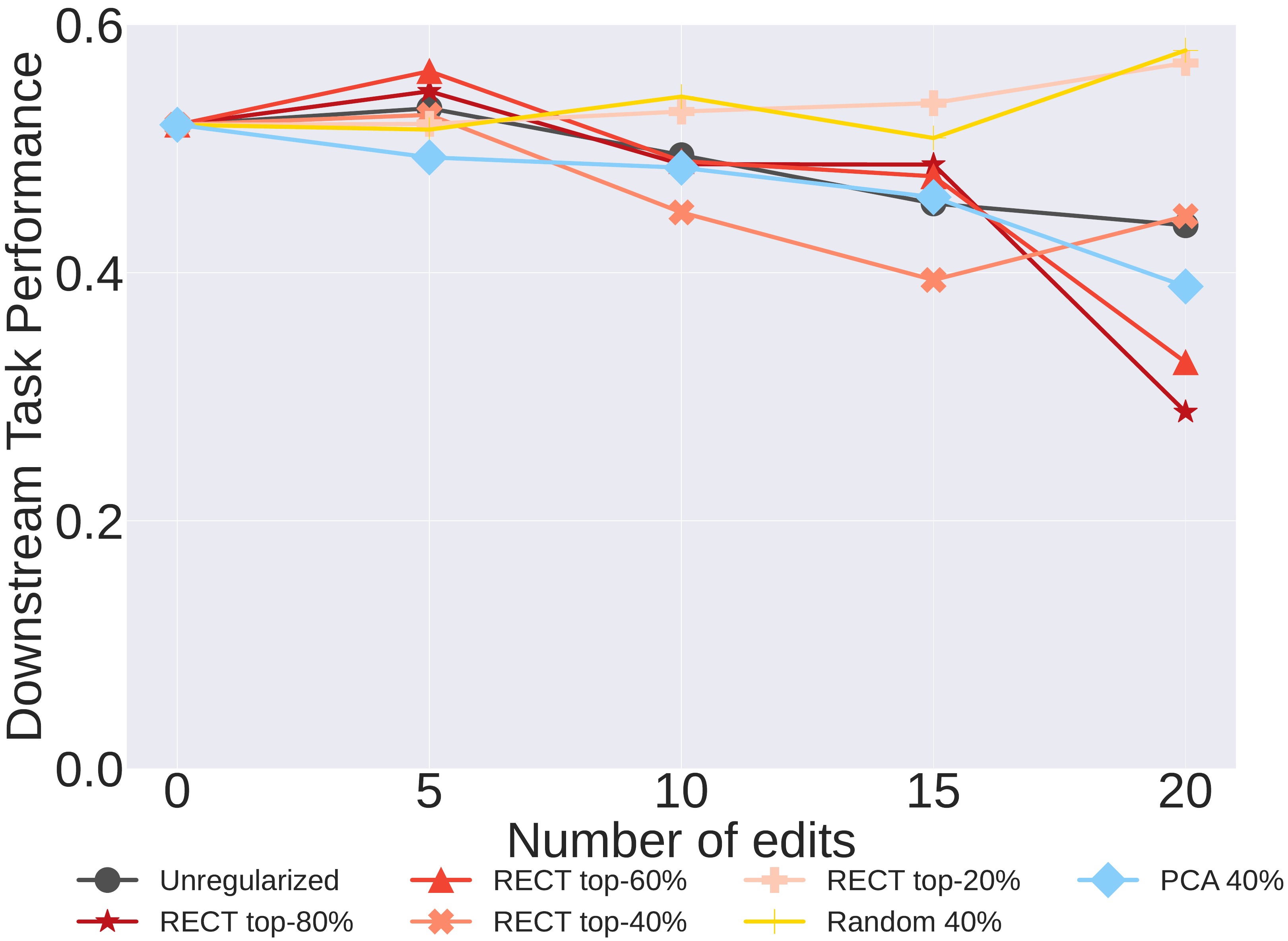}}
  \subfigure[Sentiment Analysis]{
  \includegraphics[width=3.8cm]{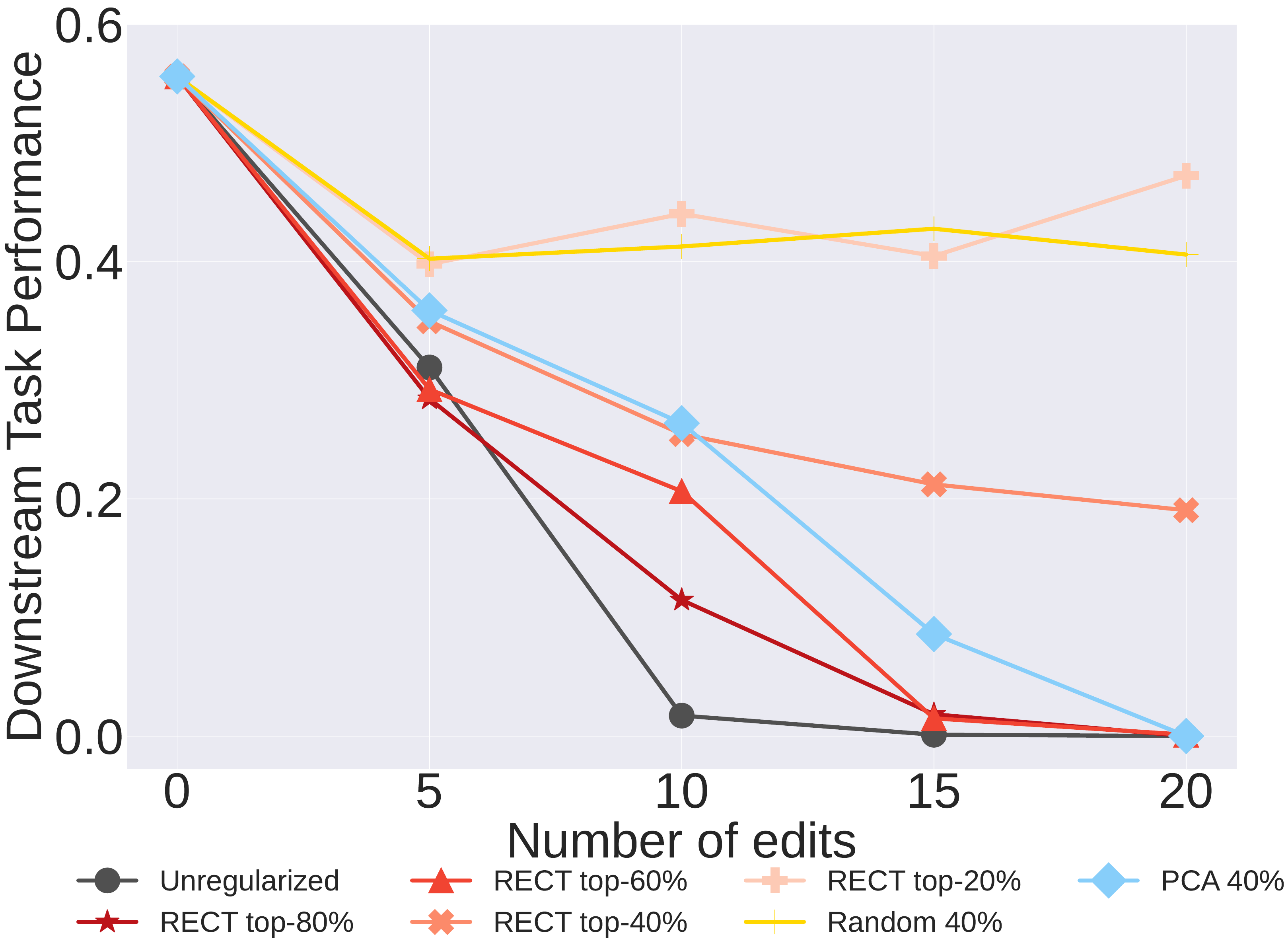}}
  \vspace{-2mm}
  \caption{Comparison of introducing various regularization methods and how the \emph{downstream task performance} change with respect to different top-\emph{k}\% for \MODELNAME{}.
  }
  \vspace{-2mm}
  \label{fig-rect-downstream}
\end{figure*}

%%%%%%%%%%%%%%%%%%%%%%%%%%%%%%%%%%%%%%%%%%%%%%%%%%%%%%%%%%%%%%%%%%%%%%%%%%%%

  To this end, this paper designs a regularization method named \textbf{RE}lative \textbf{C}hange in weigh\textbf{T} (\MODELNAME{}) to prevent editing overfitting.
  Figure~\ref{fig-rect} illustrates the overview of this regularization method.
  Typically, one editing operation is to add an edit update weight $\Delta W$ to the original weight $W$ to derive the updated weight $\overline{W}$, where $\Delta W$ is calculated aiming to insert a batch of $N$ new editing facts $\{(s, r, o^{*})_i\}_{i=1}^N$ ($N=1$ for a single editing fact).
  Formally, we have:
  \begin{align}
    \overline{W} &= W + \Delta W, \\ \label{equ-edit}
    \Delta W     &= f ( \{(s, r, o^{*})_i\}_{i=1}^N ),
  \end{align}
  where function $f$ denotes the calculation method of update weight $\Delta W$ for different editing methods, e.g., ROME~\cite{DBLP:conf/nips/MengBAB22}.

  Here, we define the absolute value of the relative change in weight $ \delta = |\frac{\Delta W}{W}|$ to characterize the degree of change of each element in the update weight $\Delta W$.
  To some extent, $\delta$ can be used to indicate the importance of each element in $\Delta W$ when inserting the new editing facts.
  On the one hand, a portion of the elements in the update weight $\overline{W}$ are assumed to constitute the core components of the new editing facts. 
  On the other hand, the remaining elements are assumed minor contributions to editing.
  Specifically, the top-\emph{k}\% elements in $\Delta W$ that change the most according to $\delta$ are considered as the principal editing information and keep their original values.
  While for the remaining elements in $\Delta W$, they are treated as minor contributions to editing and set to zero for regularization.
  Mathematically, we have the \emph{regularized} edit update weight $\Delta \overline{W}$ as:
  \begin{equation}
  \begin{aligned}
    \Delta \overline{W}_{ij} = 
        \begin{cases}
            \Delta W_{ij}  & \text{if $\delta_{ij}$ in the top-\emph{k}\%}, \\
            0              & \text{else}.
        \end{cases}
  \end{aligned}
  \end{equation}  
  Finally, the regularized edit update weight $\Delta \overline{W}$ is added to the original weight to derive the regularized updated weight.
  Essentially, \MODELNAME{} functions to deter the implementation of excessively intricate editing updates that have a higher propensity to result in overfitting. 
  By imposing constraints on the complexity of editing updates, it serves as a safeguard against the model's inclination to adapt too closely to the editing data, thus promoting more generalizable and reliable model performance.

  \subsection{Regularization Baselines}
  To demonstrate the effectiveness and efficiency of the proposed method \MODELNAME{}, we compared it with the following baselines, including:
  \textbf{Unregularized} keeps the full elements of the edit update weight $\Delta W$.
  \textbf{Random \emph{k}\%} selects the random \emph{k}\% elements of $\Delta W$.
  \textbf{PCA \emph{k}\%} compresses the most important editing information in $\Delta W$ into \emph{k}\% elements via principal component analysis (PCA), and sets the remaining elements to zero.

  \subsection{Results of \MODELNAME{}}
  The effectiveness of a regularization method should be illustrated from two perspectives. 
  First, regularizing the edit update weight should not harm its editing performance, i.e., edited models should still remember the new editing facts and generalize to related facts.
  Second, the regularized edited models should be able to preserve the general abilities compared with unregularized ones.

  \paragraph{Editing Performance}
  Figure~\ref{fig-rect-edit} presents the the results of regularizing ROME or MEMIT on GPT-2 XL or LLaMA-1 (7B). Readers can refer to Appendix~\ref{sec-appendix-result-rect} for the results on LLaMA-2 (7B).
  From these results we can have the following findings.
  First, compared with unregularized $\Delta W$, \MODELNAME{} that keeps the original values of an appropriate amount of top-40\% elements in $\Delta W$ and sets the remaining elements to zero can help maintain over 94\% majority of reliability and generalization, and even improve locality.    
  It is natural that reliability and generalization slightly drop when setting partial elements to zero since partial editing information is removed.
  The reason why locality is improved is probably because those elements with low $\delta$ corresponding to some noise and outliers in the editing data are removed to prevent from editing overfitting, so the edited models are more robust. 
  However, setting excessive elements in $\Delta W$ to zero, e.g., \MODELNAME{} top-20\%, might hurt the editing performance as partial important editing information is accidentally removed.
  Furthermore, compared with Random 40\% and PCA 40\%, \MODELNAME{} top-40\% achieves the best performance, indicating its effectiveness in selecting the most principal editing information.
  It is notable that \MODELNAME{} also exhibits advantages in terms of efficiency, since it eliminates the complex calculations required in PCA.

  \paragraph{General Downstream Task Performance}
  Figure~\ref{fig-rect-downstream} presents how the downstream task performance change with respect to introducing various regularization methods to edit GPT-2 XL. Readers can refer to Appendix~\ref{sec-appendix-result-rect} for the results on more downstream tasks.
  From these results we can have the following findings.
  As the proportion of elements in $\Delta W$ set to 0 increases, the more editing overfitting is regularized, the smaller the change to the original weight, so the general abilities can be more preserved.
  Results show that regularized edited models are able to preserve the general abilities compared with unregularized ones in most tasks such as summarization, open- and closed-domain QA.
  It is worth noting that it still poses a challenge for some tasks such as sentiment analysis, and remains unclear whether it works for larger number of edits, which will be left to future work.

%% file: 6-conclusion.tex
\section{Conclusion}
  Model updating technology has been catalyzing the continuous iteration of advanced and trustworthy LLMs.
  This paper studies model editing and for the first time raises concerns whether model editing has any side effects on the general abilities of LLMs.
  The systematical evaluation reveals that current methods unintentionally hurt the general abilities of LLMs no matter in instance- or batch-editing, and single- or sequential-editing.
  Our analysis of the causes reveals that model editing results in overfitting to the editing facts, and the accumulation of overfitting across multiple edits can amplify the negative impact.
  The proposed \MODELNAME{} regularization method has been proven to effectively prevent overfitting of new editing facts, thus preserving both the editing and general downstream task performance.

\section*{Impact Statements}
  As LLMs play an increasingly crucial role in various applications, mitigating the hallucinations caused by missing, false or outdated knowledge encapsulated within the parameters is imperative for ensuring the reliability of their outputs. 
  However, the potential trade-off between improving the factuality and degrading the general abilities underscores the need for a balanced approach. 
  Striking the right balance in model editing is crucial to prevent unintended consequences and to preserve the broader abilities of LLMs, contributing to the sustainable advancement of AI technology. 
  This paper highlights the importance of considering not only the immediate gains in factuality but also the long-term impacts on the general performance and applicability of LLMs, encouraging a thoughtful and comprehensive exploration of model editing techniques for responsible AI development.
  More importantly, this paper calls for more efforts and underscores the collective focus on \emph{strengthening the robustness of LLMs to weight perturbations, developing innovative paradigms for model editing, and designing comprehensive evaluation of model editing}.
  By doing so, we can collectively advance the continual development of LLMs, paving the way for more reliable applications in real-world scenarios.

\section*{Limitations}
  This paper studies the side effects of editing based on the ZsRE editing dataset, while more complex and diverse side effects are hypothesized to exist and thus need to be explored on more editing datasets in future work.
  In addition, although sometimes a method of side effect mitigation is effective for a certain number of edits, it remains to be seen whether the method will still be effective for a larger number of edits.
  It is expected that one editing method outperforms another in terms of the number of edits given the same requirements of maintaining editing performance and general abilities.
  This paper does not further explore whether the proposed method can still be effective for more edits, which is worth further study.
  While we primarily propose to mitigate the side effects of model editing from a statistical perspective, the bottleneck of the general abilities of edited models should be analyzed theoretically.

%% file: 7-appendix.tex
\section{Details of Evaluation Metrics} \label{sec-appendix-metrics}
  
  \paragraph{Reliability} Given an editing fact (\emph{s = United States, r = President of, o = Donald Trump, $o^*$ = Joe Biden}), it could be regarded edited effectively if the edited model $f^*$ assigns a higher probability to the statement ``\emph{The President of the United States is Joe Biden}'' than the original prediction (\emph{Donald Trump}).
  
  \paragraph{Generalization} Edited models should be able to recall the updated knowledge when prompted within the editing scope.
  For example, 
  the paraphrased prompts like ``\emph{Who currently holds the office of President of the United States?}'' or ``\emph{Who is the current president of the US?}'' can be used.
  The edited model $f^{*}$ is considered to have generalized successfully if it can recall the editing memory, in this case, ``\emph{Joe Biden}''.
  
  \paragraph{Locality} The edited model $f^{*}$ should remain unchanged in response to prompts that are irrelevant or outside the scope of its editing. 
  For example, the answer to the question ``\emph{Who is the President of France?}'' should still correctly be ``\emph{Emmanuel Macron}''.

\section{Task Prompts} \label{sec-appendix-prompt}

The prompts for each task were illustrated in Table~\ref{tab-prompt}.

\begin{table*}[h]
\centering
\begin{tabular}{p{0.95\linewidth}}
\toprule

Reasoning:\\
Q: \{\texttt{QUESTION}\} A: Let's think step by step. \{\texttt{HINT}\} Therefore, the answer (arabic numerals) is:\\

\midrule

NLI:\\
\{\texttt{SENTENCE1}\} entails the \{\texttt{SENTENCE2}\}. True or False? answer:\\

\midrule

Open-domain QA:\\
Refer to the passage below and answer the following question. Passage: \{\texttt{DOCUMENT}\} Question: \{\texttt{QUESTION}\}\\

\midrule

Closed-domain QA:\\
Please answer the given question based on the passage. The answer should be exact 'yes' or 'no'. passage: \{\texttt{PASSAGE}\} question: \{\texttt{QUESTION}\}. answer:\\

\midrule

Dialogue:\\
Q: \{\texttt{ARTICLE}\} Which choice is correct? Answer Choices: (A)\{\texttt{OPTION0}\} (B)\{\texttt{OPTION1}\} (C)\{\texttt{OPTION2}\} (D)\{\texttt{OPTION3}\} A: Among A through D, the answer is\\

\midrule

Summarization:\\
\{\texttt{DIALOGUE}\} TL;DR:\\

\midrule

NER:\\
Please identify Person Entity from the given text. Text: \{\texttt{SENTENCE}\} Entity:\\
Please identify Location Entity from the given text. Text: \{\texttt{SENTENCE}\} Entity:\\
Please identify Organization Entity from the given text. Text: \{\texttt{SENTENCE}\} Entity:\\
Please identify Miscellaneous Entity from the given text. Text: \{\texttt{SENTENCE}\} Entity:\\

\midrule

Sentiment analysis:\\
For each snippet of text, label the sentiment of the text as positive or negative. The answer should be exact 'positive' or 'negative'. text: \{\texttt{TEXT}\} answer:\\

\bottomrule
\end{tabular}
\caption{The prompts to LLMs for evaluating their zero-shot performance on these general tasks.}
\label{tab-prompt}
\end{table*}

\clearpage
\section{Editing Methods \& Datasets} \label{sec-edit-method}

    \begin{table}[!hbt]
      \vspace{2mm}
      \centering
      \resizebox{0.6\linewidth}{!}{
      \begin{tabular}{lcccc}
      \toprule
        \multirow{2}{*}{\textbf{Paradigm}}   & \textbf{Editing}  & \textbf{Additional} & \textbf{Batch} & \textbf{Editor}      \\
                                             & \textbf{Methods}  & \textbf{Training}   & \textbf{Edit}  & \textbf{Parameters}  \\
      \midrule
        Meta-learn                           & MEND     & Yes        & Yes   & \emph{Model$_{hyper}$ + L*MLP}  \\  
      \midrule
        Locate-                              & KN       & No         & No    & \emph{L*neuron}        \\
        then-                                & ROME     & No         & No    & \emph{MLP$_{proj}$}    \\
        edit                                 & MEMIT    & No         & Yes   & \emph{L*MLP$_{proj}$}  \\
      \bottomrule
      \end{tabular}
      }
      \caption{Comparisons between several popular model editing methods following~\citet{DBLP:conf/emnlp/YaoWT0LDC023}.
      \textbf{Additional Training} refers to whether the methods need training before conducting specific edits. 
      \textbf{Batch Edit} refers to editing multiple target knowledge simultaneously. 
      \textbf{Editor Parameters} refers to the parameters that need to be updated for editing. \emph{L} denotes the number of layers to update. \emph{MLP} is FFN and \emph{MLP$_{proj}$} is the second linear layer in FFN. \emph{neuron} denotes the key-value pair in FFN. 
      }
      \vspace{-2mm}
      \label{tab-edit-method}
    \end{table}

  Four popular editing methods as compared in Table~\ref{tab-edit-method} were selected to measure their performance in improving factuality as well as their impairment in the general abilities of LLMs, including: 
  \textbf{KN}~\cite{DBLP:conf/acl/DaiDHSCW22}\footnote{https://github.com/EleutherAI/knowledge-neurons} first selected neurons that were associated with knowledge expression via gradient-based attributions, and then modified MLP layer at the rows corresponding to those neurons by adding scaled embedding vectors.
  \textbf{MEND}~\cite{DBLP:conf/iclr/MitchellLBFM22}\footnote{https://github.com/eric-mitchell/mend} learned a hypernetwork to produce weight updates by decomposing the fine-tuning gradients into rank-1 form.
  \textbf{ROME}~\cite{DBLP:conf/nips/MengBAB22}\footnote{https://github.com/kmeng01/rome} first localized the factual knowledge at a specific layer in the transformer MLP modules, and then updated the knowledge by directly writing new key-value pairs in the MLP module.
  \textbf{MEMIT}~\cite{DBLP:conf/iclr/MengSABB23}\footnote{https://github.com/kmeng01/memit} expanded the capabilities of ROME by enabling the editing of large amounts of factual data through the updating of a sequence of MLP layers.

  All experiments were conducted using the EasyEdit tool~\cite{DBLP:journals/corr/abs-2308-07269}, ensuring standardized and reproducible evaluation.
  All editing instances were randomly sampled from the editing dataset.

  The popular model editing dataset Zero-Shot Relation Extraction \textsc{(ZsRE)}~\cite{DBLP:conf/conll/LevySCZ17} used in previous work~\cite{DBLP:conf/emnlp/CaoAT21,DBLP:conf/nips/MengBAB22,DBLP:conf/emnlp/YaoWT0LDC023} was adopted in our experiments.
  \textsc{ZsRE} is a QA dataset using question rephrasings generated by back-translation as the equivalence neighborhood.
  Each input is a question about an entity, and plausible alternative edit labels are sampled from the top-ranked predictions of a BART-base model trained on \textsc{ZsRE}.

\clearpage
\section{Extensive Evaluation Results} \label{sec-appendix-result}

\subsection{Results of Side Effects of Model Editing}

\begin{figure*}[!hbt]
  \centering
  \subfigure{
  \includegraphics[width=4.5cm]{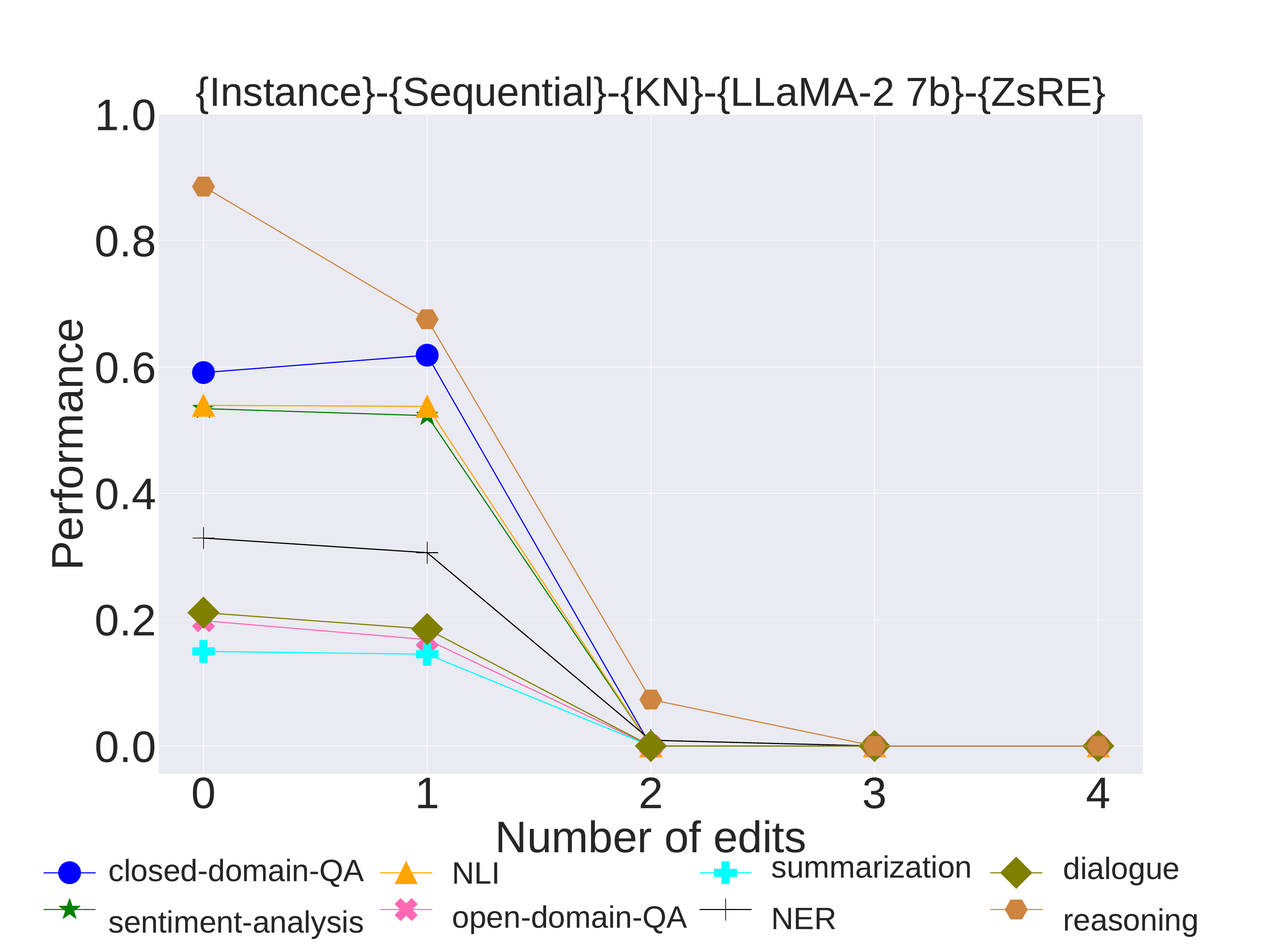}}
  \subfigure{
  \includegraphics[width=4.5cm]{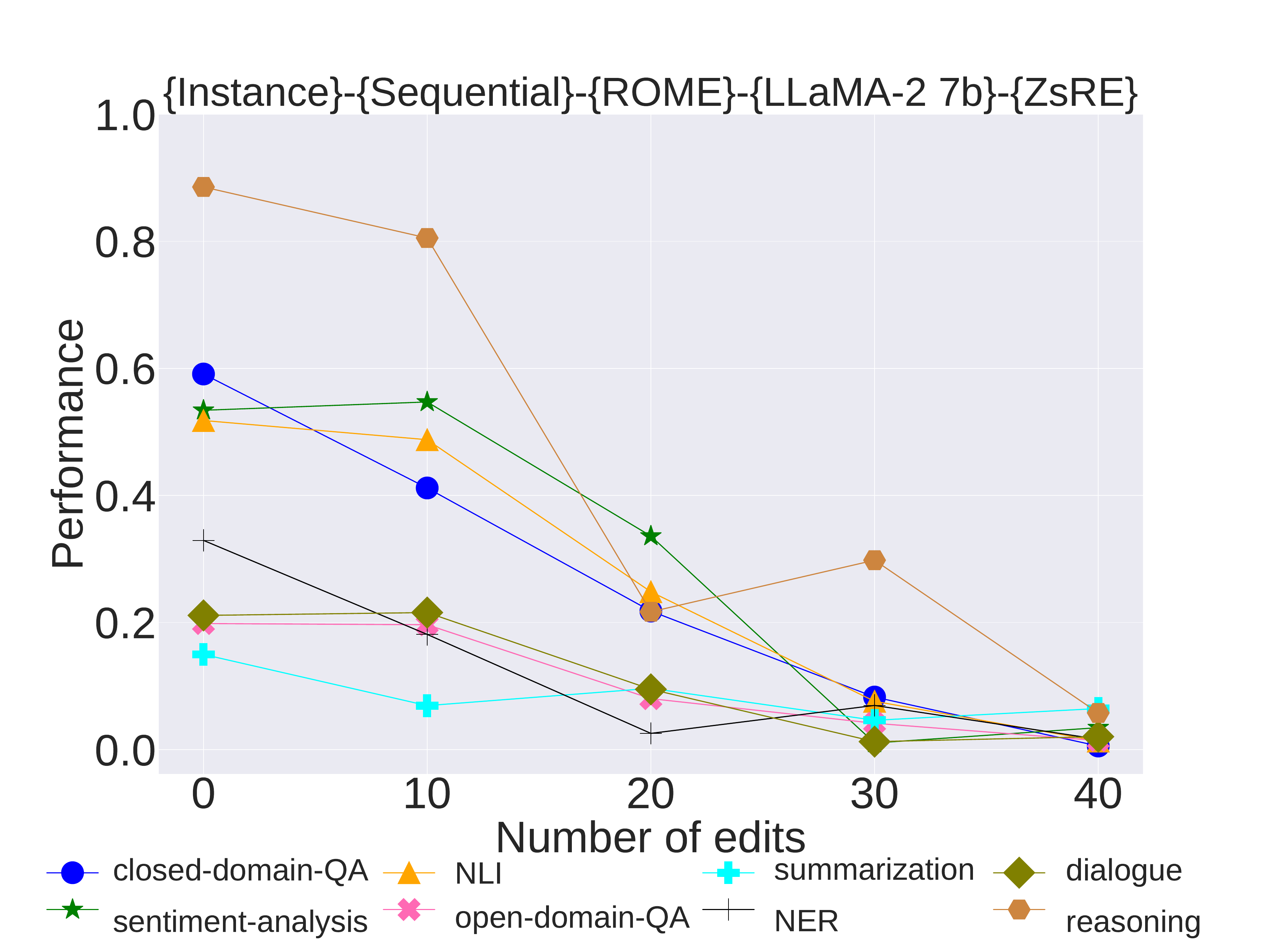}}
  \vspace{-2mm}
  \caption{Performance on general tasks of edited models using KN or ROME to edit LLaMA-2 (7B) as the number of edits increases in \emph{instance- and sequential-editing}.}
  \vspace{-3mm}
  \label{fig-instance-sequential-2}
\end{figure*}

%%%%%%%%%%%%%%%%%%%%%%%%%%%%%%%%%%%%%%%%%%%%%%%%%%%%%%%%%%%%%%%%%%%%%%%%%%%%

\begin{figure*}[!hbt]
  \centering
  \subfigure{
  \includegraphics[width=4.5cm]{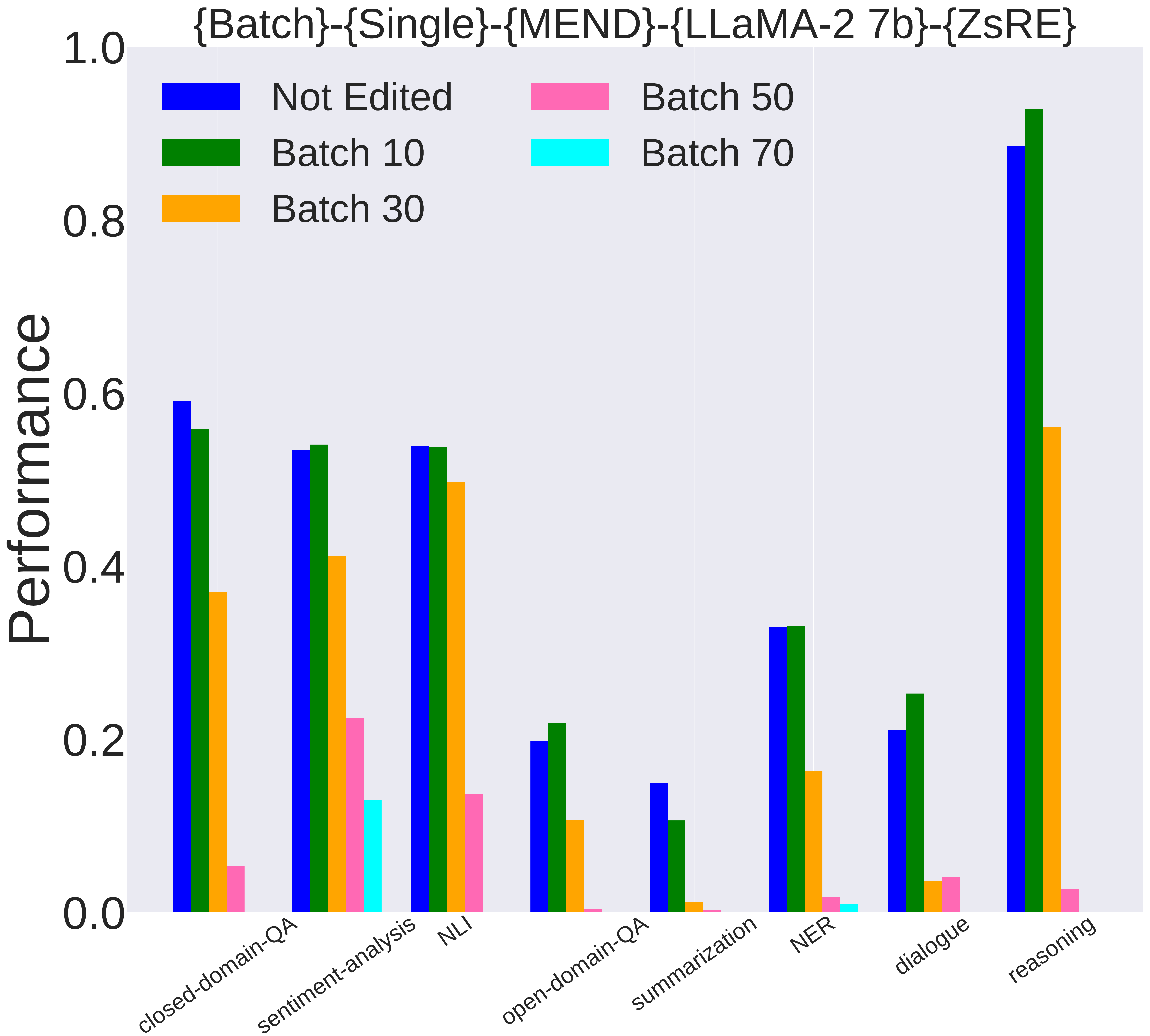}}
  \subfigure{
  \includegraphics[width=4.5cm]{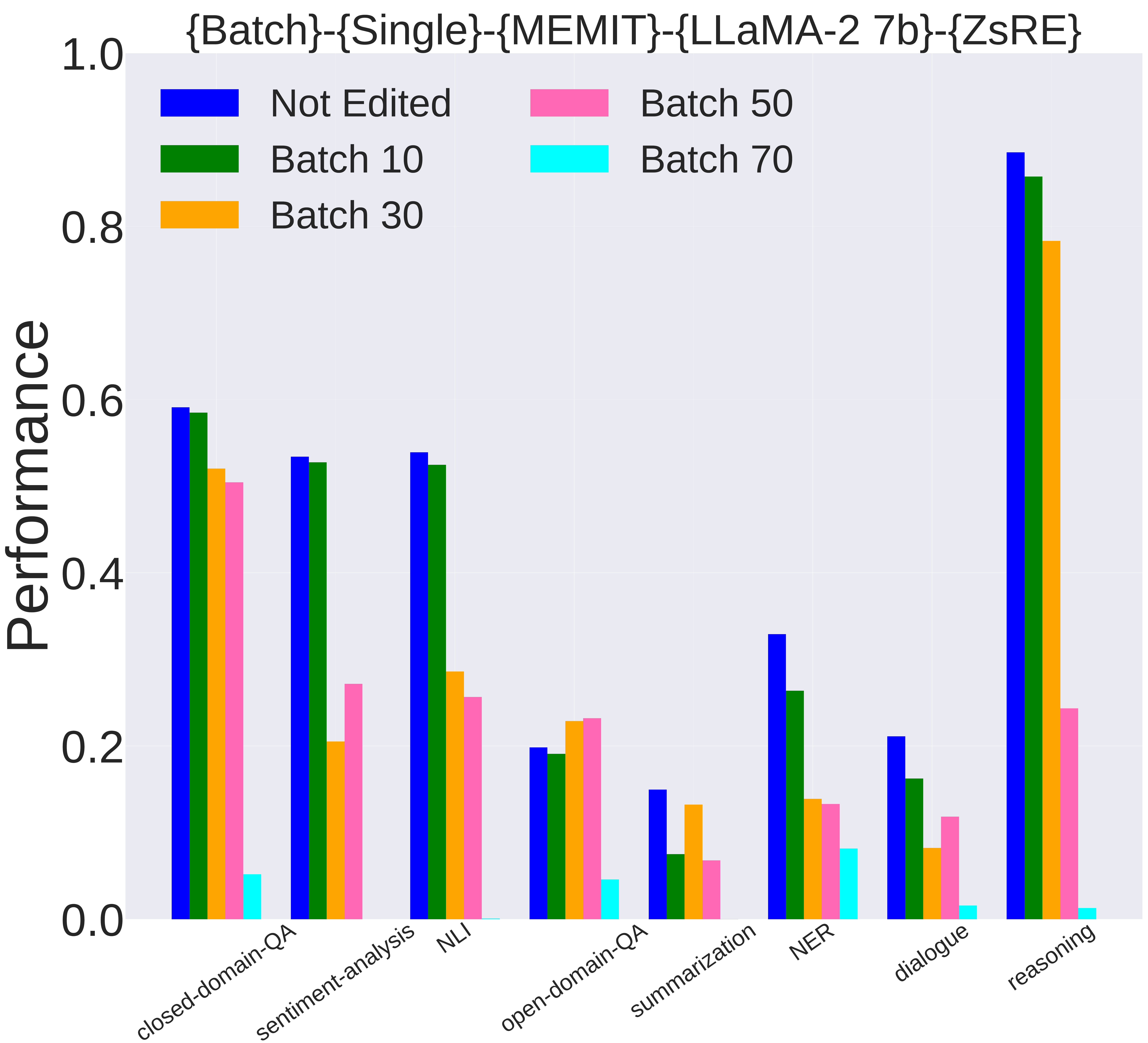}}
  \vspace{-4mm}
  \caption{Performance on general tasks of edited models using MEND or MEMIT to edit LLaMA-2 (7B) with different batch sizes in \emph{batch- and single-editing}.}
  \vspace{-4mm}
  \label{fig-batch-single-2}
\end{figure*}

%%%%%%%%%%%%%%%%%%%%%%%%%%%%%%%%%%%%%%%%%%%%%%%%%%%%%%%%%%%%%%%%%%%%%%%%%%%%

\clearpage

%%%%%%%%%%%%%%%%%%%%%%%%%%%%%%%%%%%%%%%%%%%%%%%%%%%%%%%%%%%%%%%%%%%%%%%%%%%%

\begin{figure*}[!hbt]
  \centering
  \subfigure[Reasoning]{
  \includegraphics[width=3.6cm]{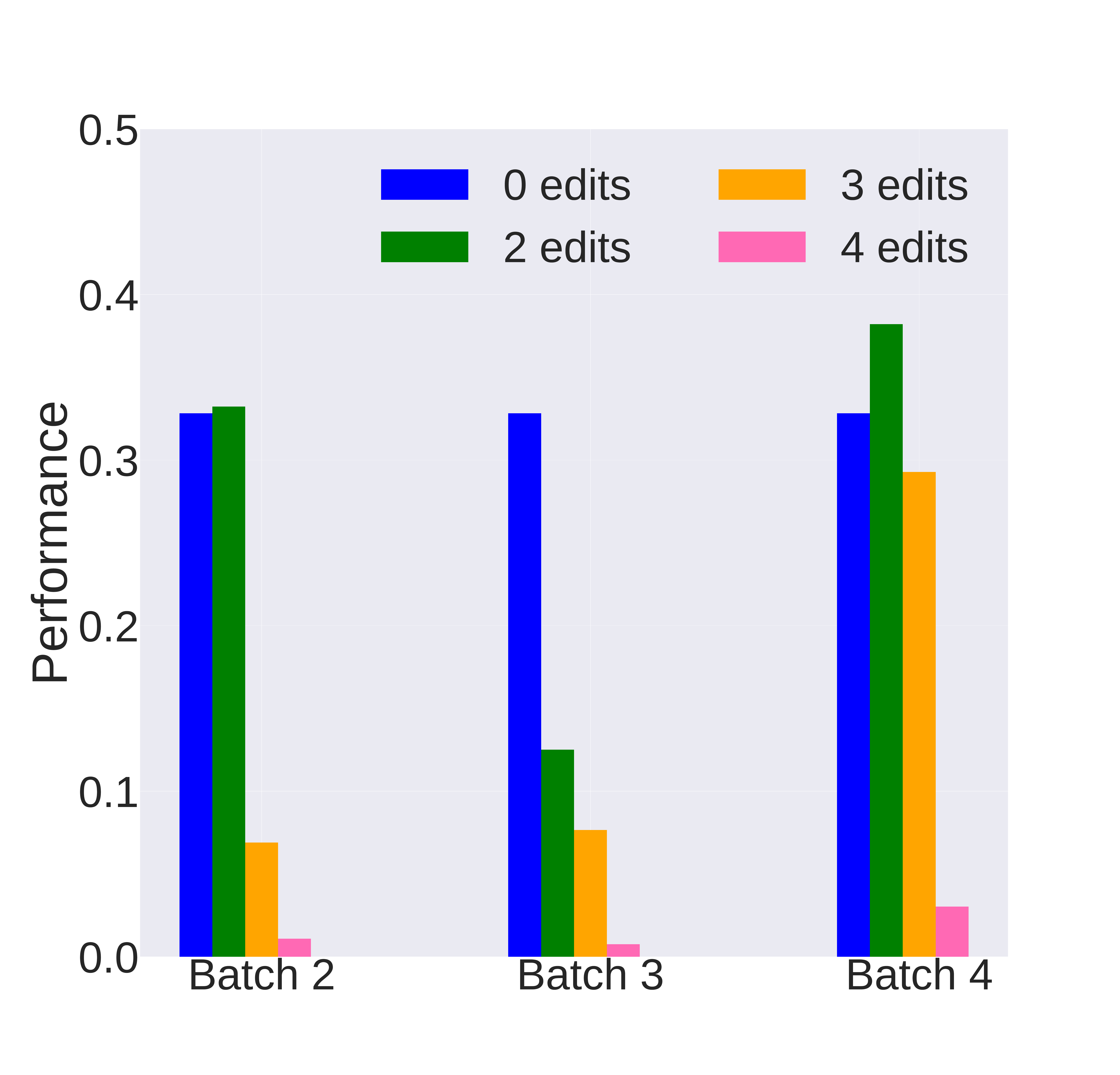}}
  \subfigure[NLI]{
  \includegraphics[width=3.6cm]{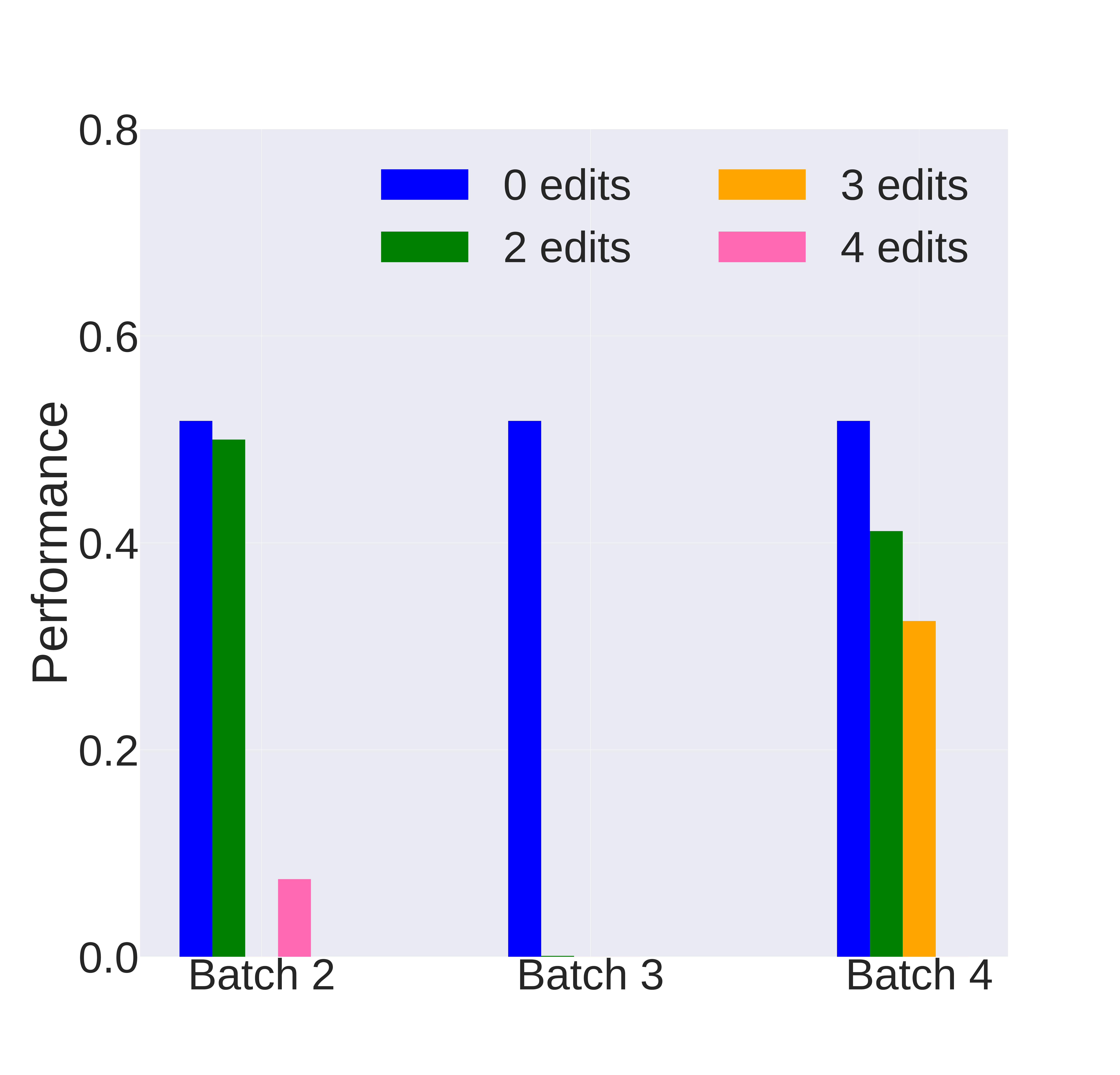}}
  \subfigure[Open-domain QA]{
  \includegraphics[width=3.6cm]{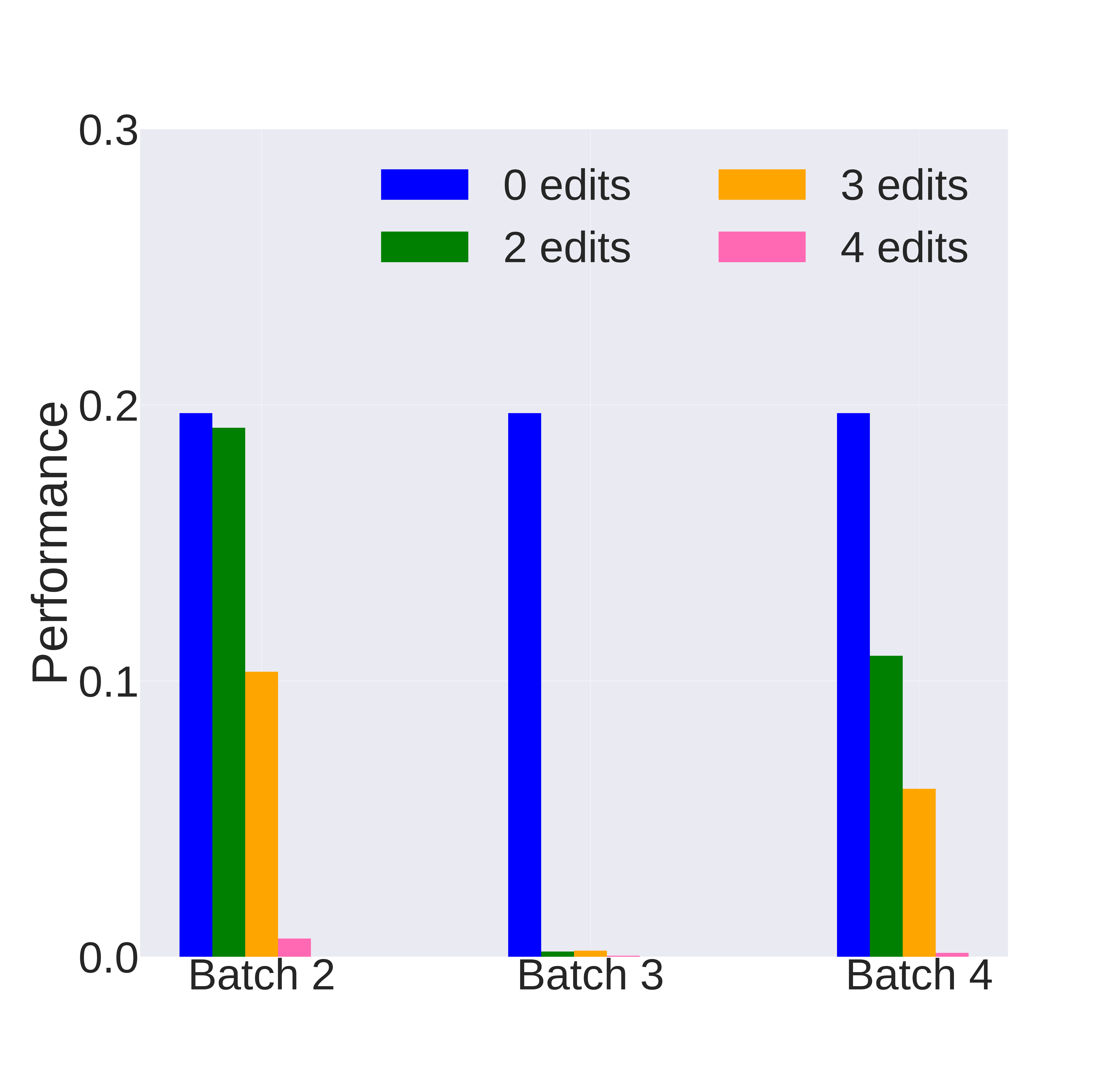}}
  \subfigure[Closed-domain QA]{
  \includegraphics[width=3.6cm]{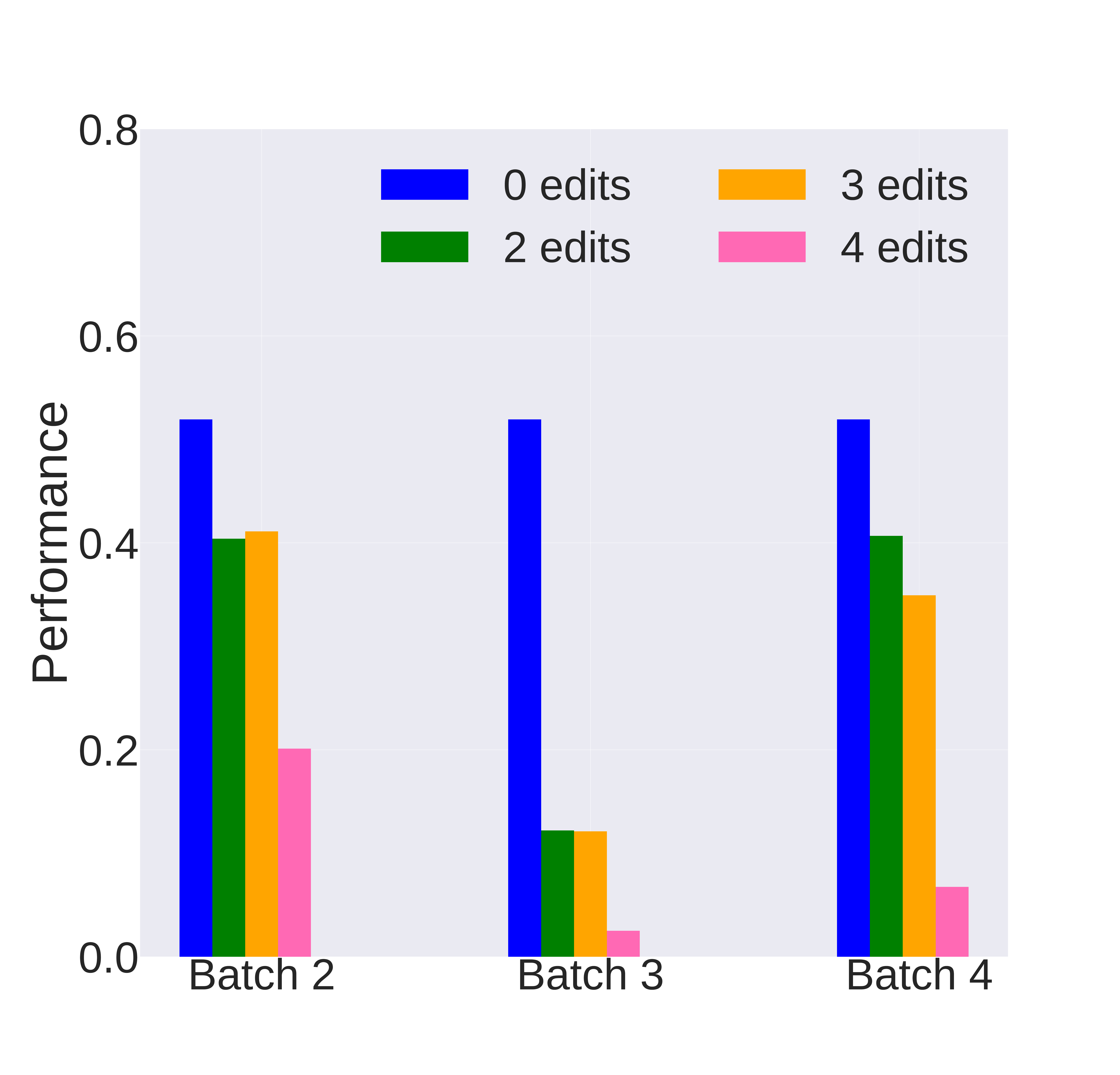}}
  \subfigure[Dialogue]{
  \includegraphics[width=3.6cm]{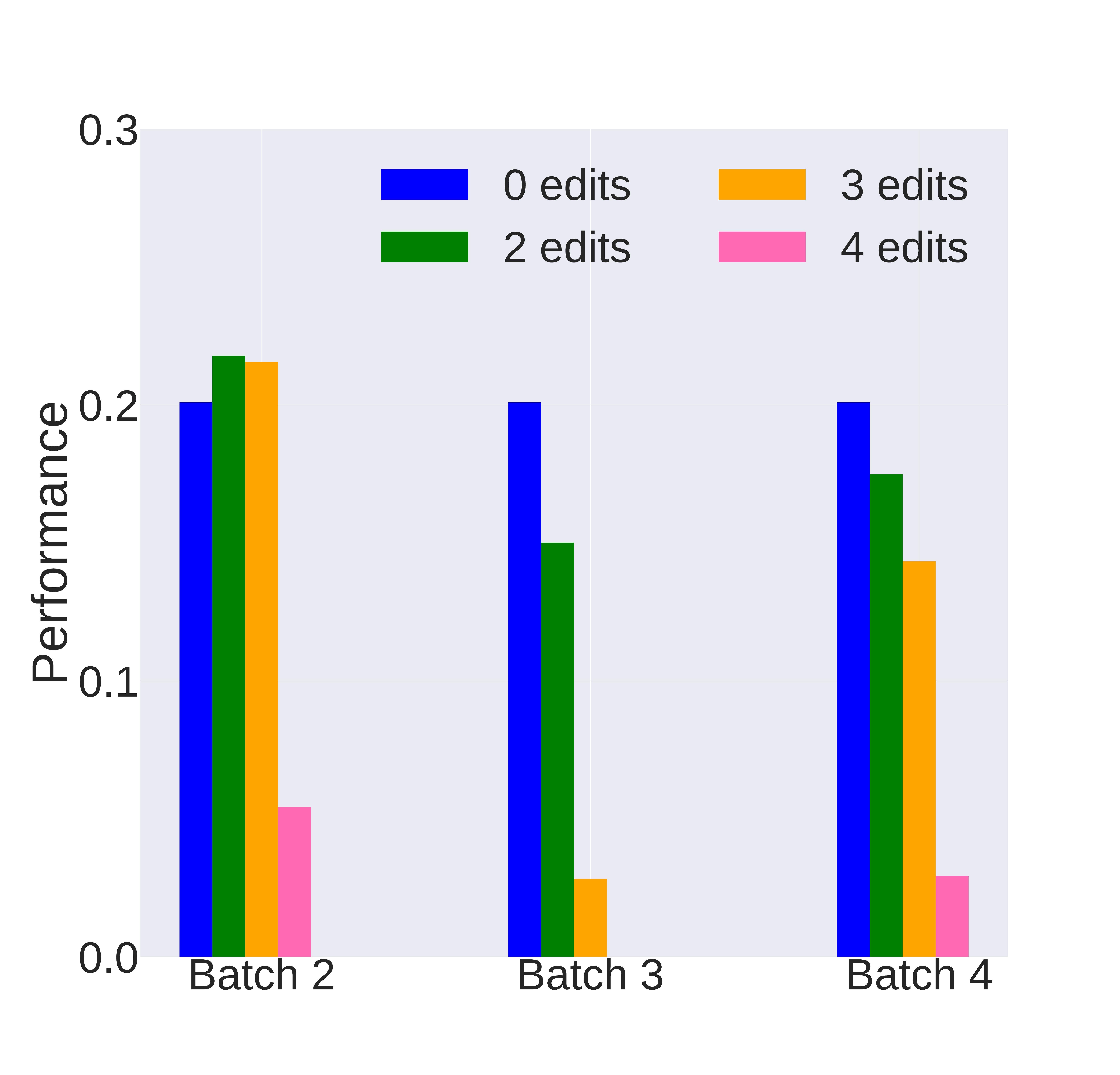}}
  \subfigure[Summarization]{
  \includegraphics[width=3.6cm]{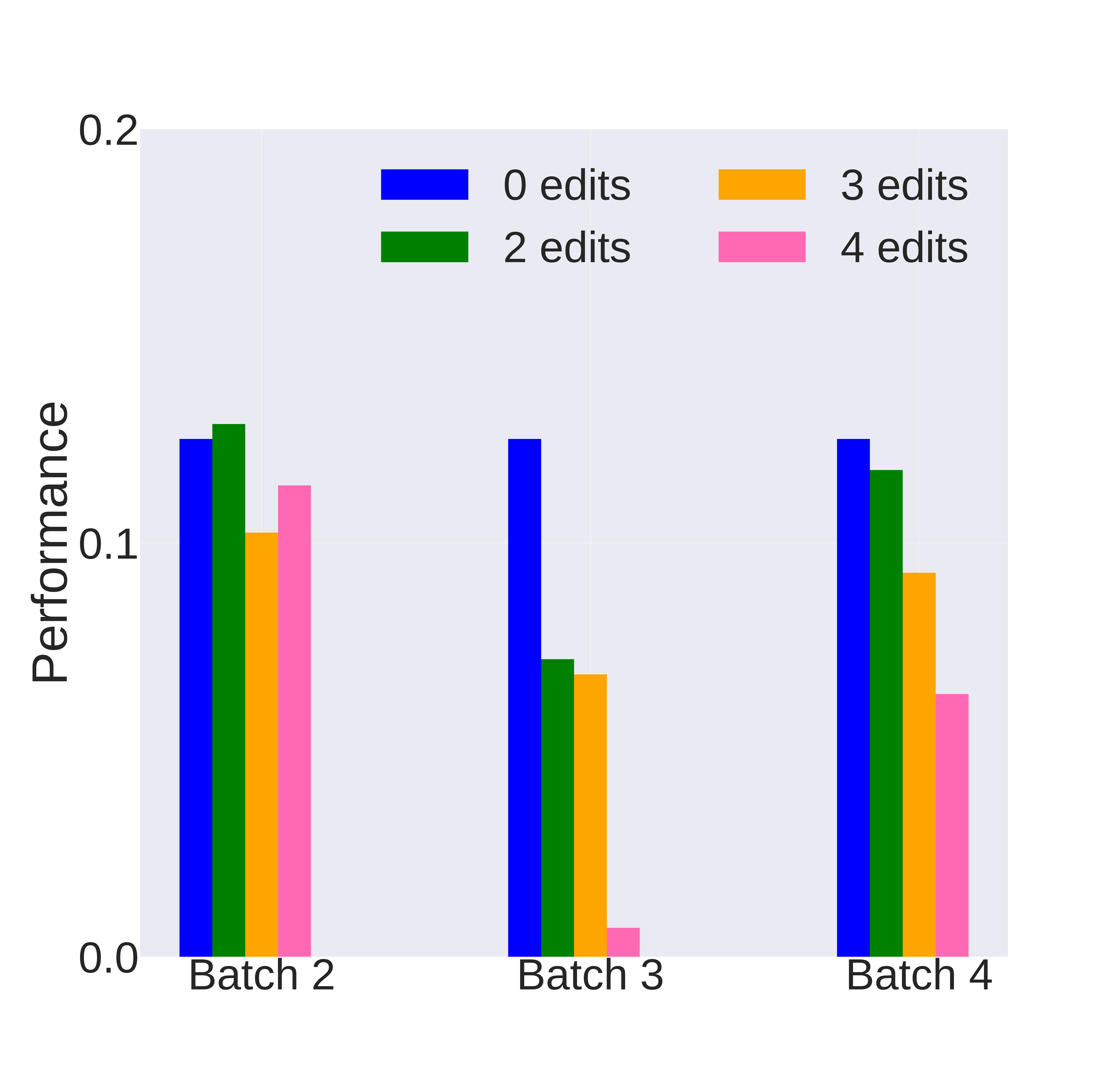}}
  \subfigure[NER]{
  \includegraphics[width=3.6cm]{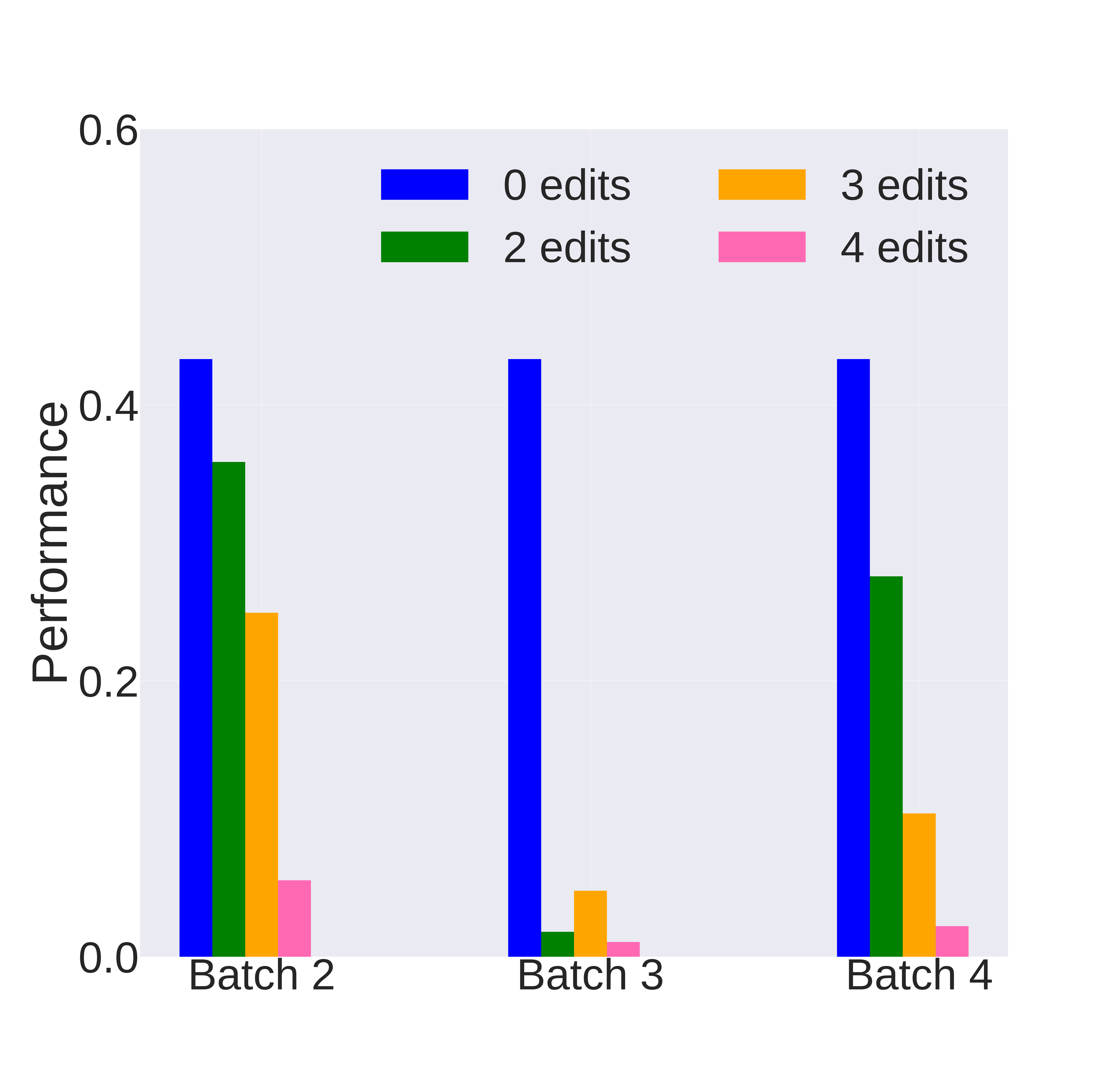}}
  \subfigure[Sentiment analysis]{
  \includegraphics[width=3.6cm]{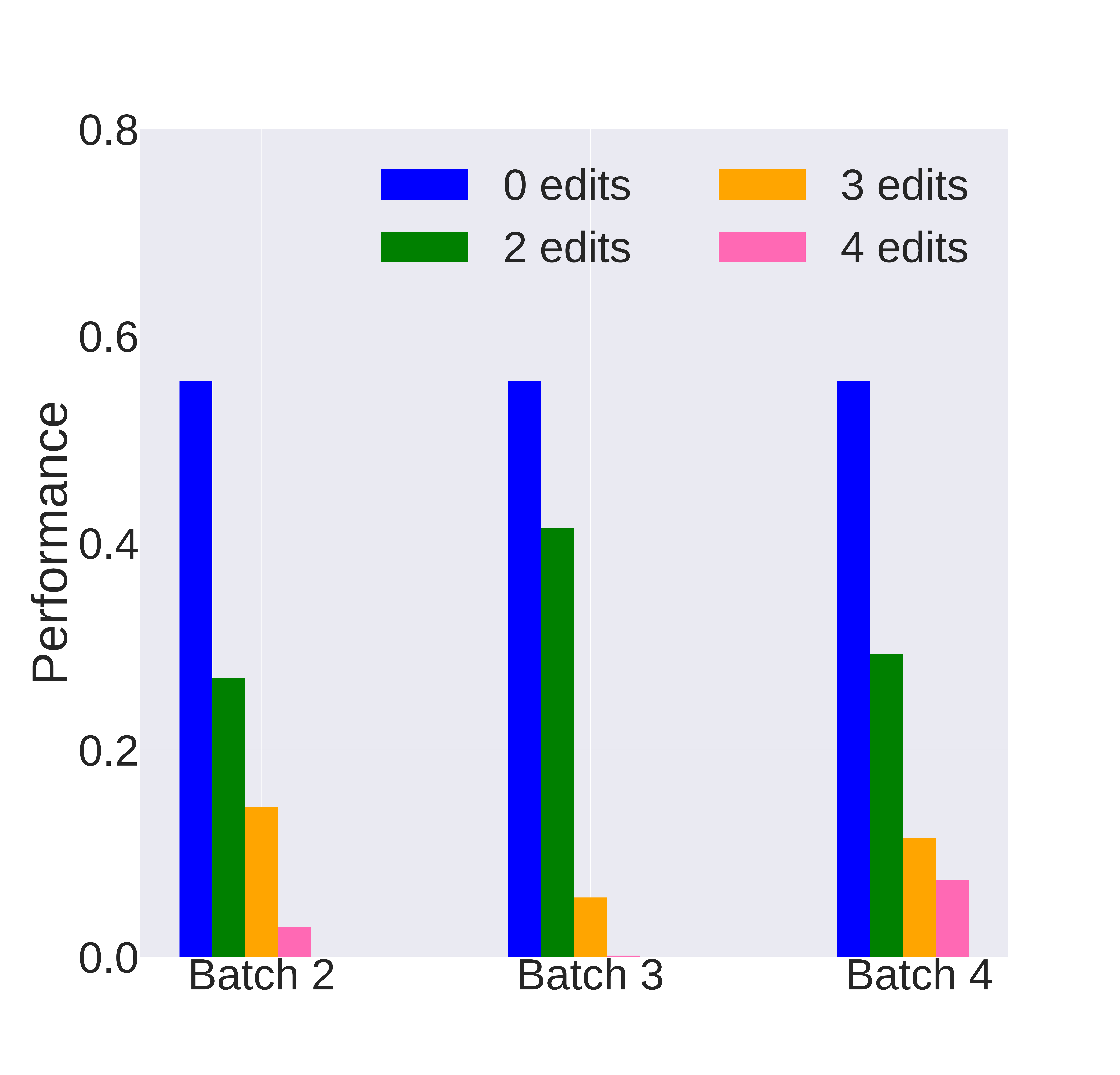}}
  \vspace{-4mm}
  \caption{Performance on general tasks of edited models using MEND to edit GPT2-XL as the number of edits increases in \emph{batch- and sequential-editing}.}
  \vspace{-4mm}
  \label{fig-batch-sequential-mend-gpt2xl}
\end{figure*}

%%%%%%%%%%%%%%%%%%%%%%%%%%%%%%%%%%%%%%%%%%%%%%%%%%%%%%%%%%%%%%%%%%%%%%%%%%%%

\begin{figure*}[!hbt]
  \centering
  \subfigure[Reasoning]{
  \includegraphics[width=3.6cm]{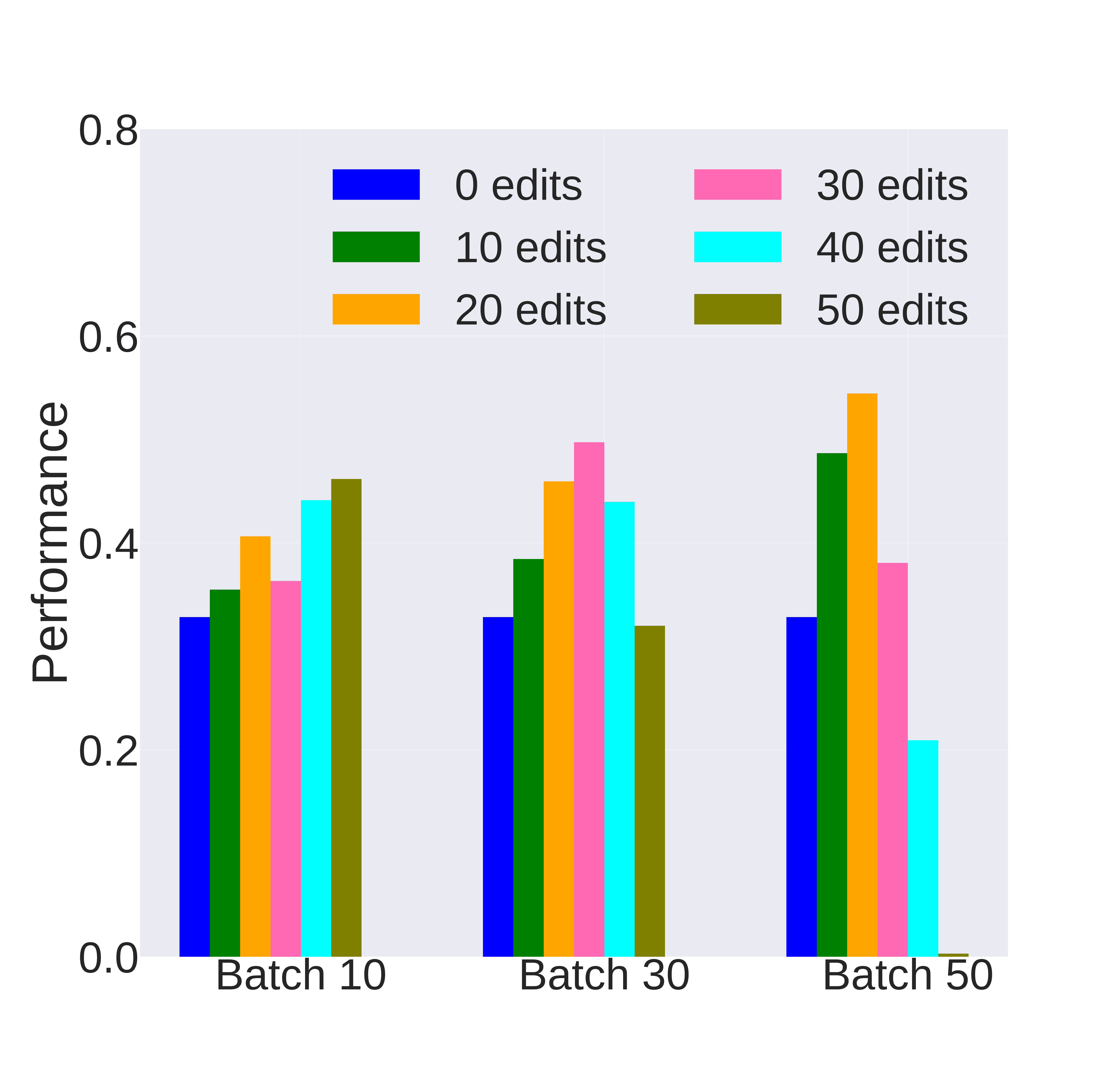}}
  \subfigure[NLI]{
  \includegraphics[width=3.6cm]{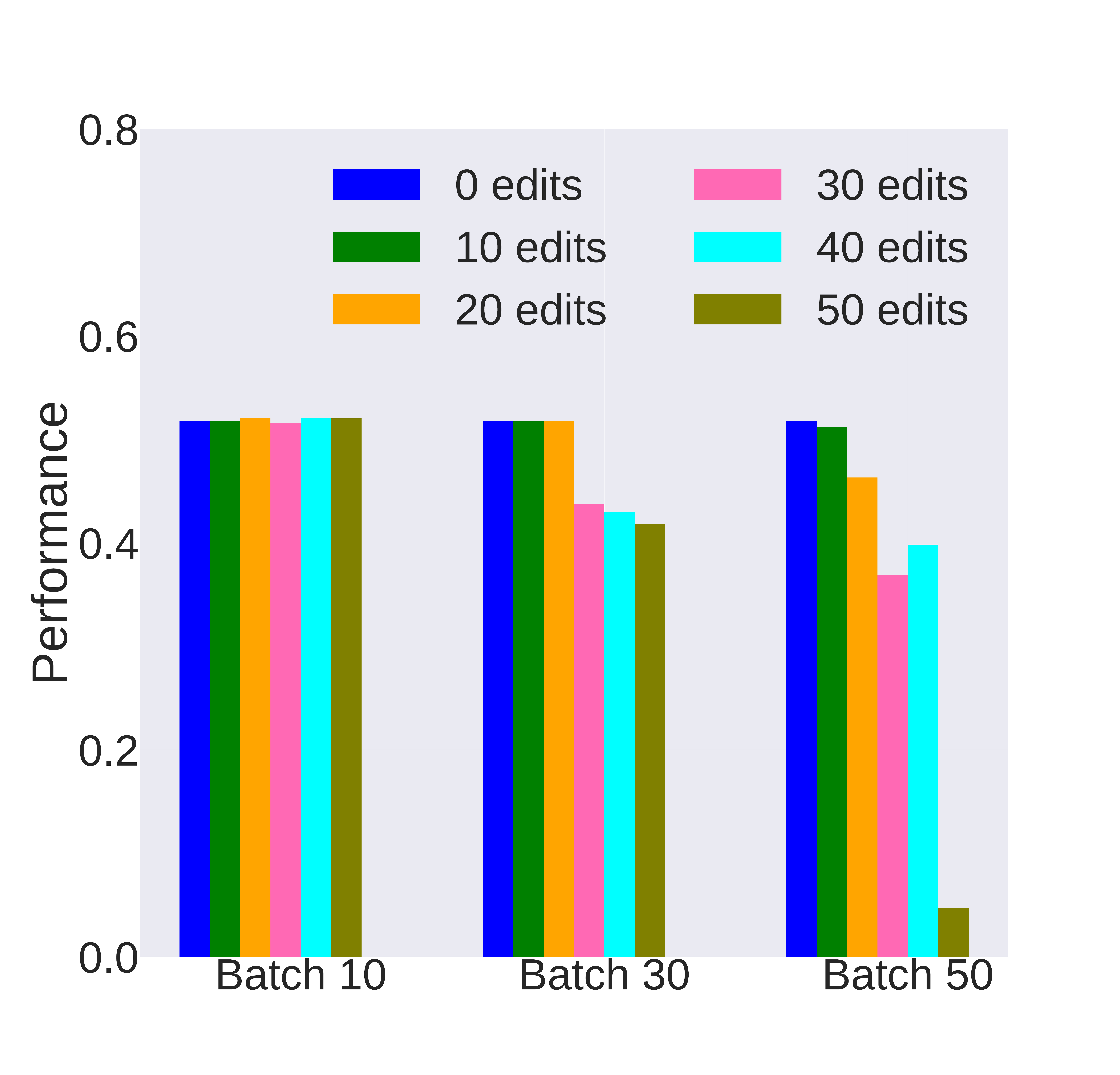}}
  \subfigure[Open-domain QA]{
  \includegraphics[width=3.6cm]{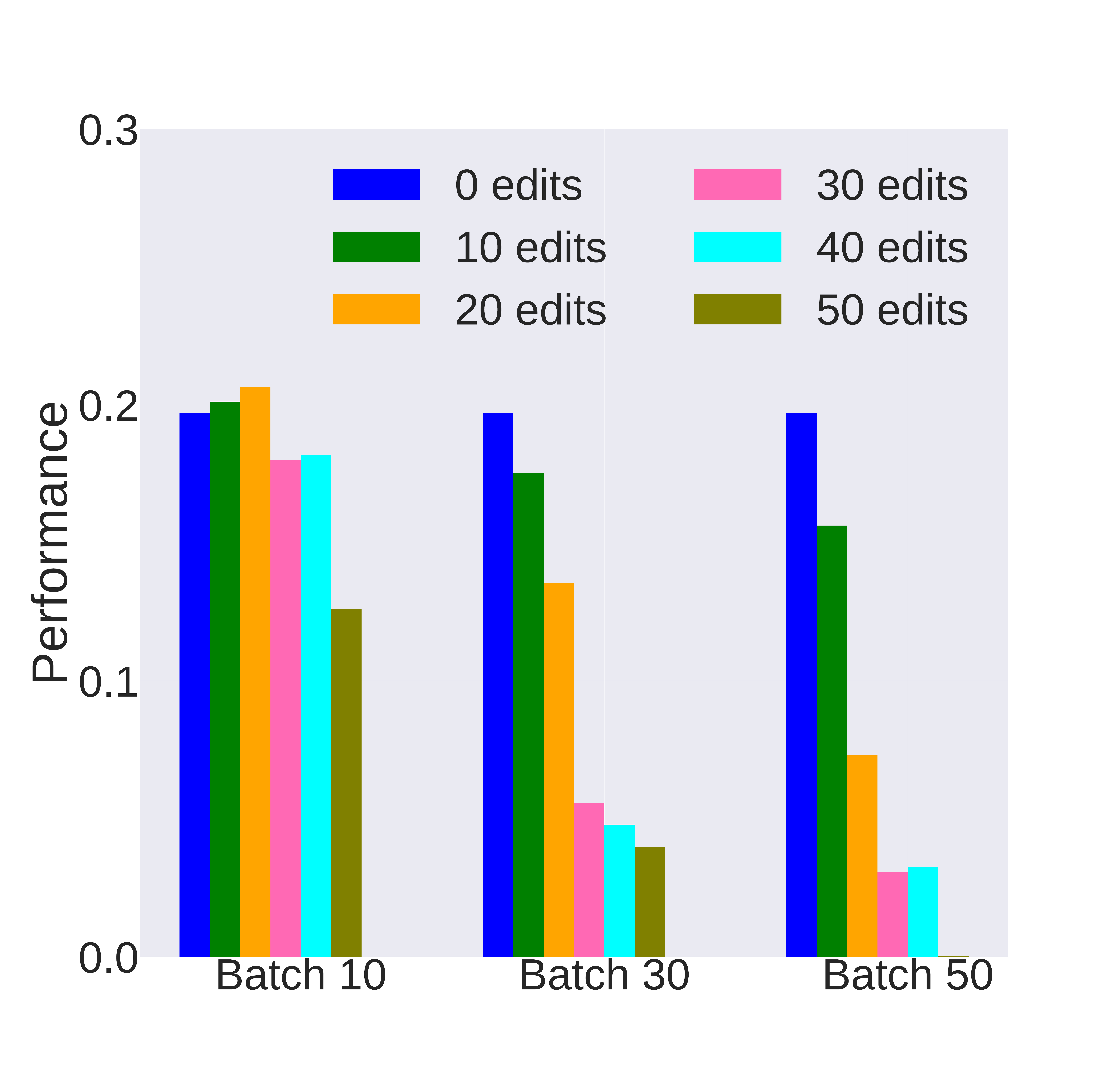}}
  \subfigure[Closed-domain QA]{
  \includegraphics[width=3.6cm]{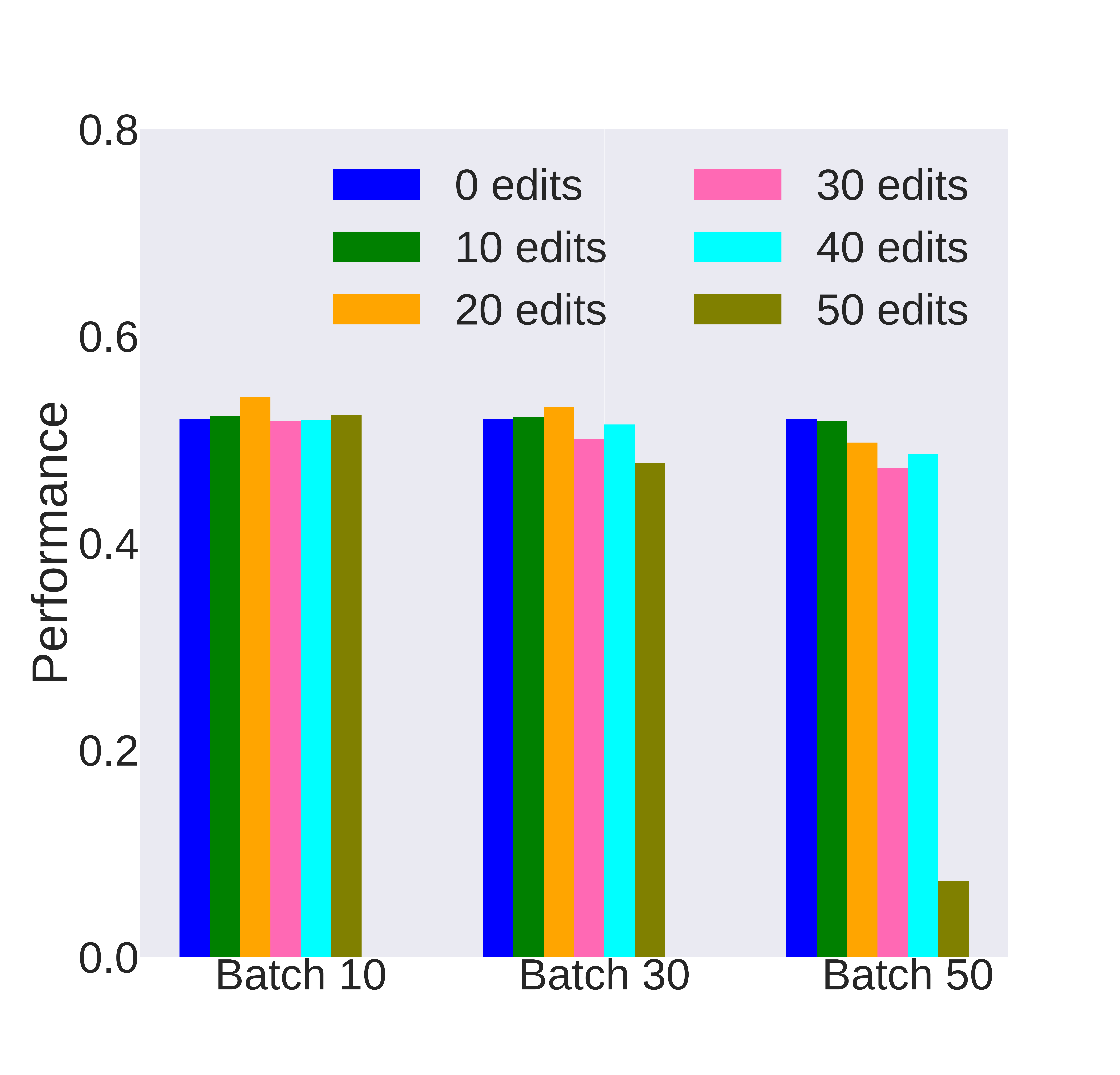}}
  \subfigure[Dialogue]{
  \includegraphics[width=3.6cm]{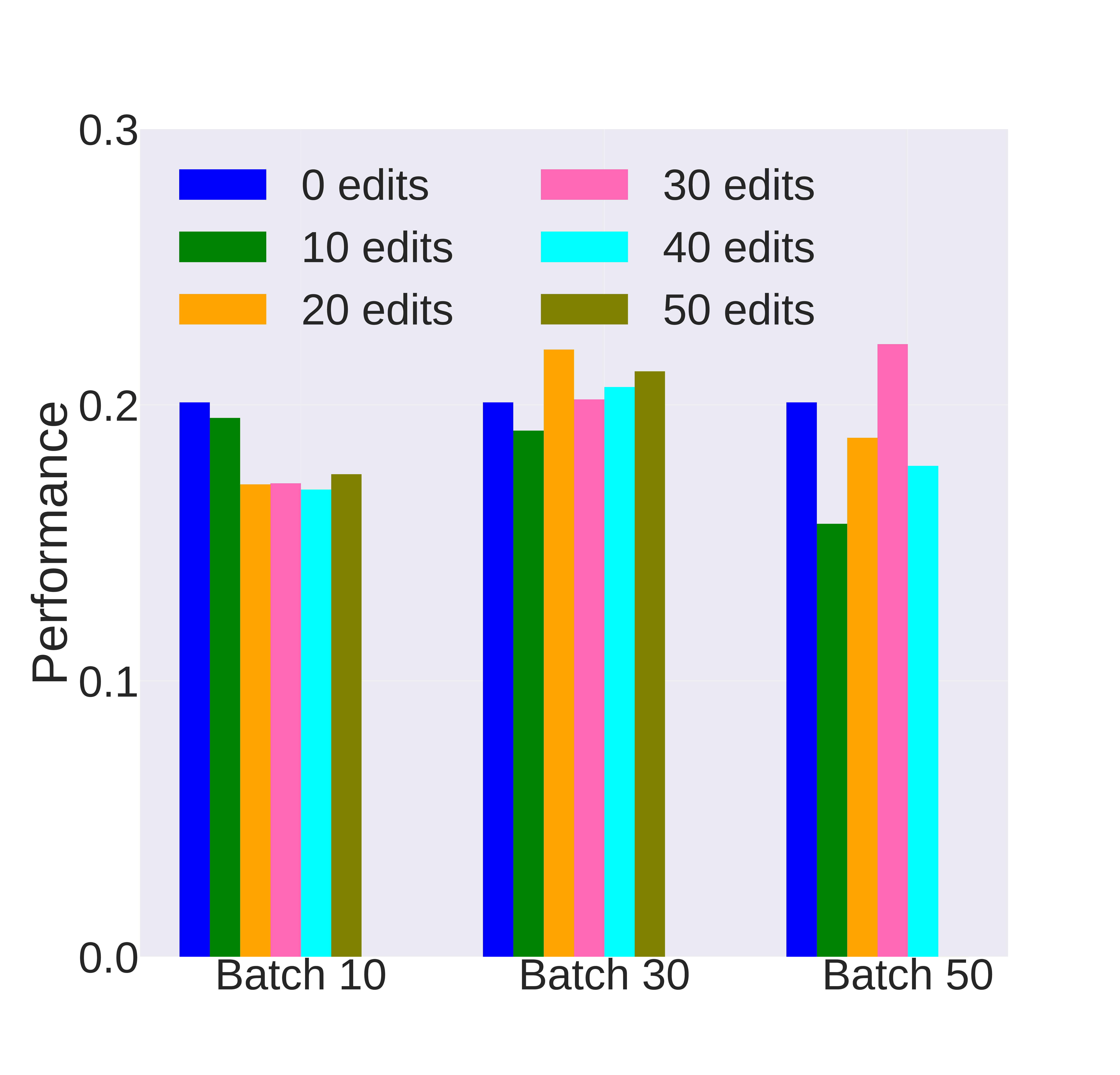}}
  \subfigure[Summarization]{
  \includegraphics[width=3.6cm]{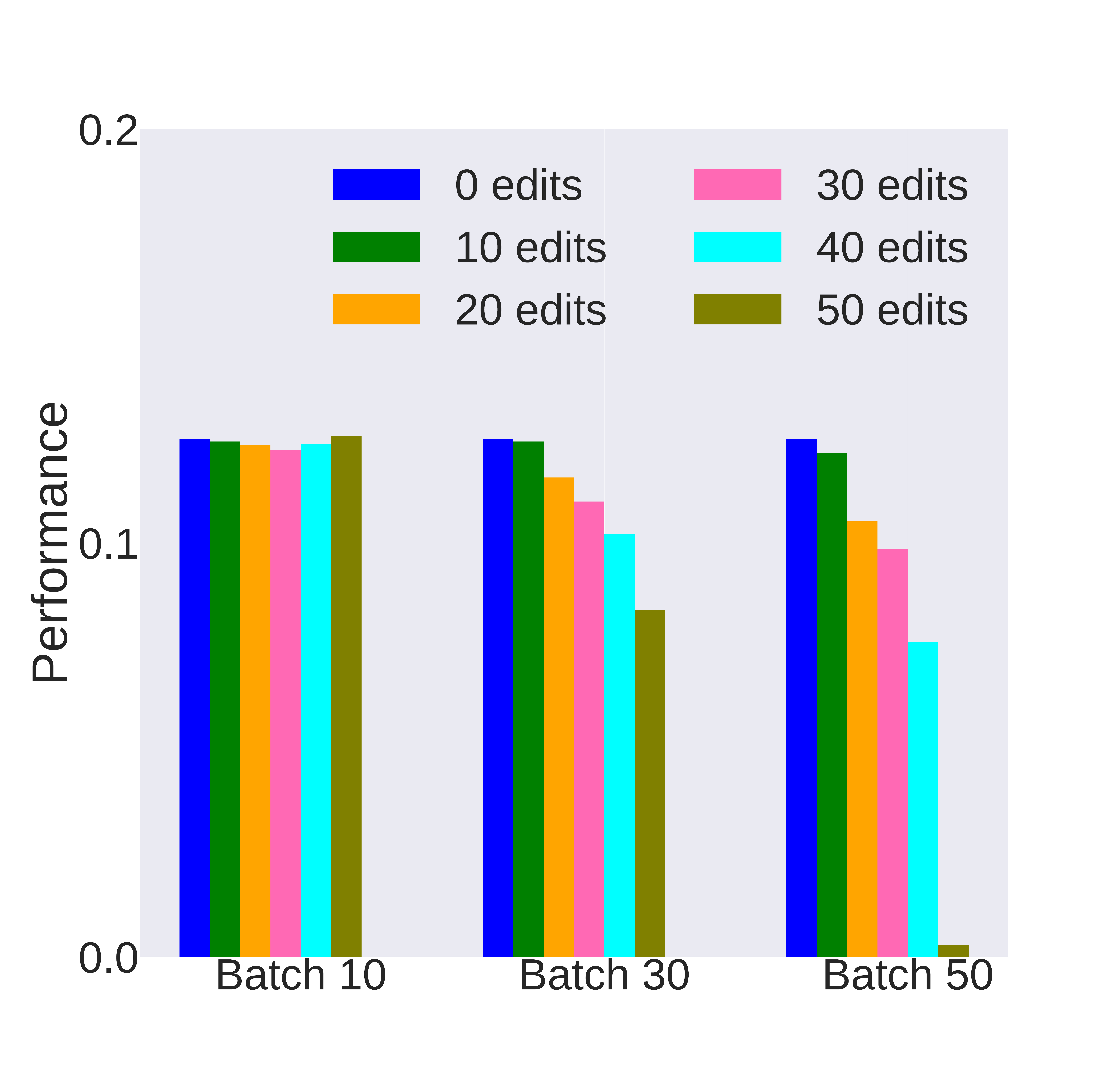}}
  \subfigure[NER]{
  \includegraphics[width=3.6cm]{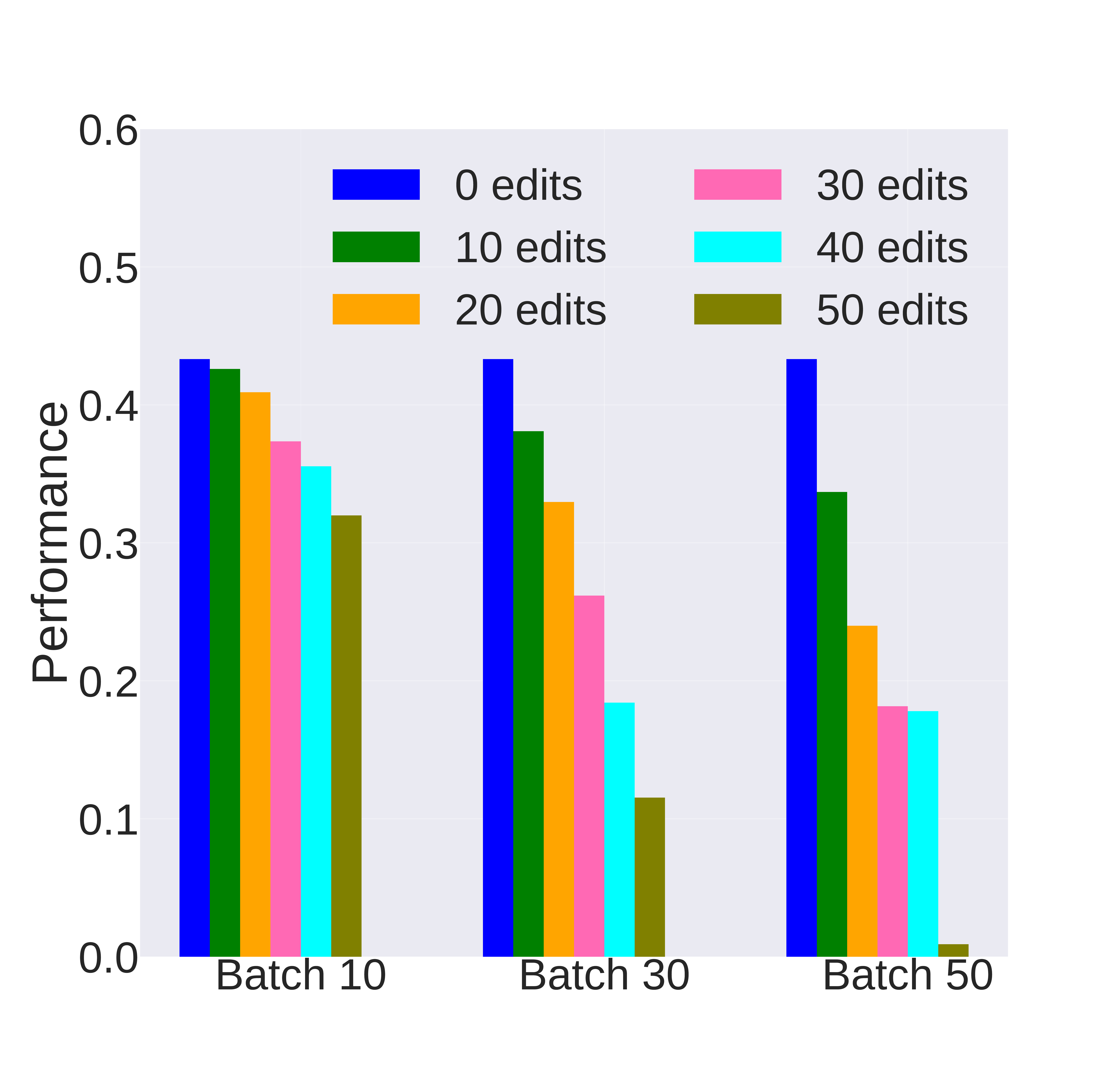}}
  \subfigure[Sentiment analysis]{
  \includegraphics[width=3.6cm]{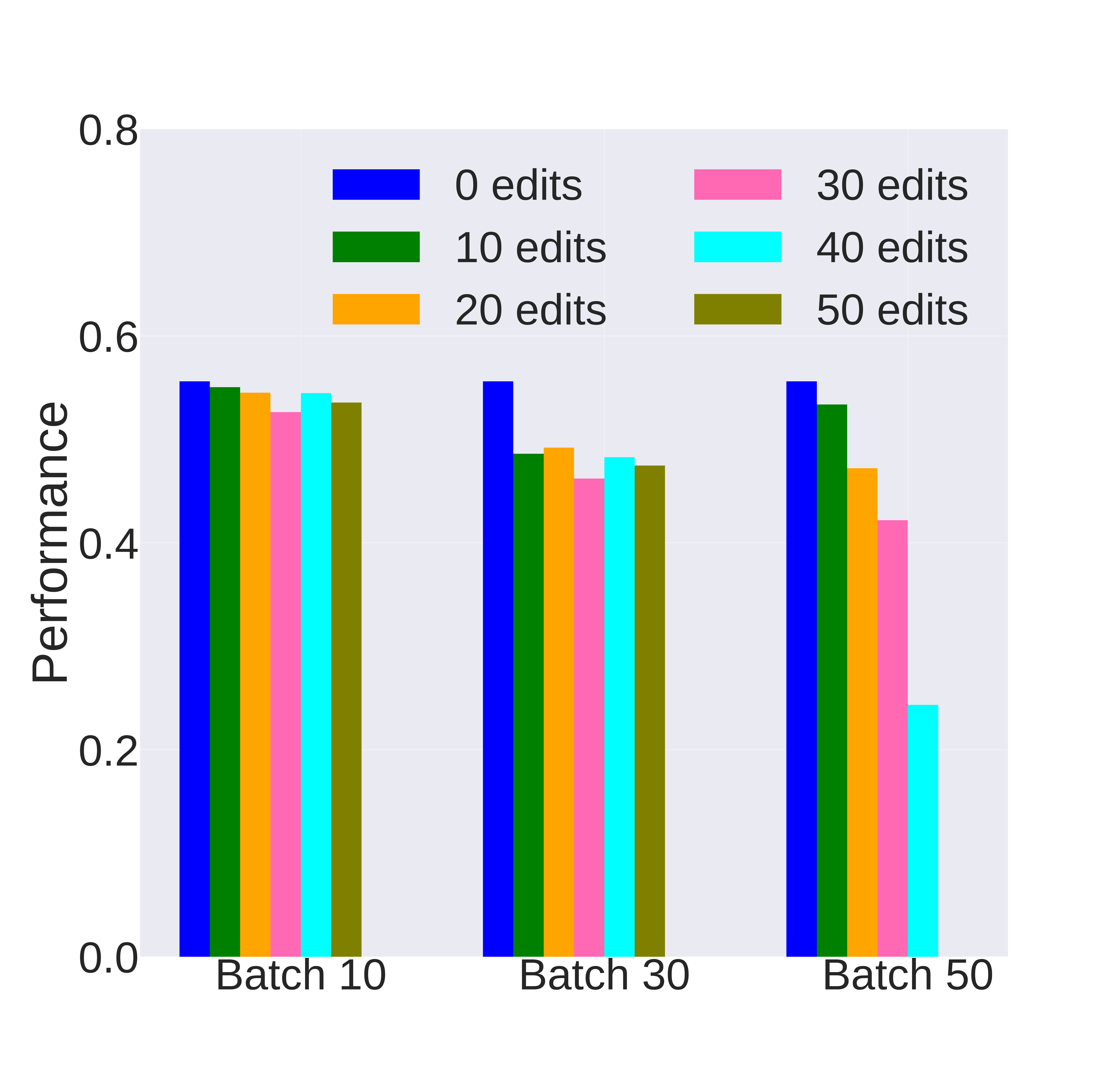}}
  \vspace{-4mm}
  \caption{Performance on general tasks of edited models using MEMIT to edit GPT2-XL as the number of edits increases in \emph{batch- and sequential-editing}.}
  \vspace{-4mm}
  \label{fig-batch-sequential-memit-gpt2xl}
\end{figure*}

\clearpage

%%%%%%%%%%%%%%%%%%%%%%%%%%%%%%%%%%%%%%%%%%%%%%%%%%%%%%%%%%%%%%%%%%%%%%%%%%%%

\begin{figure*}[!hbt]
  \centering
  \subfigure[Reasoning]{
  \includegraphics[width=3.6cm]{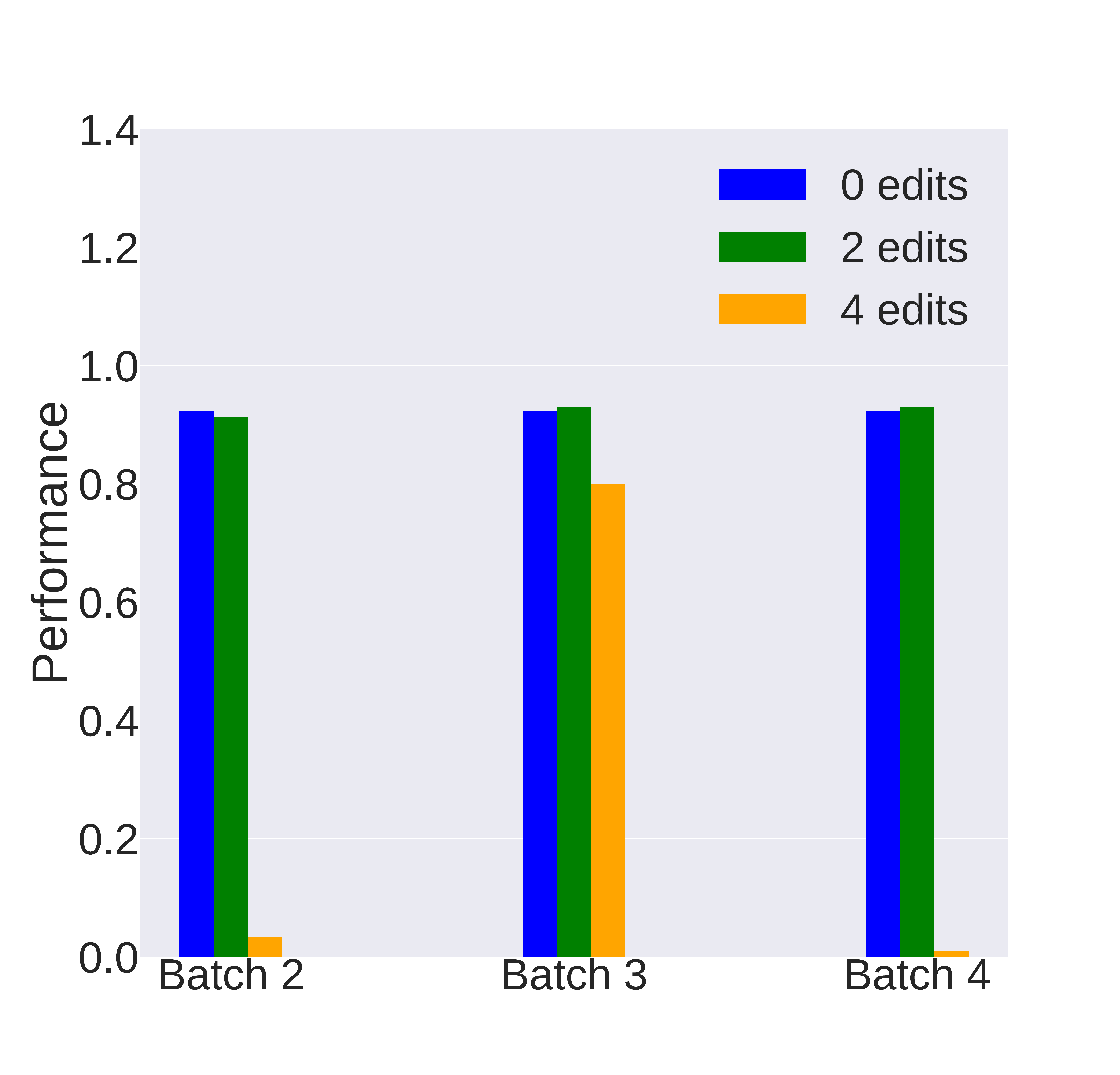}}
  \subfigure[NLI]{
  \includegraphics[width=3.6cm]{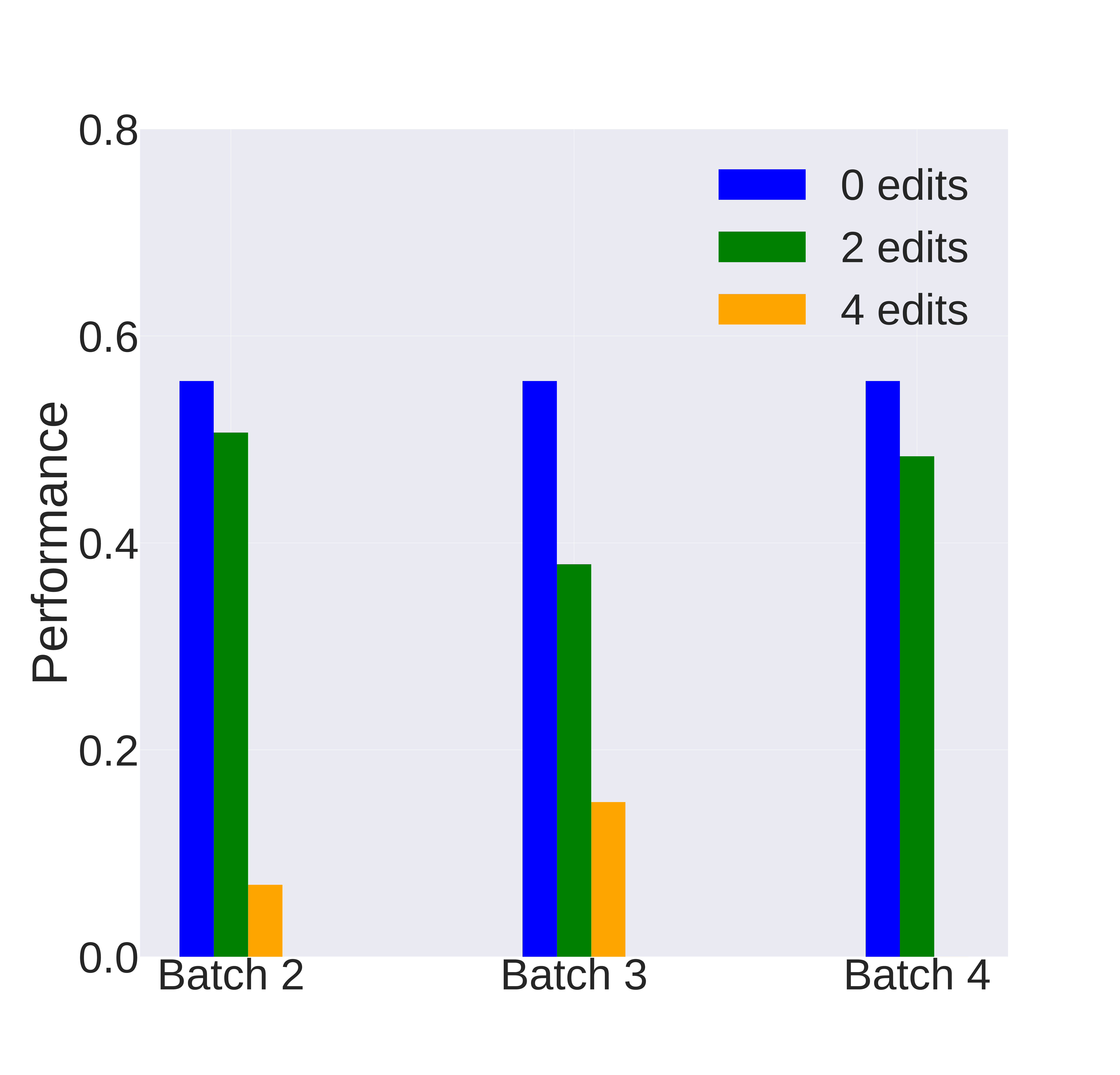}}
  \subfigure[Open-domain QA]{
  \includegraphics[width=3.6cm]{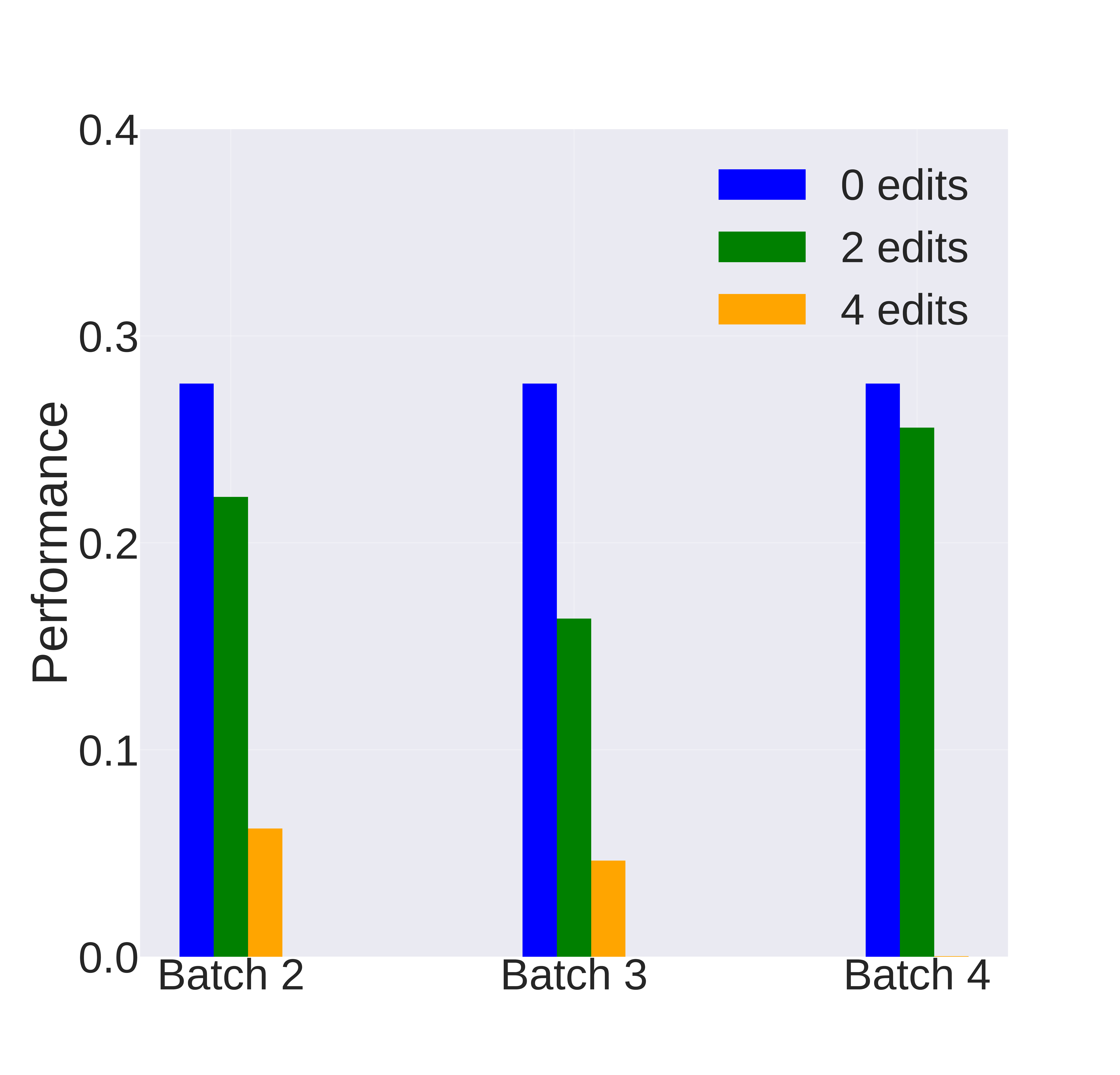}}
  \subfigure[Closed-domain QA]{
  \includegraphics[width=3.6cm]{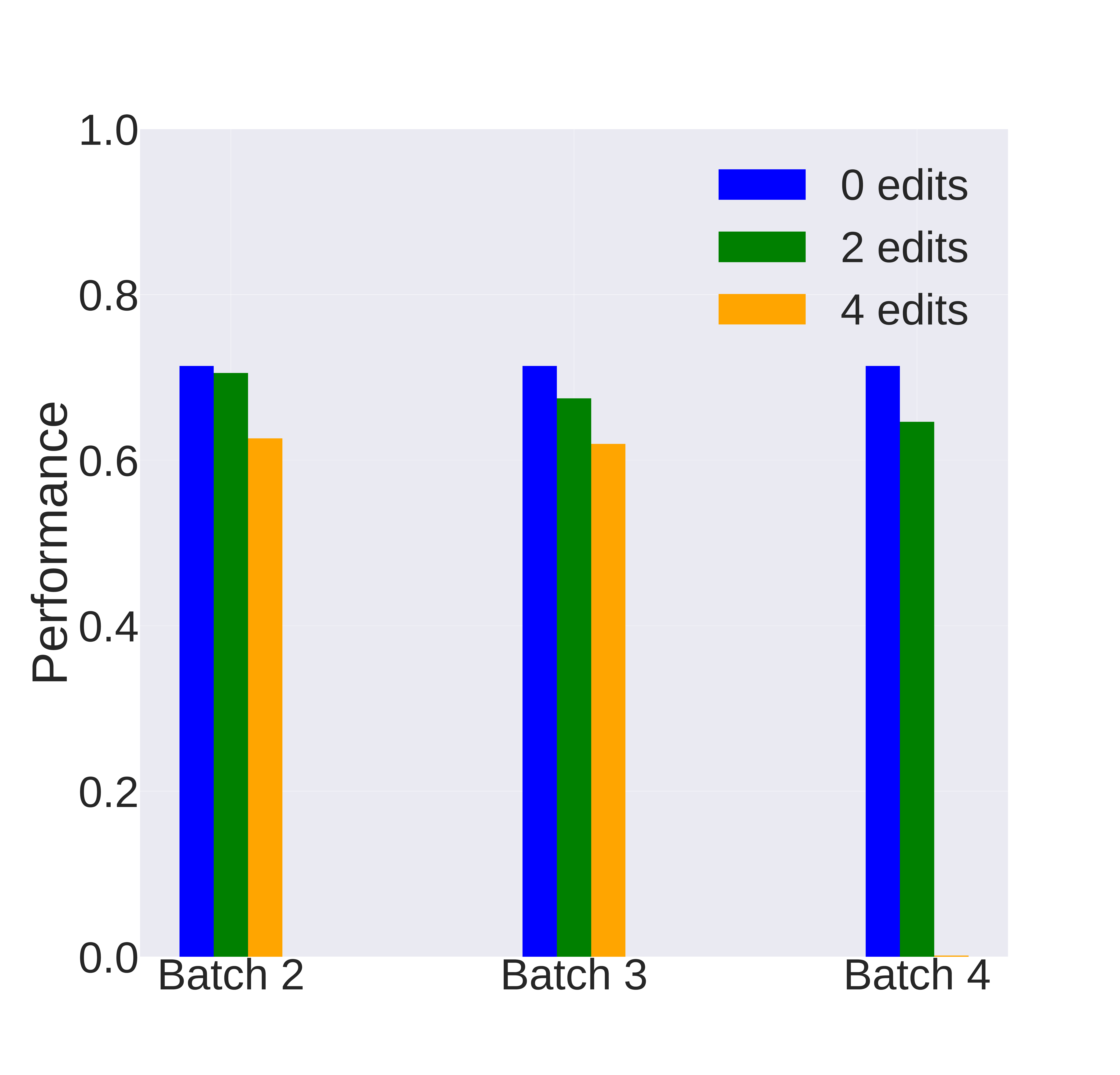}}
  \subfigure[Dialogue]{
  \includegraphics[width=3.6cm]{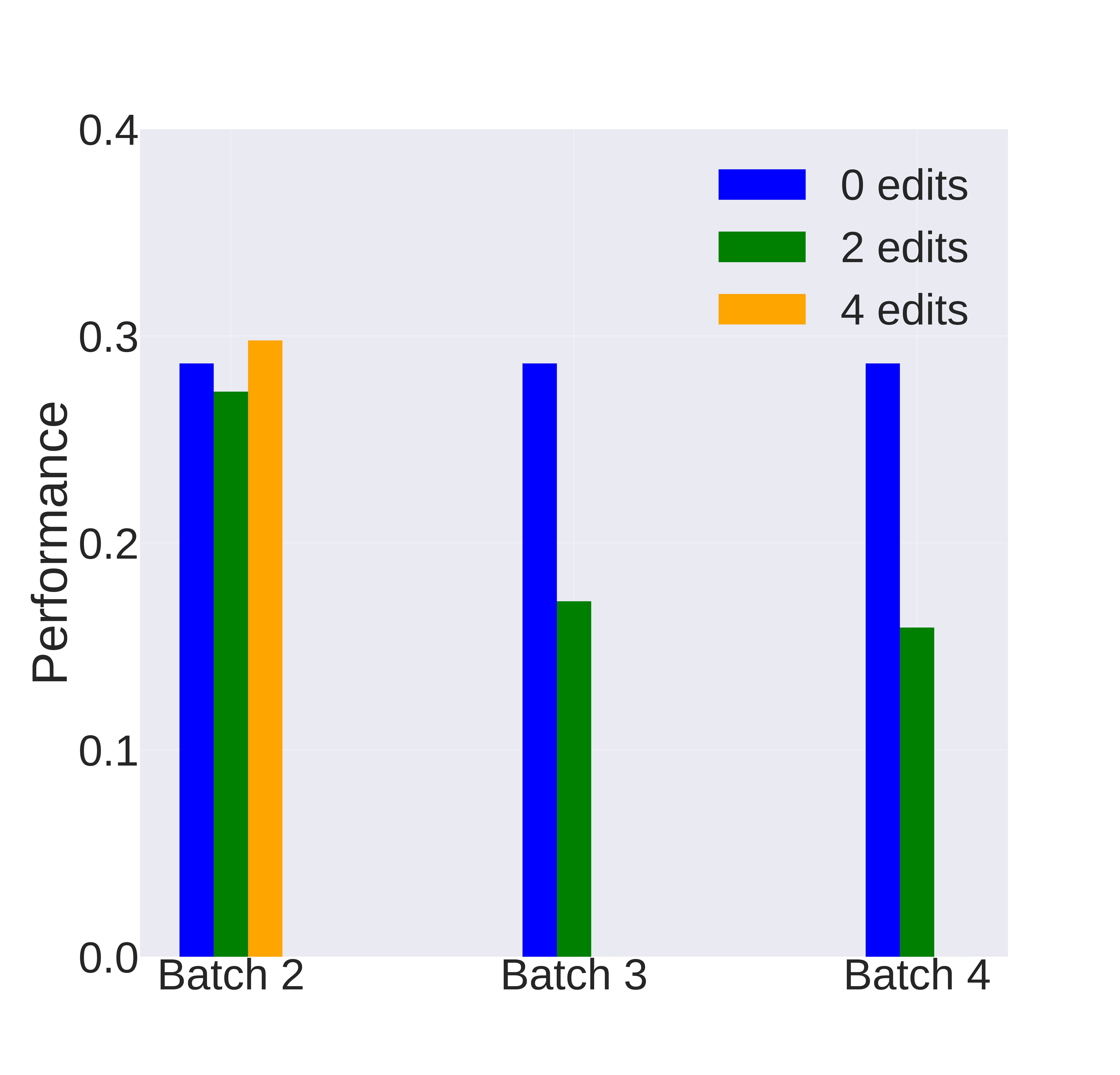}}
  \subfigure[Summarization]{
  \includegraphics[width=3.6cm]{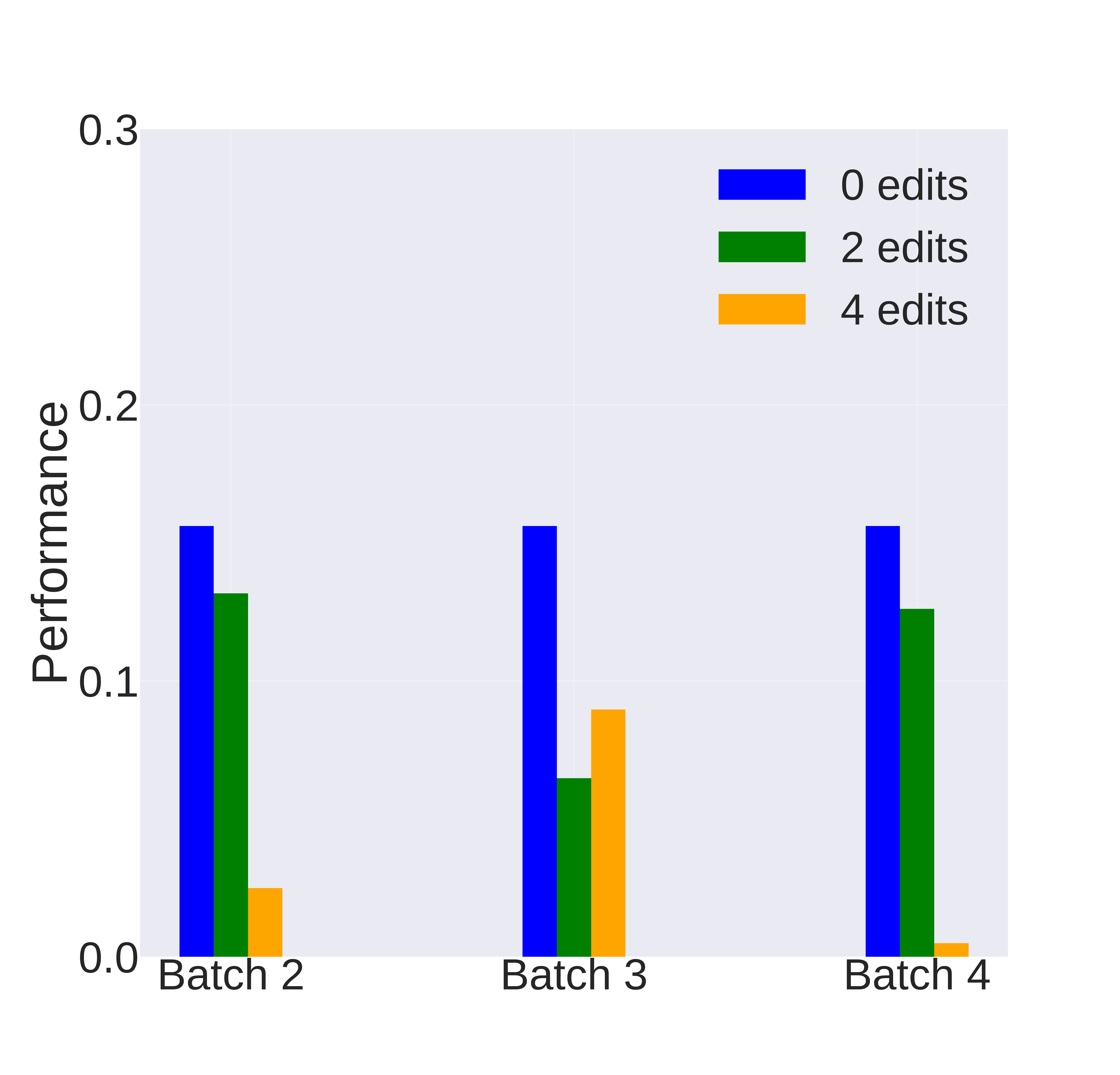}}
  \subfigure[NER]{
  \includegraphics[width=3.6cm]{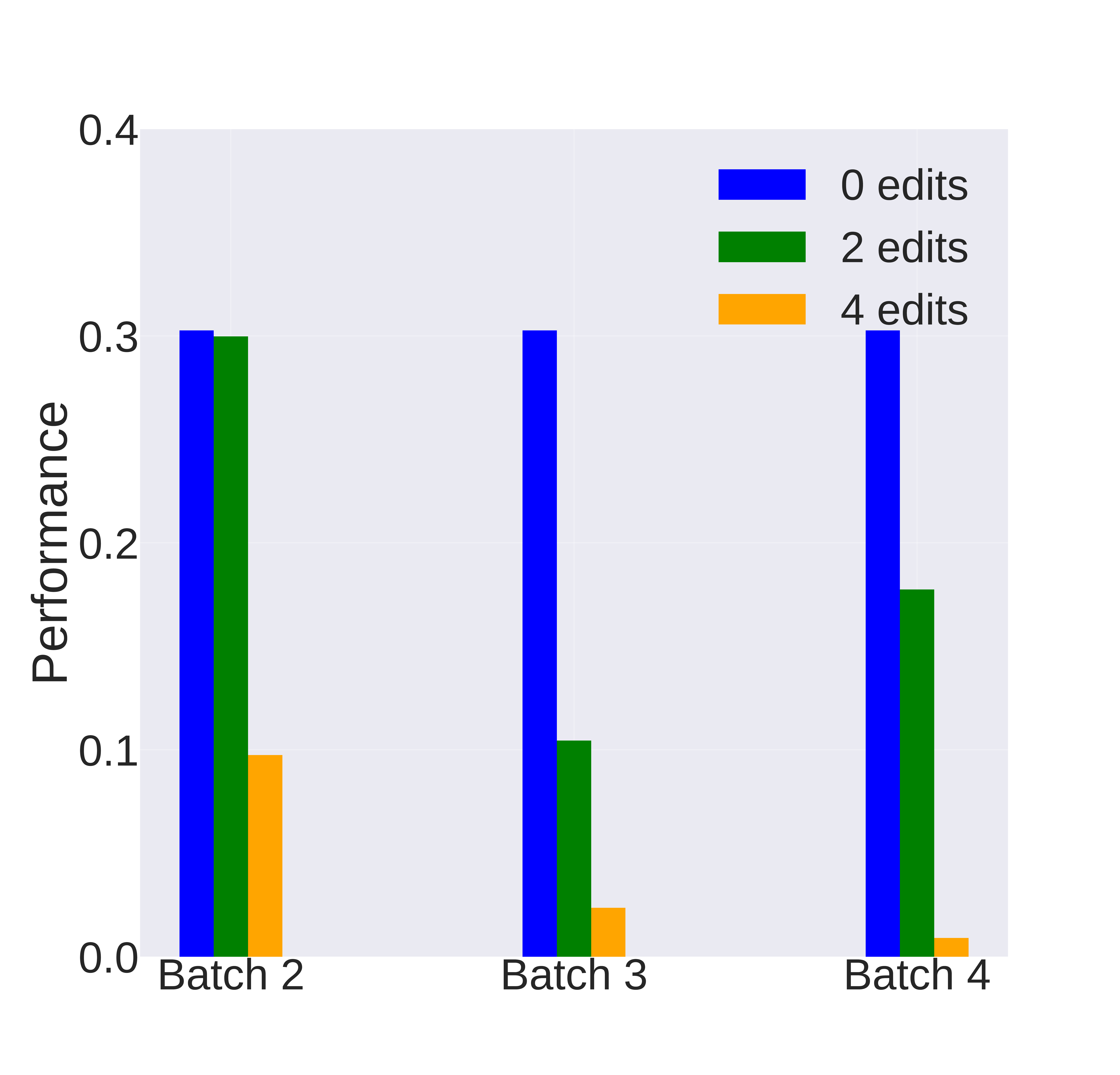}}
  \subfigure[Sentiment analysis]{
  \includegraphics[width=3.6cm]{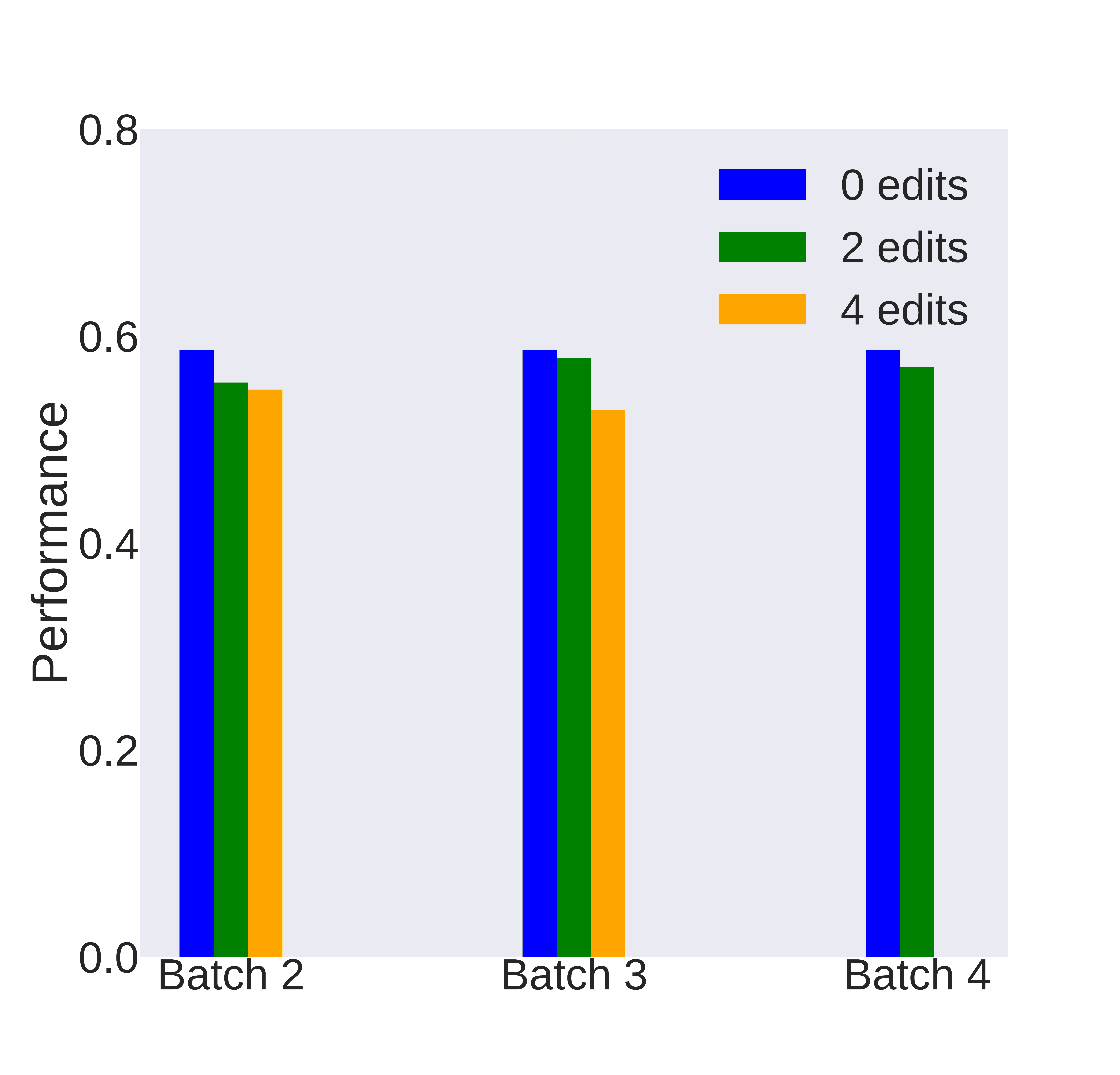}}
  \vspace{-4mm}
  \caption{Performance on general tasks of edited models using MEND to edit LLaMA-1 (7B) as the number of edits increases in \emph{batch- and sequential-editing}.}
  \vspace{-4mm}
  \label{fig-batch-sequential-mend-llama1-7b}
\end{figure*}

%%%%%%%%%%%%%%%%%%%%%%%%%%%%%%%%%%%%%%%%%%%%%%%%%%%%%%%%%%%%%%%%%%%%%%%%%%%%

\begin{figure*}[!hbt]
  \centering
  \subfigure[Reasoning]{
  \includegraphics[width=3.6cm]{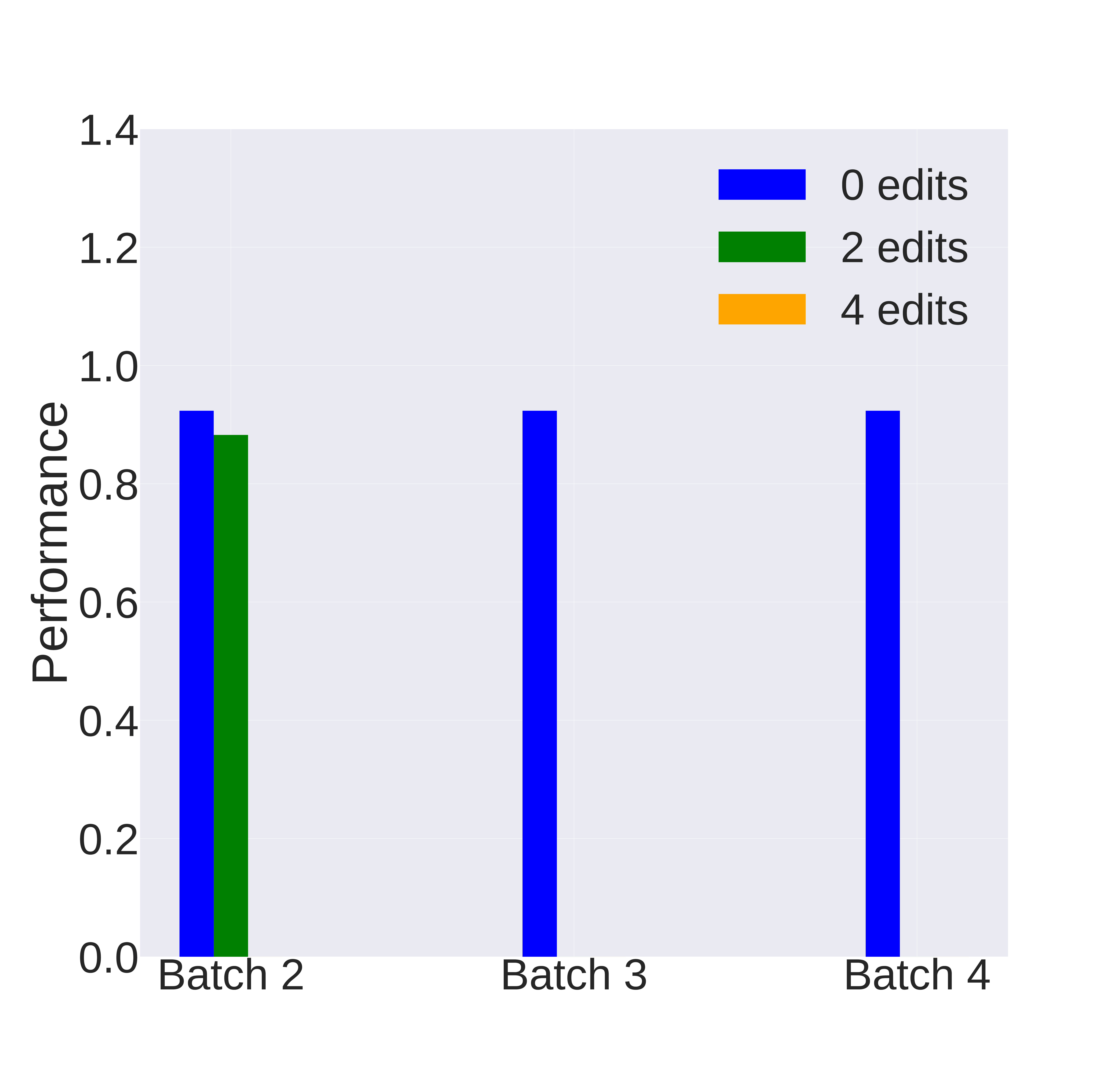}}
  \subfigure[NLI]{
  \includegraphics[width=3.6cm]{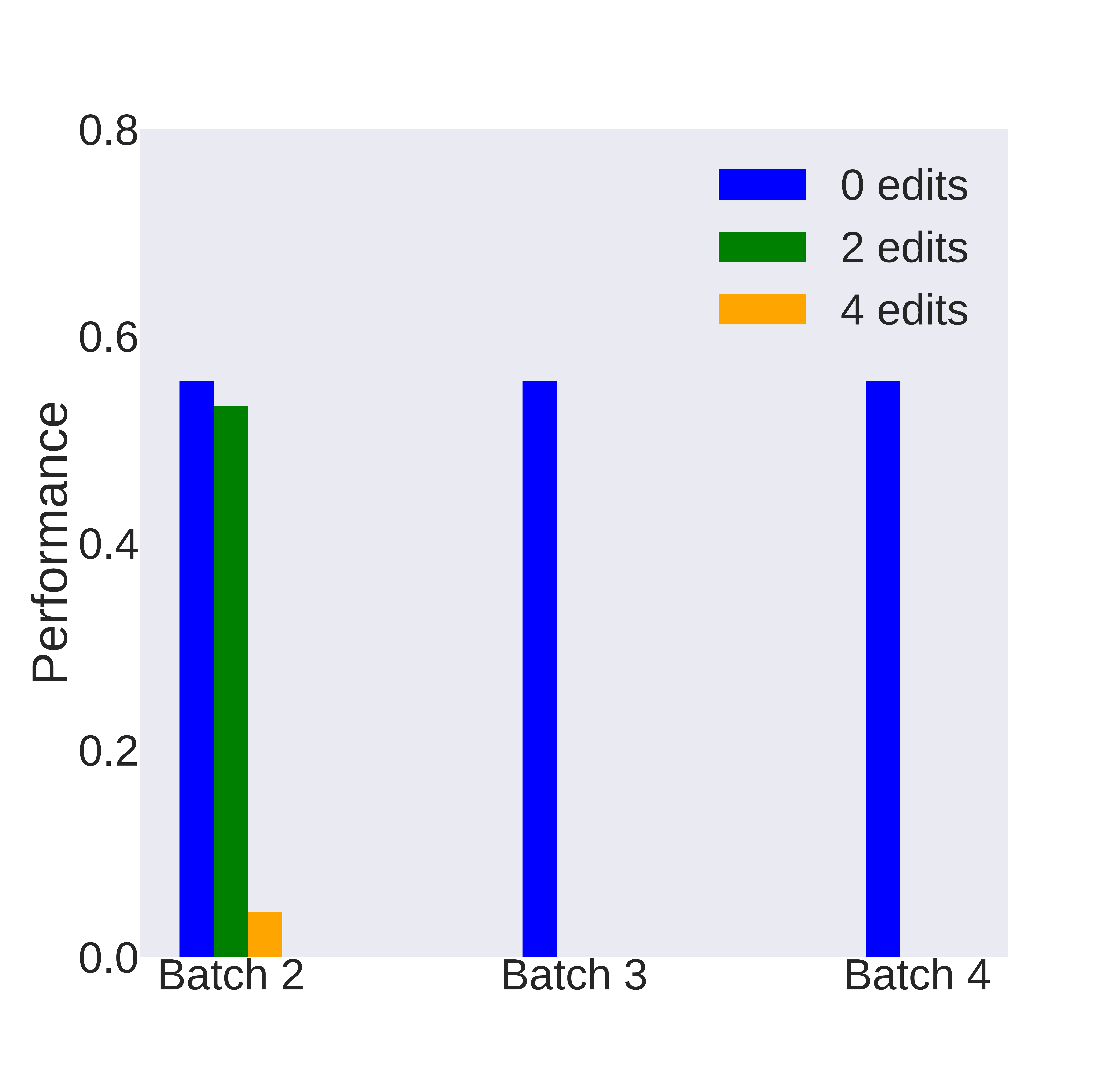}}
  \subfigure[Open-domain QA]{
  \includegraphics[width=3.6cm]{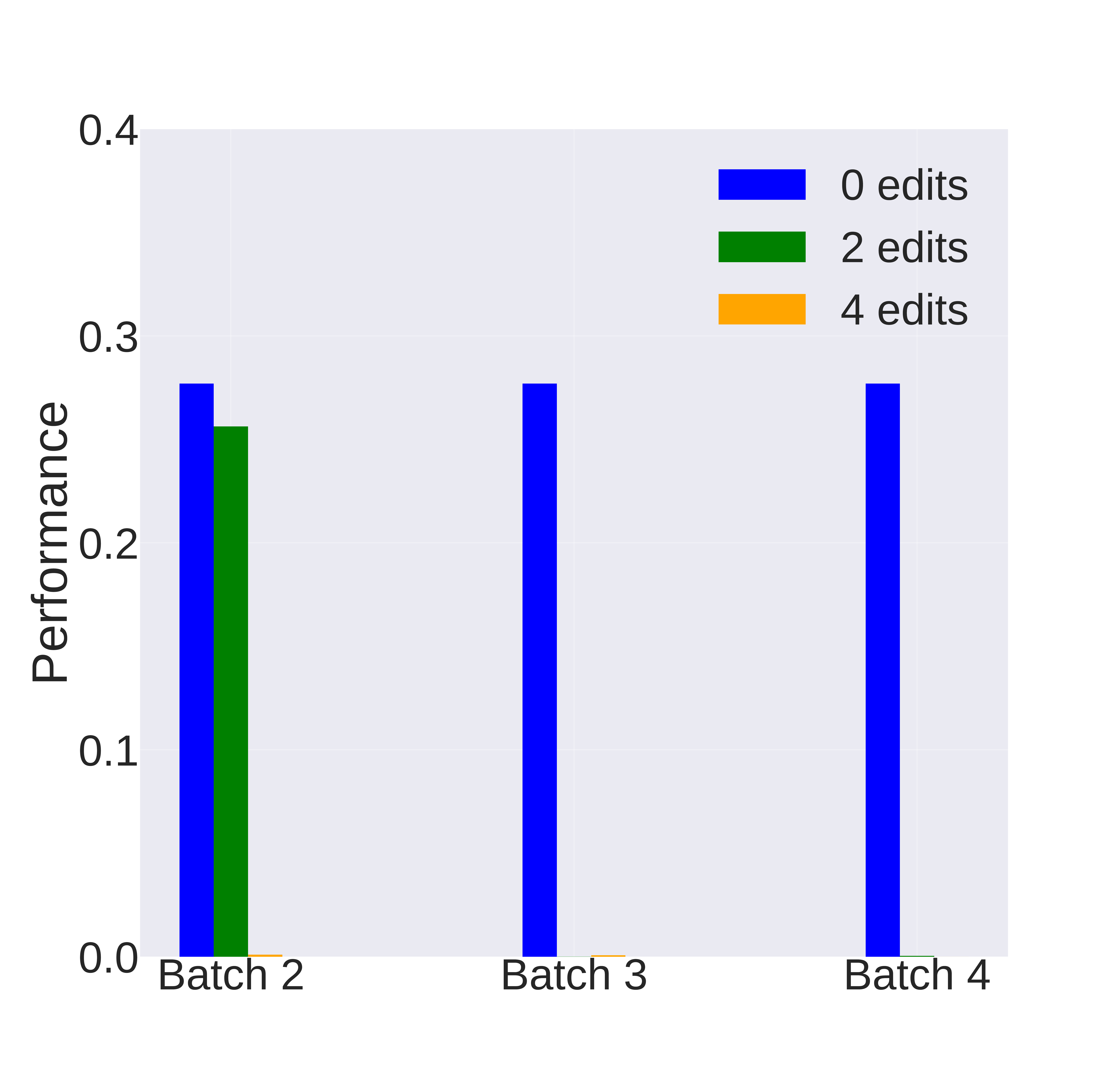}}
  \subfigure[Closed-domain QA]{
  \includegraphics[width=3.6cm]{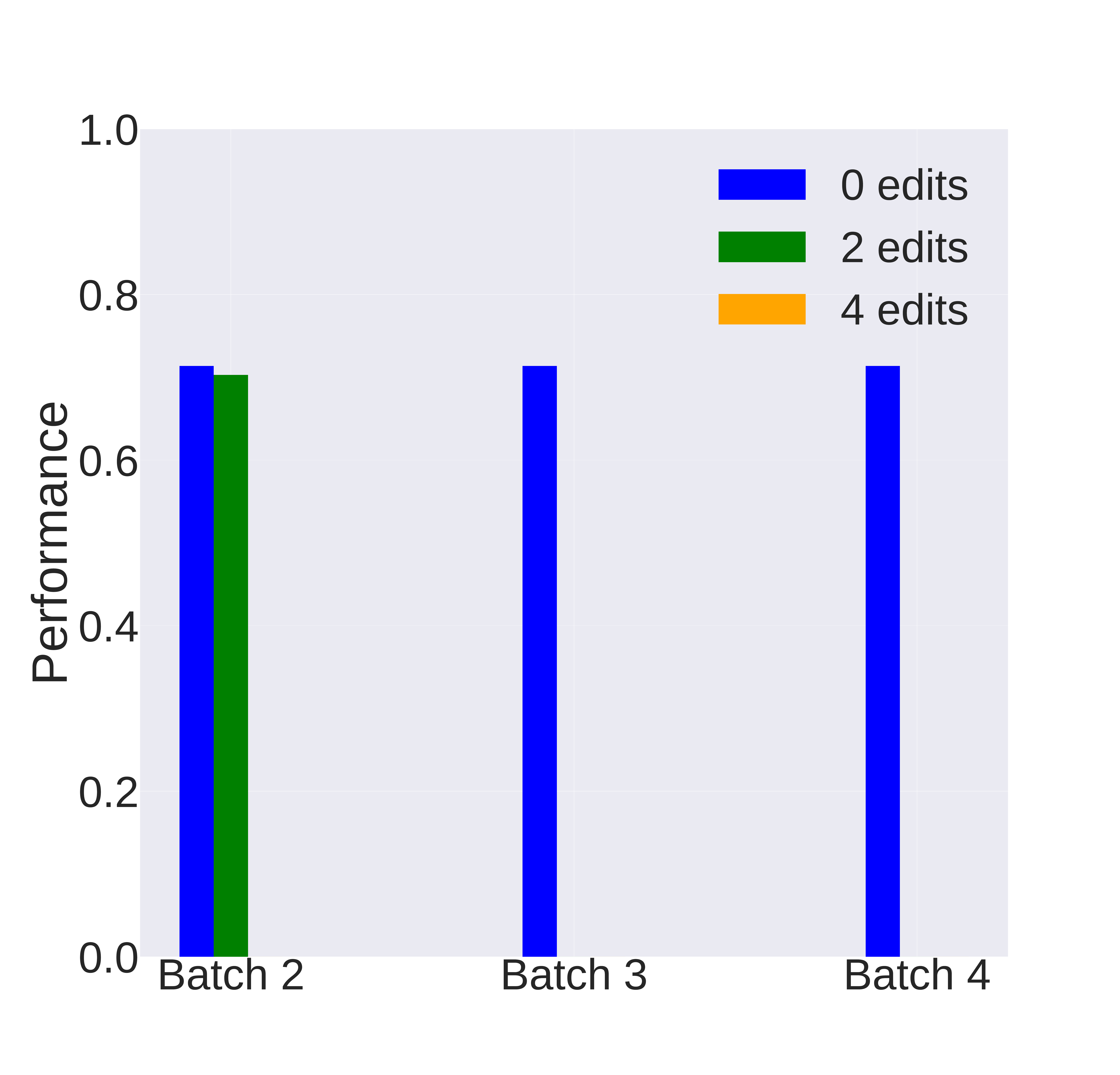}}
  \subfigure[Dialogue]{
  \includegraphics[width=3.6cm]{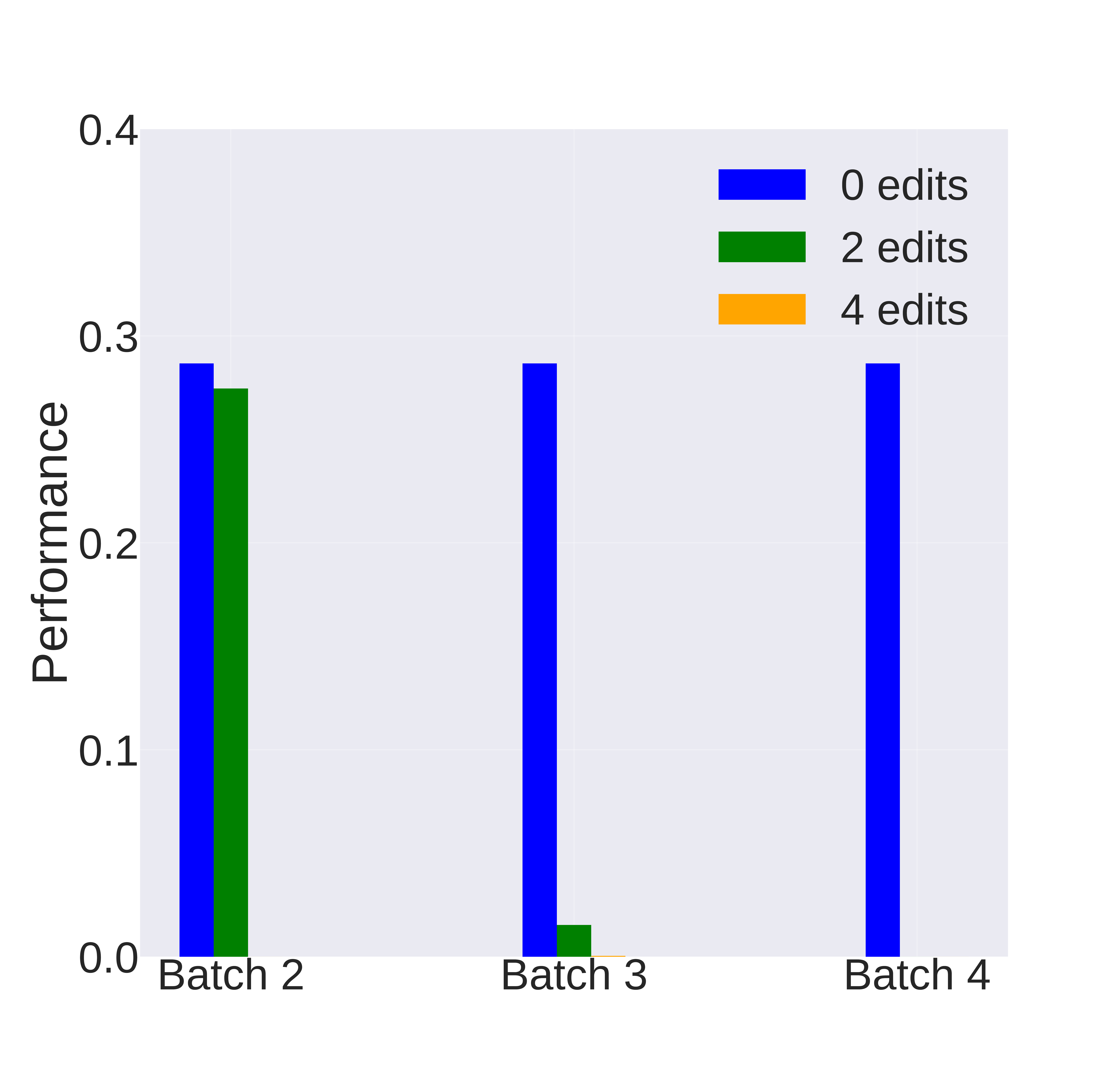}}
  \subfigure[Summarization]{
  \includegraphics[width=3.6cm]{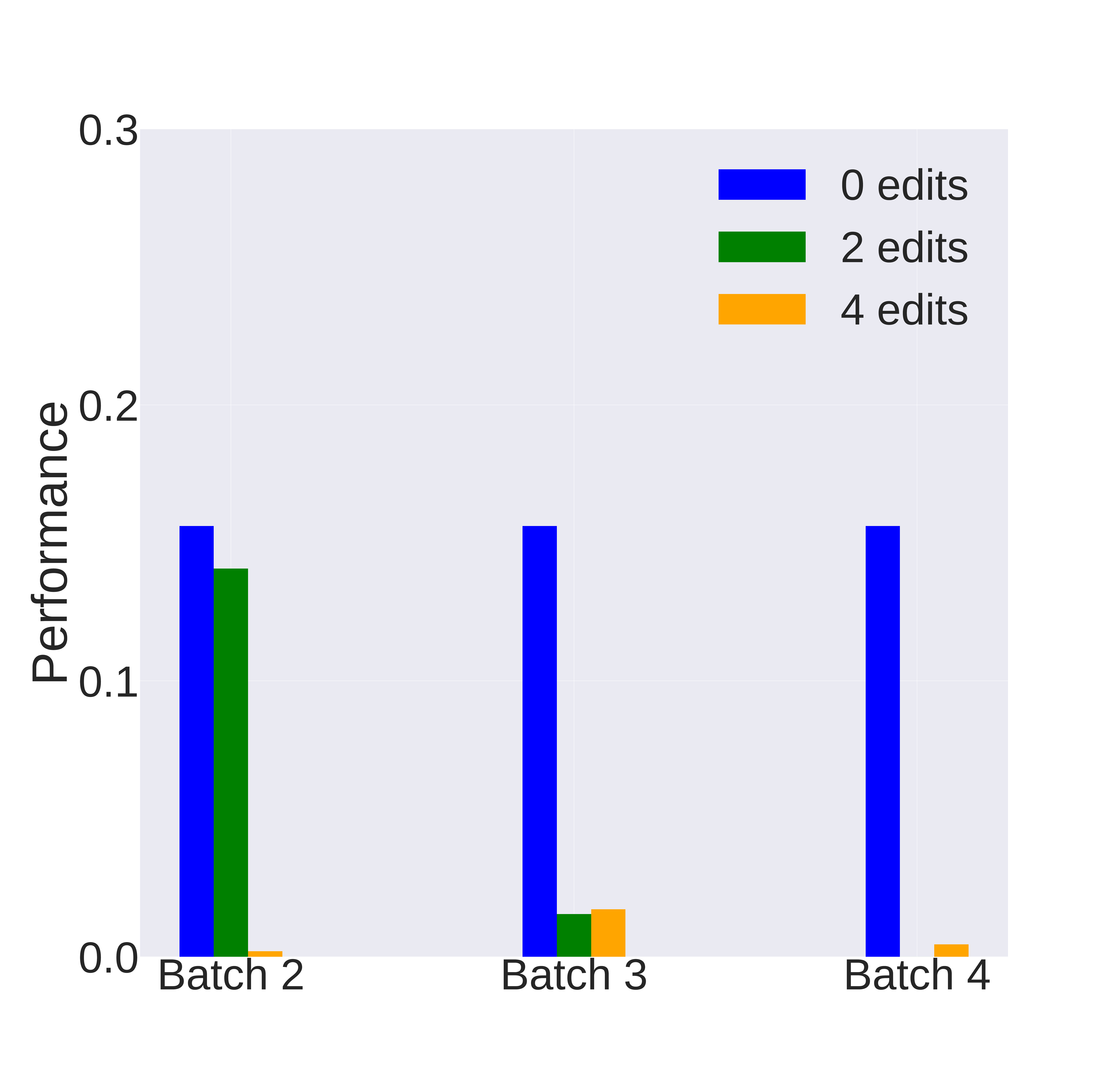}}
  \subfigure[NER]{
  \includegraphics[width=3.6cm]{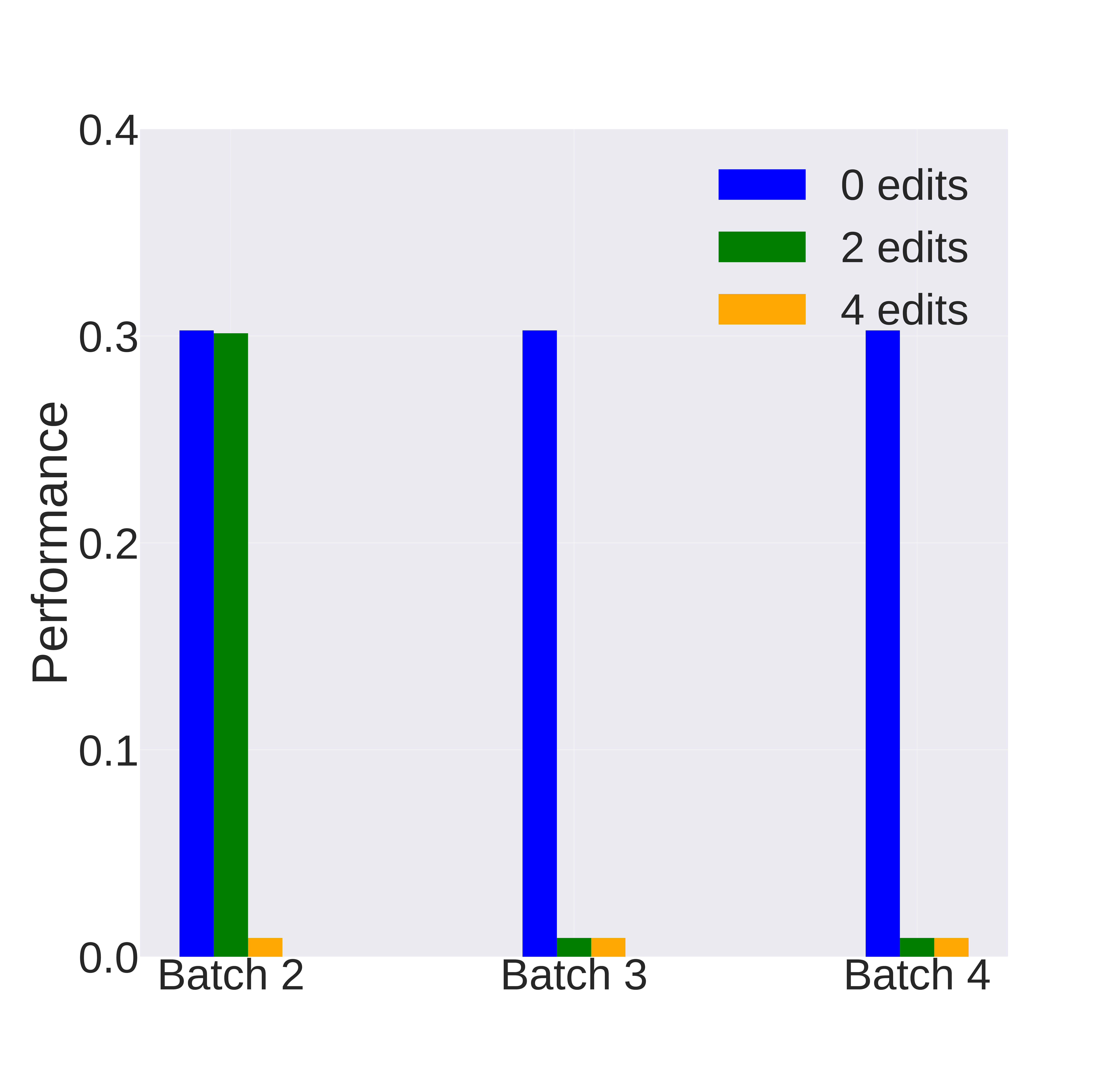}}
  \subfigure[Sentiment analysis]{
  \includegraphics[width=3.6cm]{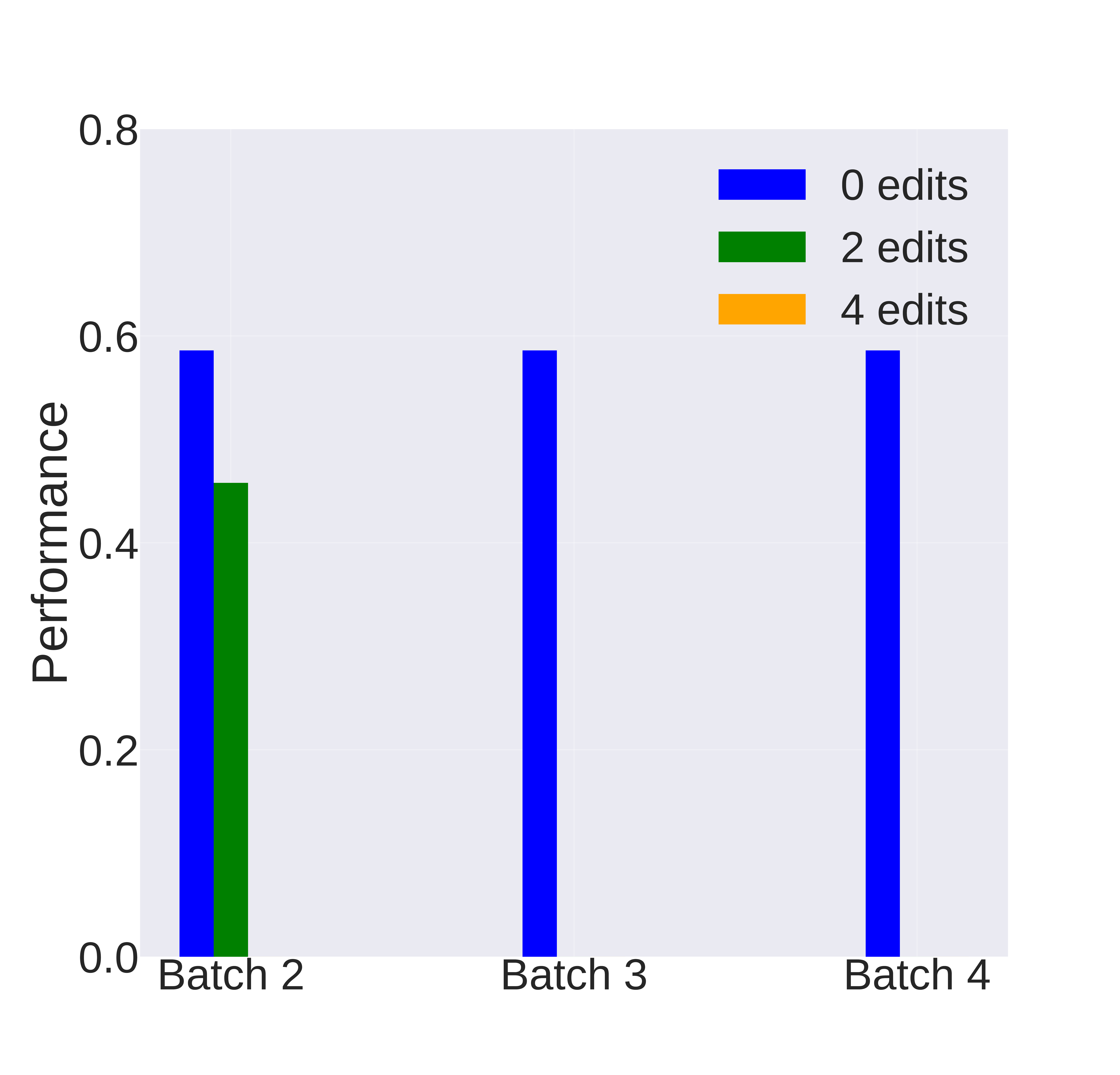}}
  \vspace{-4mm}
  \caption{Performance on general tasks of edited models using MEMIT to edit LLaMA-1 (7B) as the number of edits increases in \emph{batch- and sequential-editing}.}
  \vspace{-4mm}
  \label{fig-batch-sequential-memit-llama1-7b}
\end{figure*}

\clearpage

%%%%%%%%%%%%%%%%%%%%%%%%%%%%%%%%%%%%%%%%%%%%%%%%%%%%%%%%%%%%%%%%%%%%%%%%%%%%

\begin{figure*}[!hbt]
  \centering
  \subfigure[Reasoning]{
  \includegraphics[width=3.6cm]{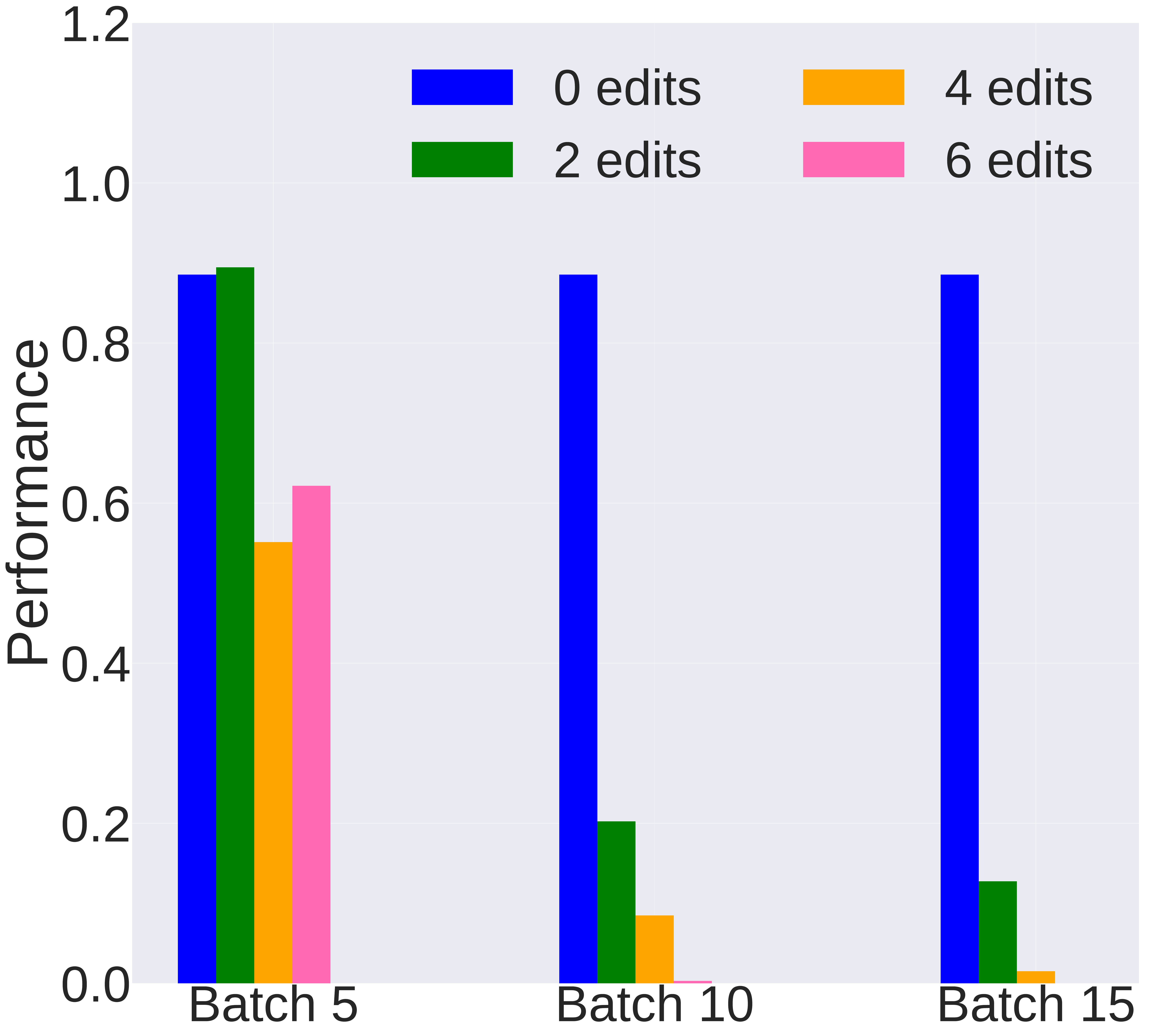}}
  \subfigure[NLI]{
  \includegraphics[width=3.6cm]{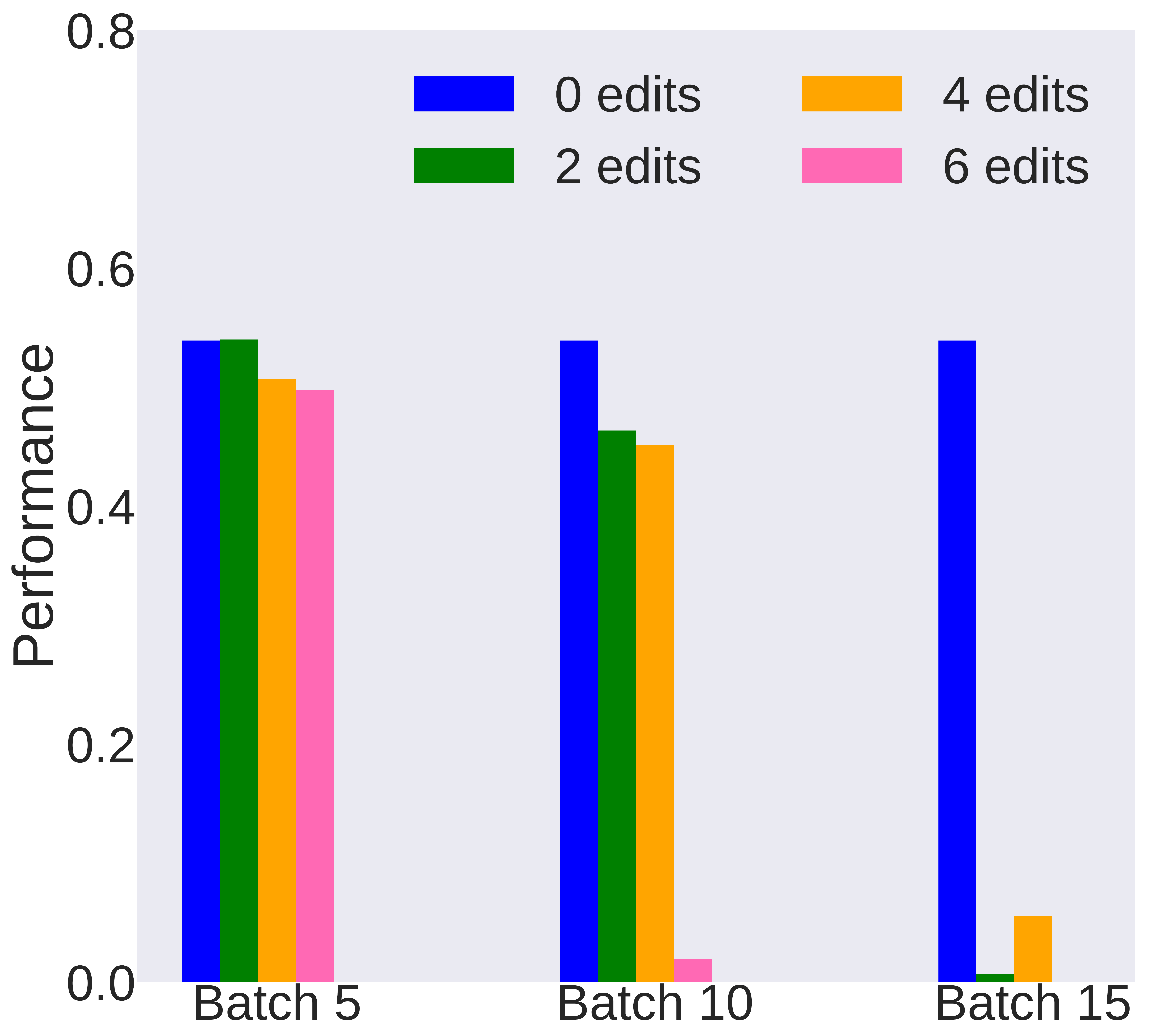}}
  \subfigure[Open-domain QA]{
  \includegraphics[width=3.6cm]{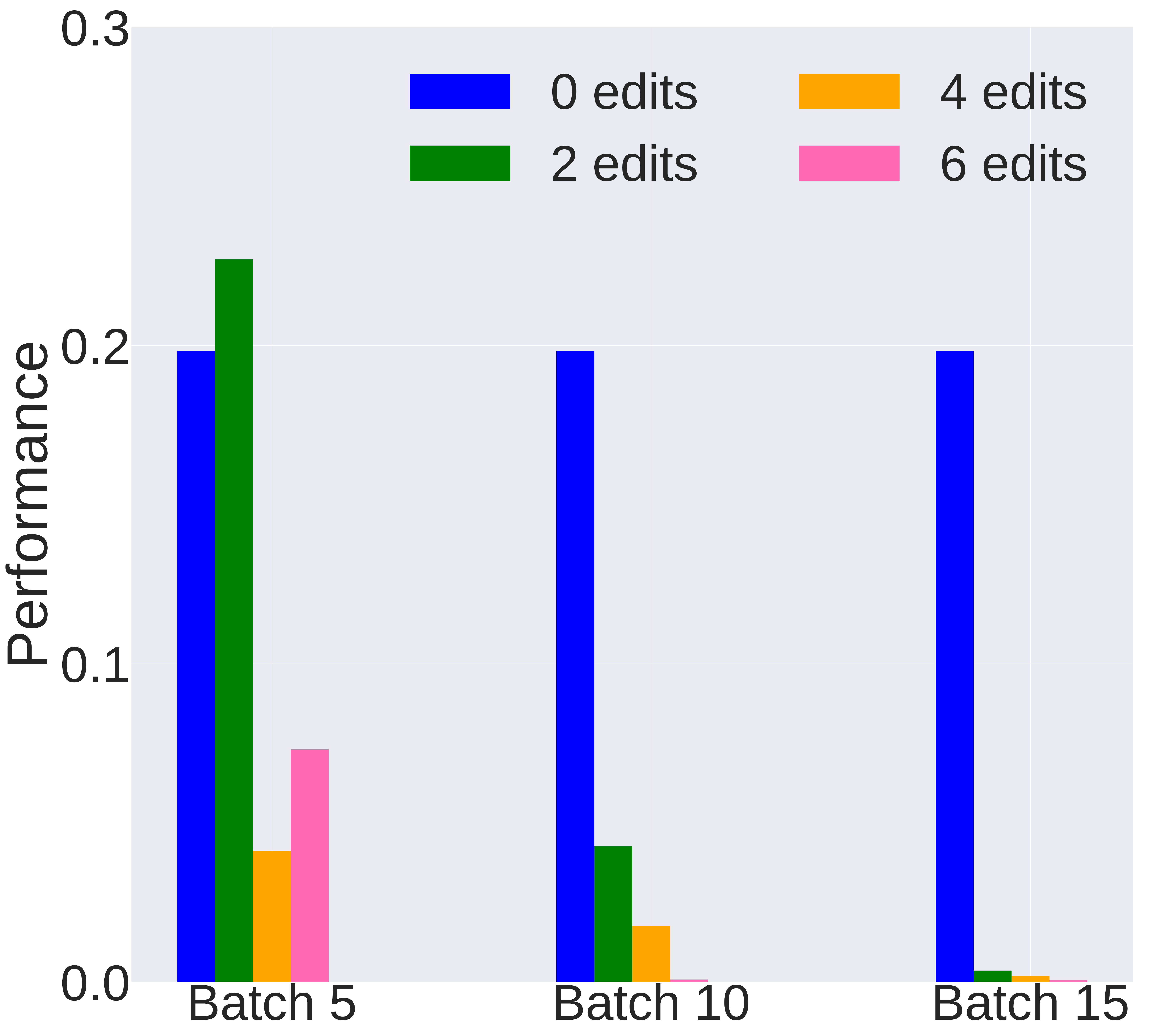}}
  \subfigure[Closed-domain QA]{
  \includegraphics[width=3.6cm]{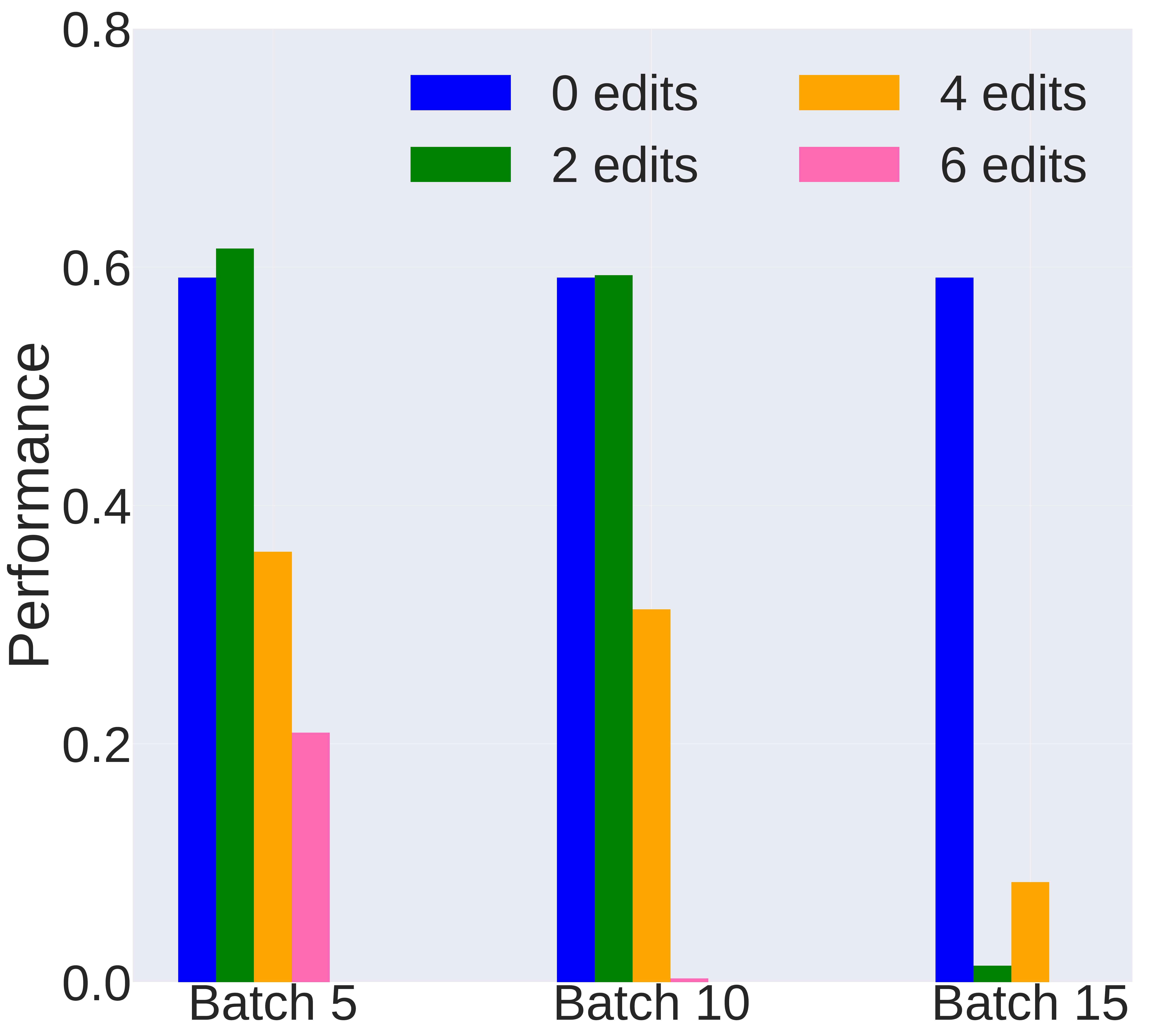}}
  \subfigure[Dialogue]{
  \includegraphics[width=3.6cm]{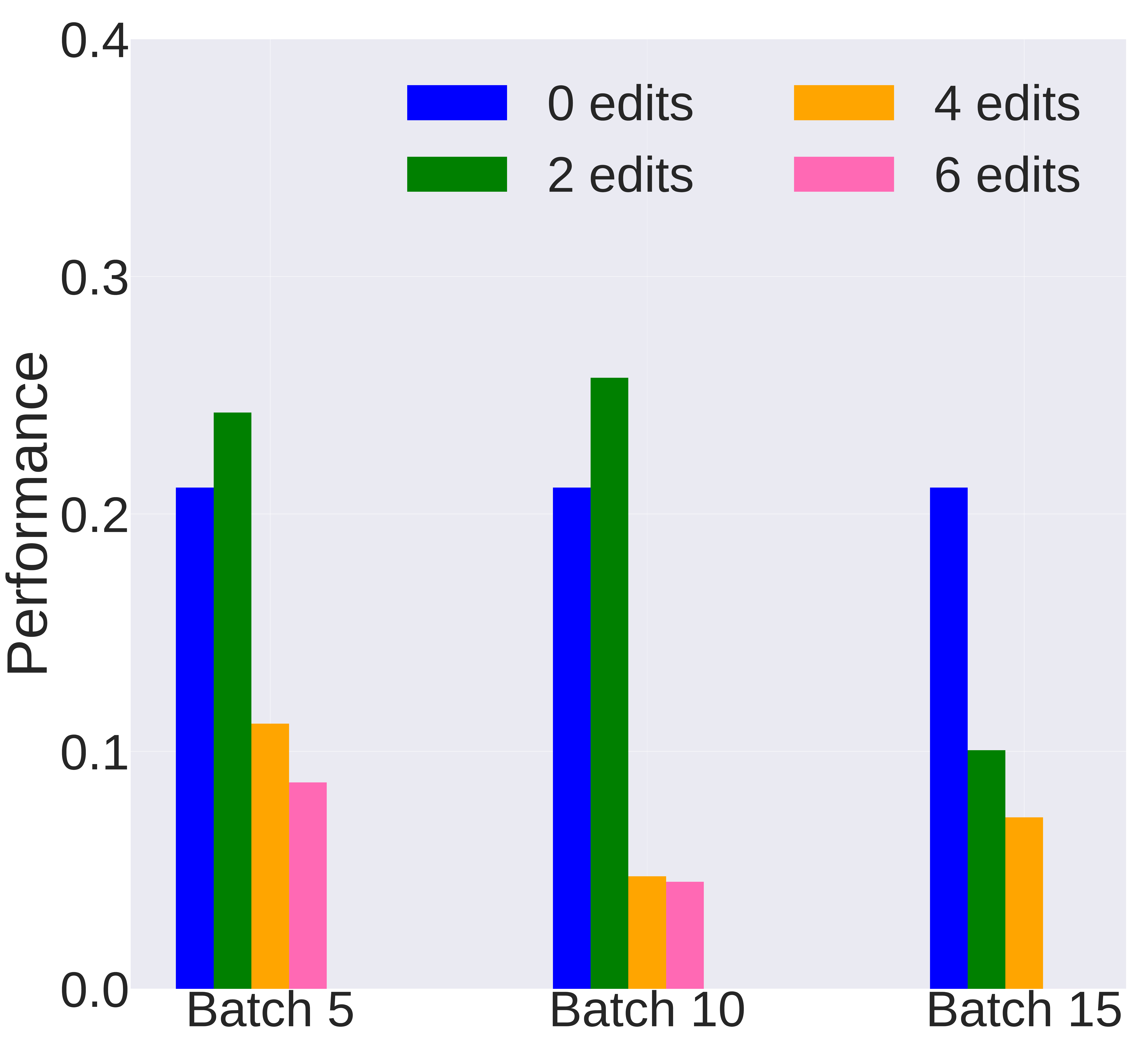}}
  \subfigure[Summarization]{
  \includegraphics[width=3.6cm]{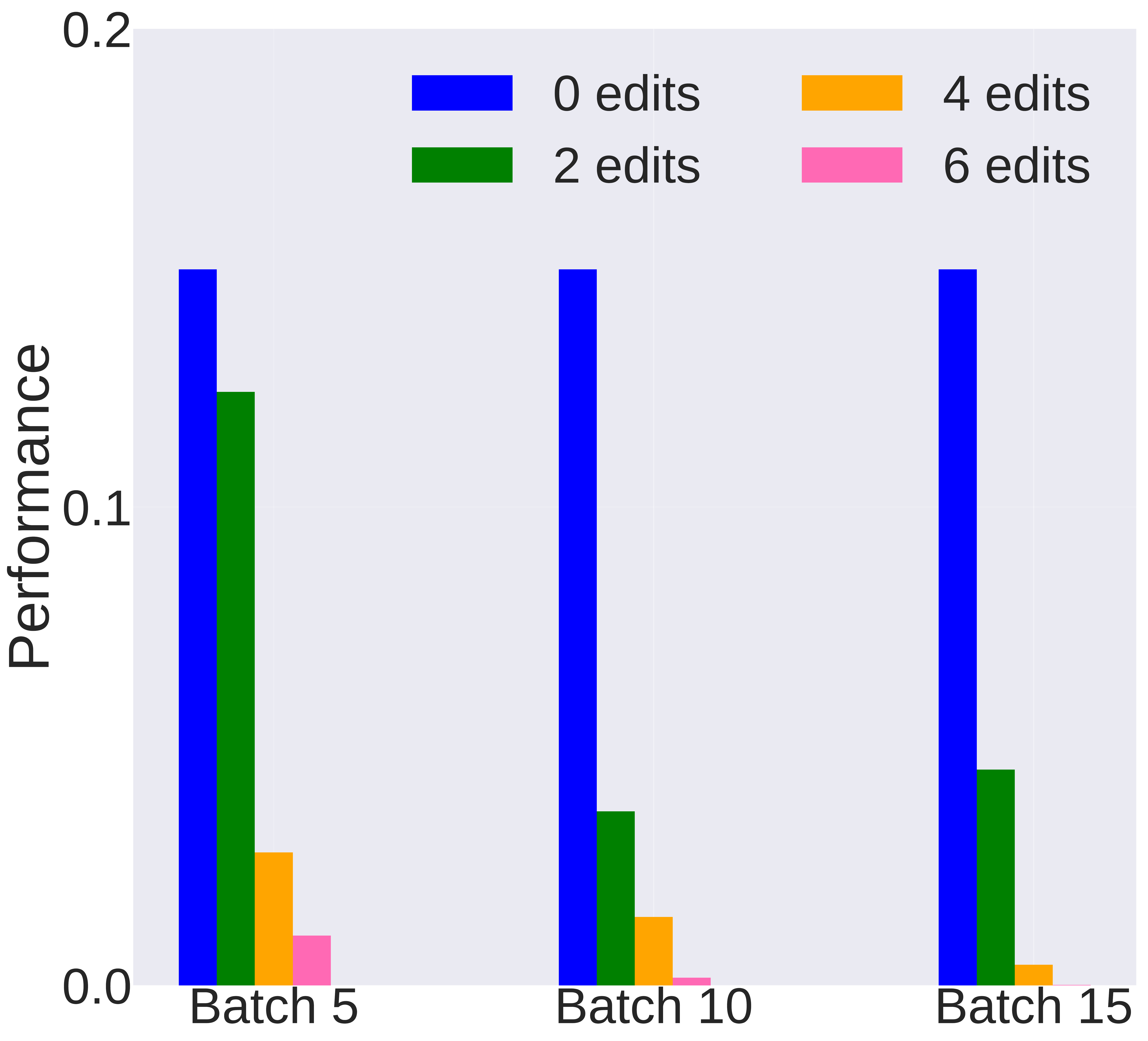}}
  \subfigure[NER]{
  \includegraphics[width=3.6cm]{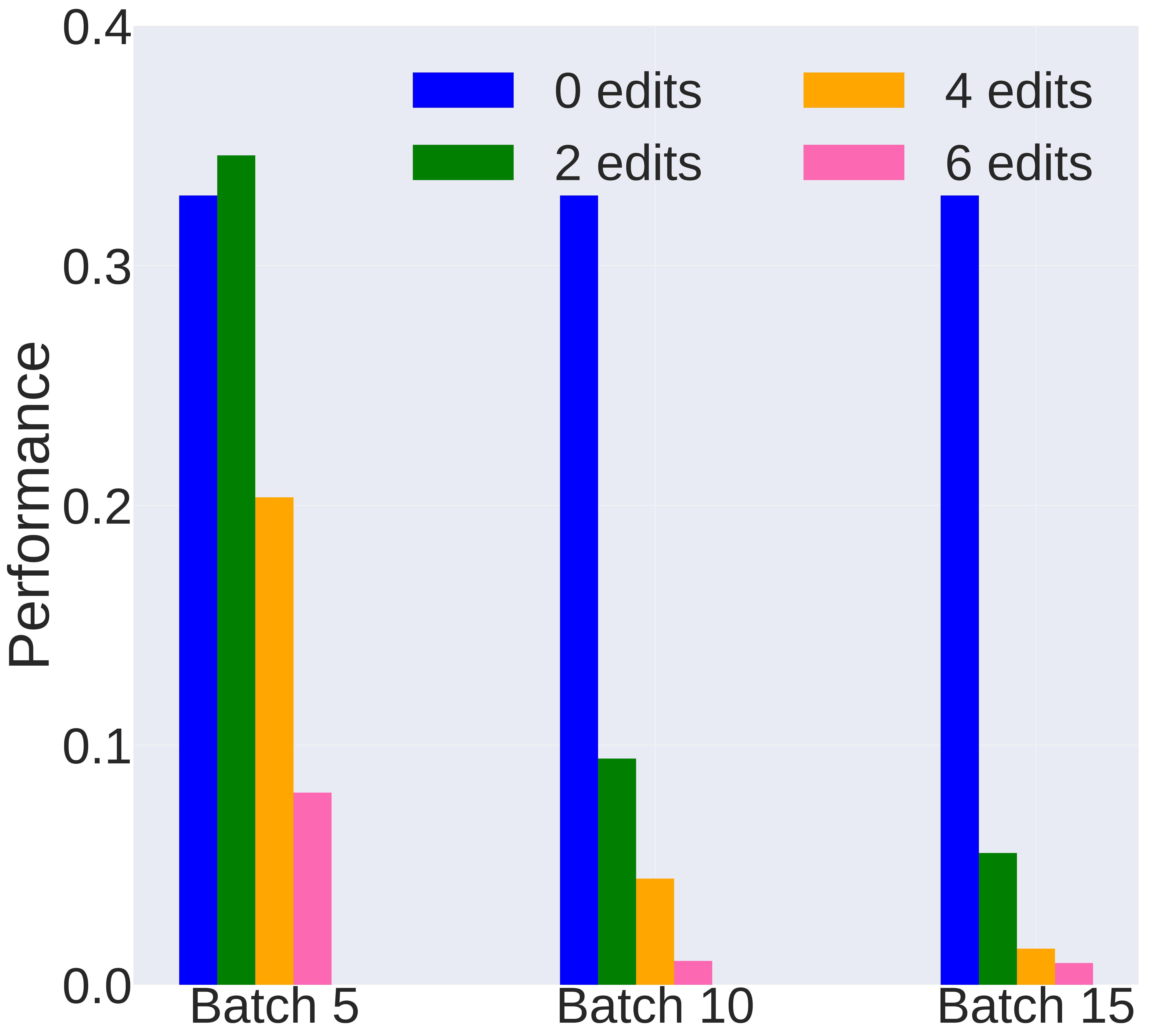}}
  \subfigure[Sentiment analysis]{
  \includegraphics[width=3.6cm]{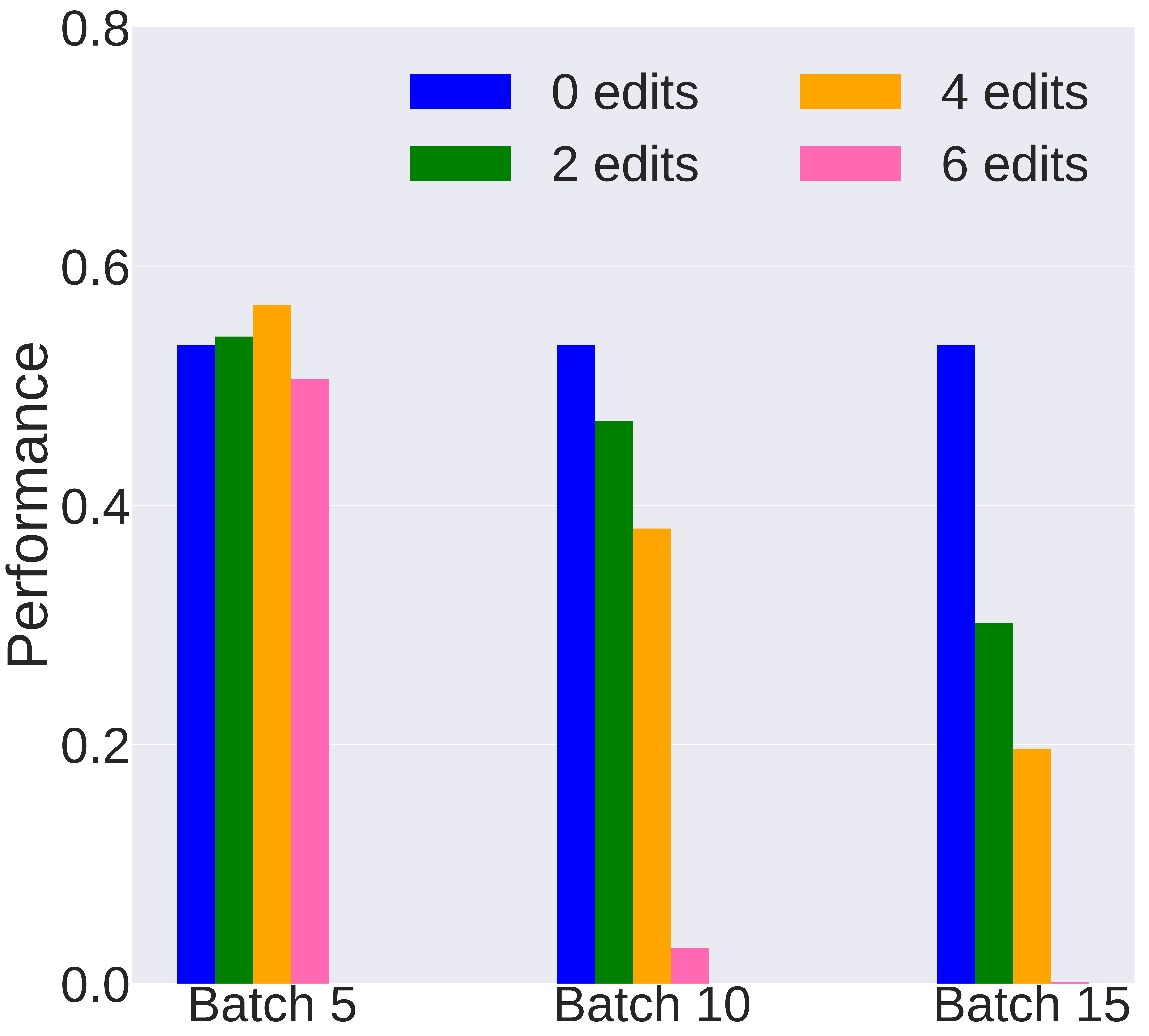}}
  \vspace{-4mm}
  \caption{Performance on general tasks of edited models using MEND to edit LLaMA-2 (7B) as the number of edits increases in \emph{batch- and sequential-editing}.}
  \vspace{-2mm}
  \label{fig-batch-sequential-mend-llama2-7b}
\end{figure*}

%%%%%%%%%%%%%%%%%%%%%%%%%%%%%%%%%%%%%%%%%%%%%%%%%%%%%%%%%%%%%%%%%%%%%%%%%%%%

\begin{figure*}[!hbt]
  \centering
  \subfigure[Reasoning]{
  \includegraphics[width=3.6cm]{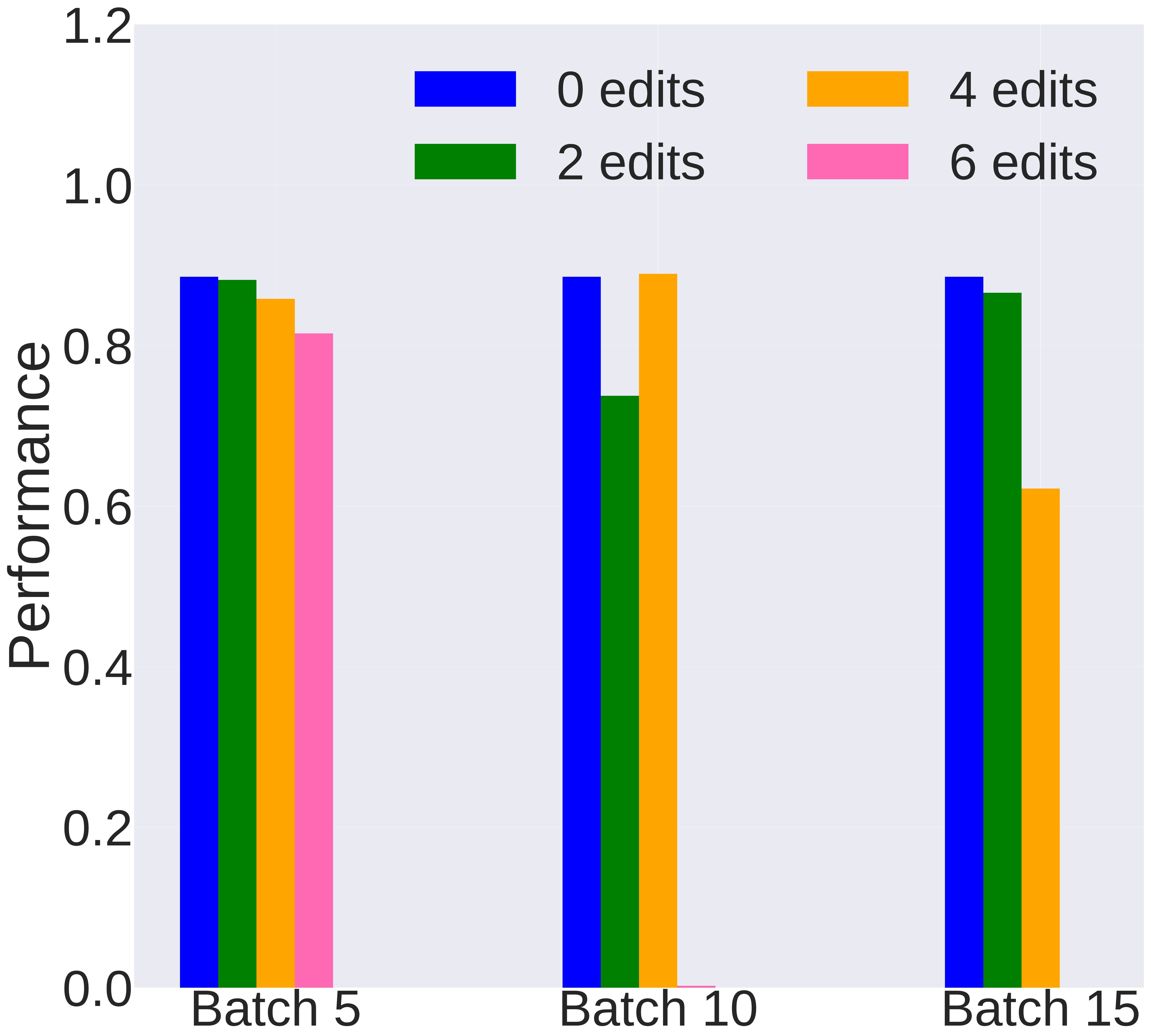}}
  \subfigure[NLI]{
  \includegraphics[width=3.6cm]{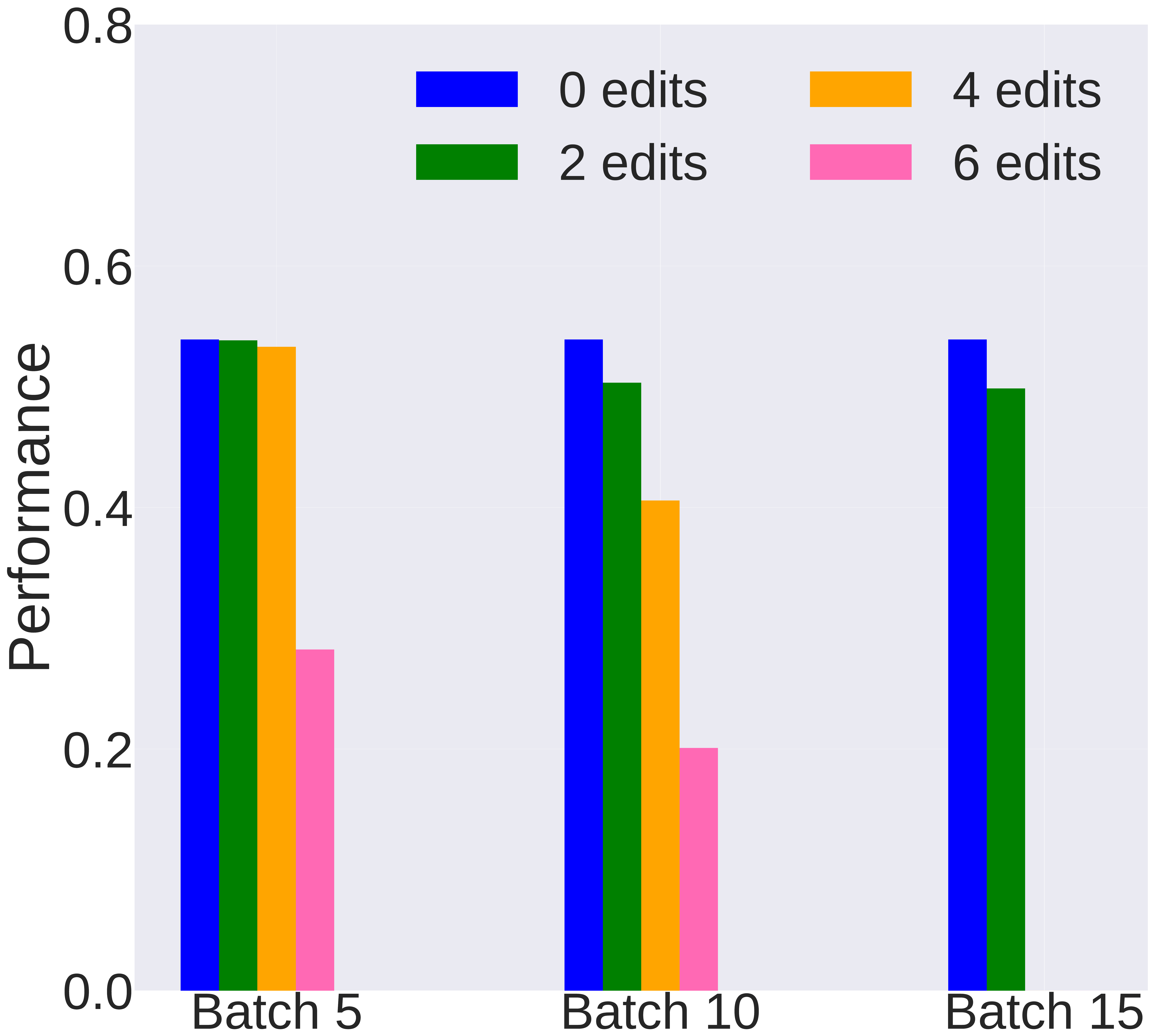}}
  \subfigure[Open-domain QA]{
  \includegraphics[width=3.6cm]{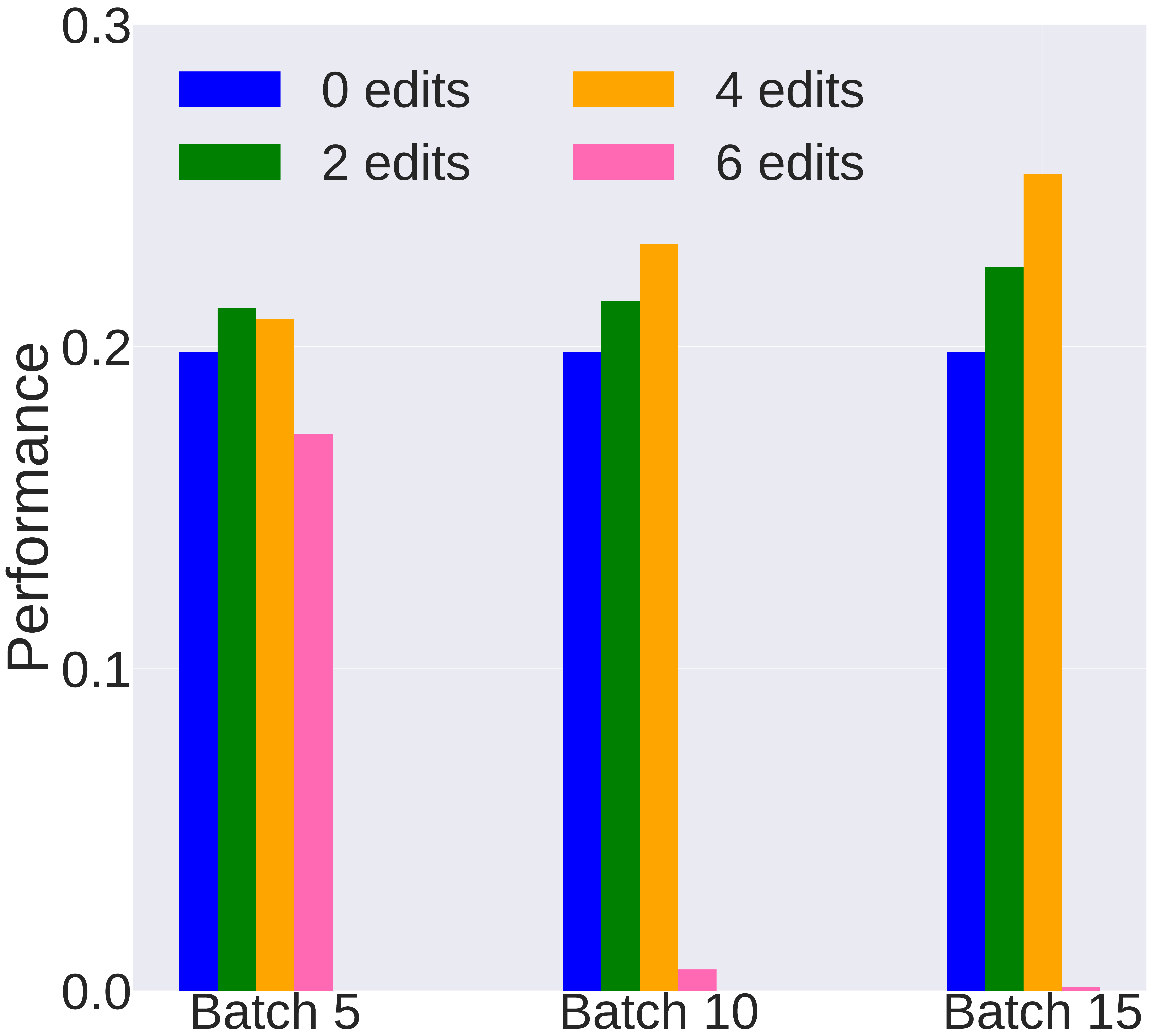}}
  \subfigure[Closed-domain QA]{
  \includegraphics[width=3.6cm]{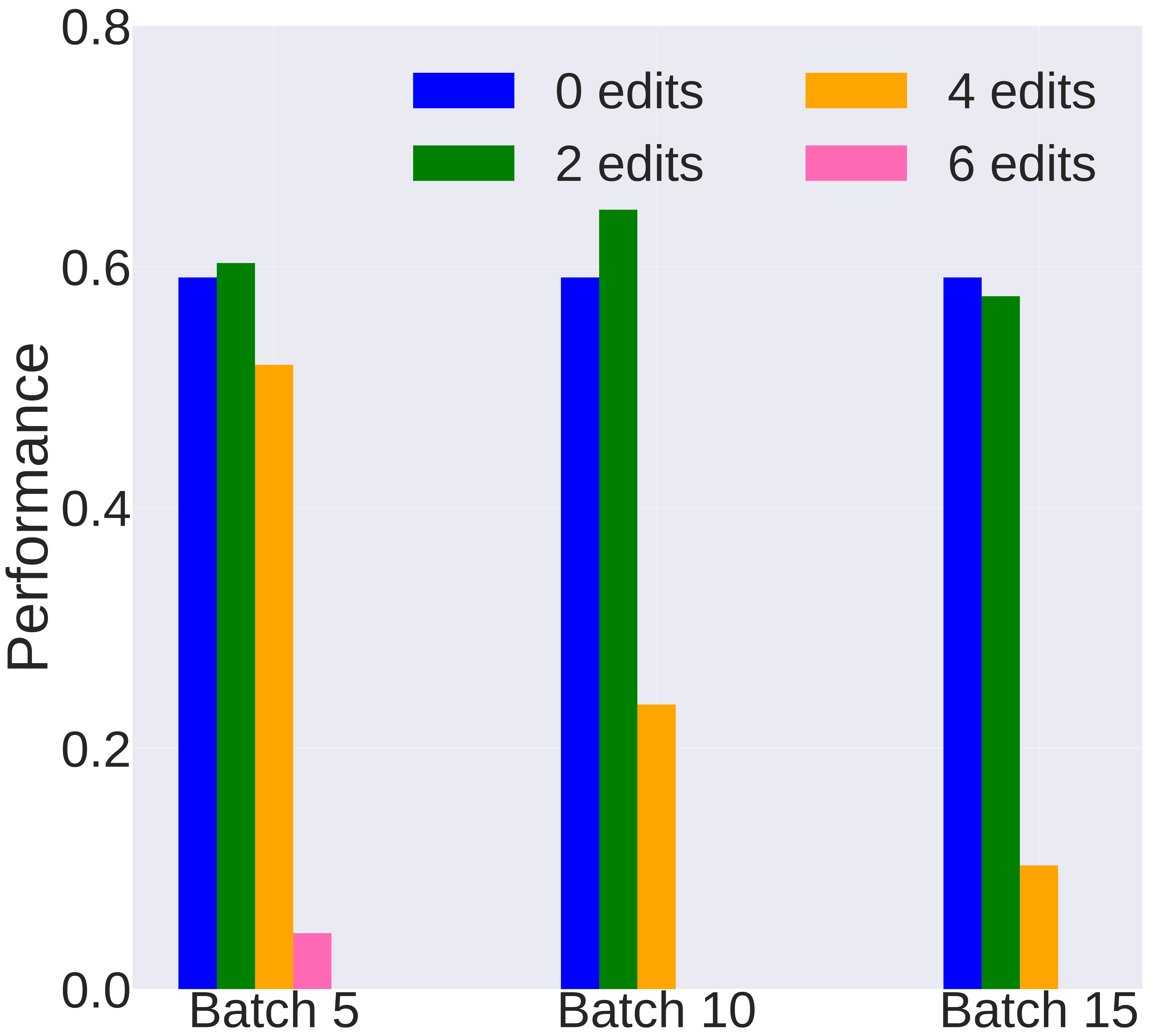}}
  \subfigure[Dialogue]{
  \includegraphics[width=3.6cm]{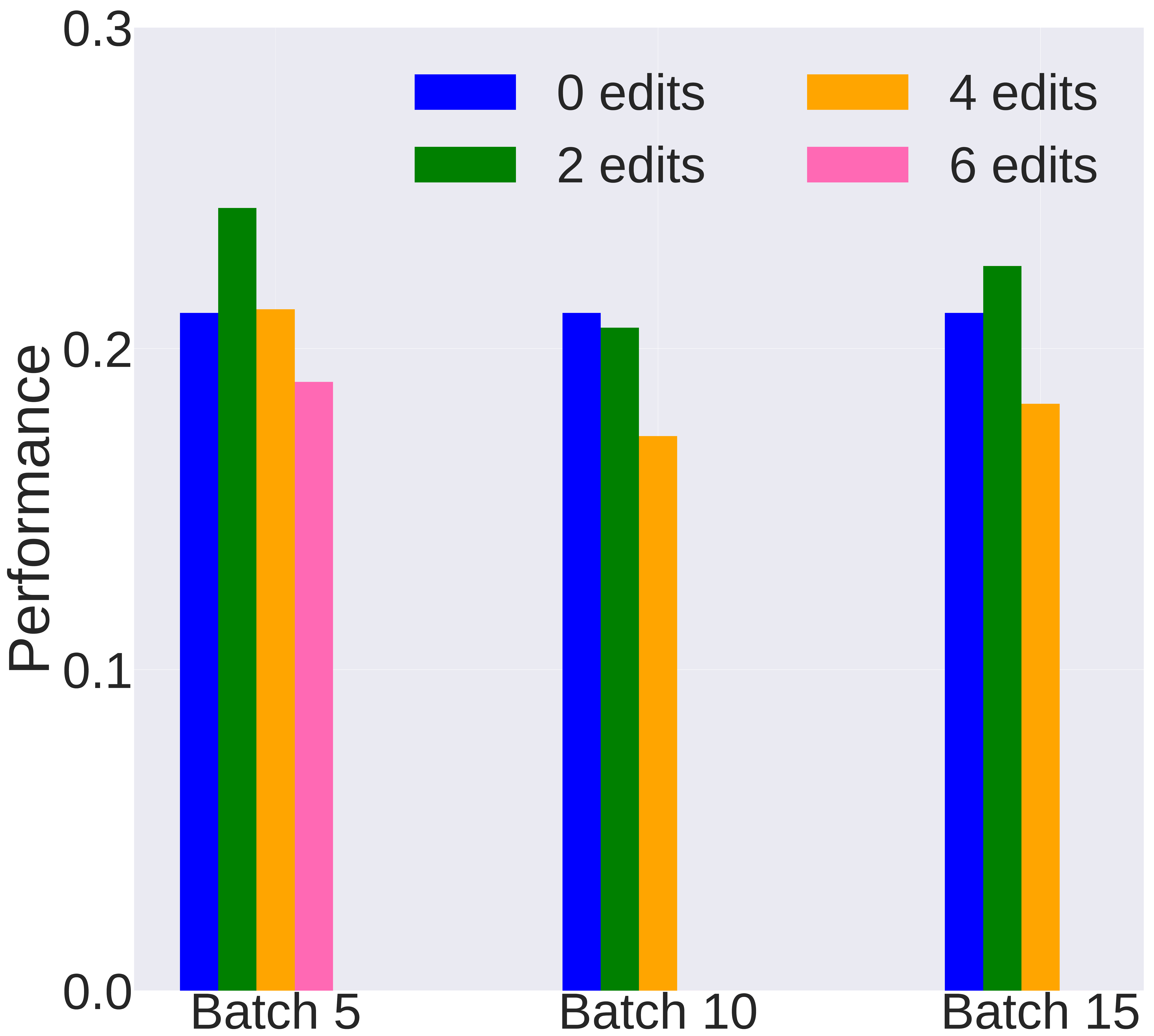}}
  \subfigure[Summarization]{
  \includegraphics[width=3.6cm]{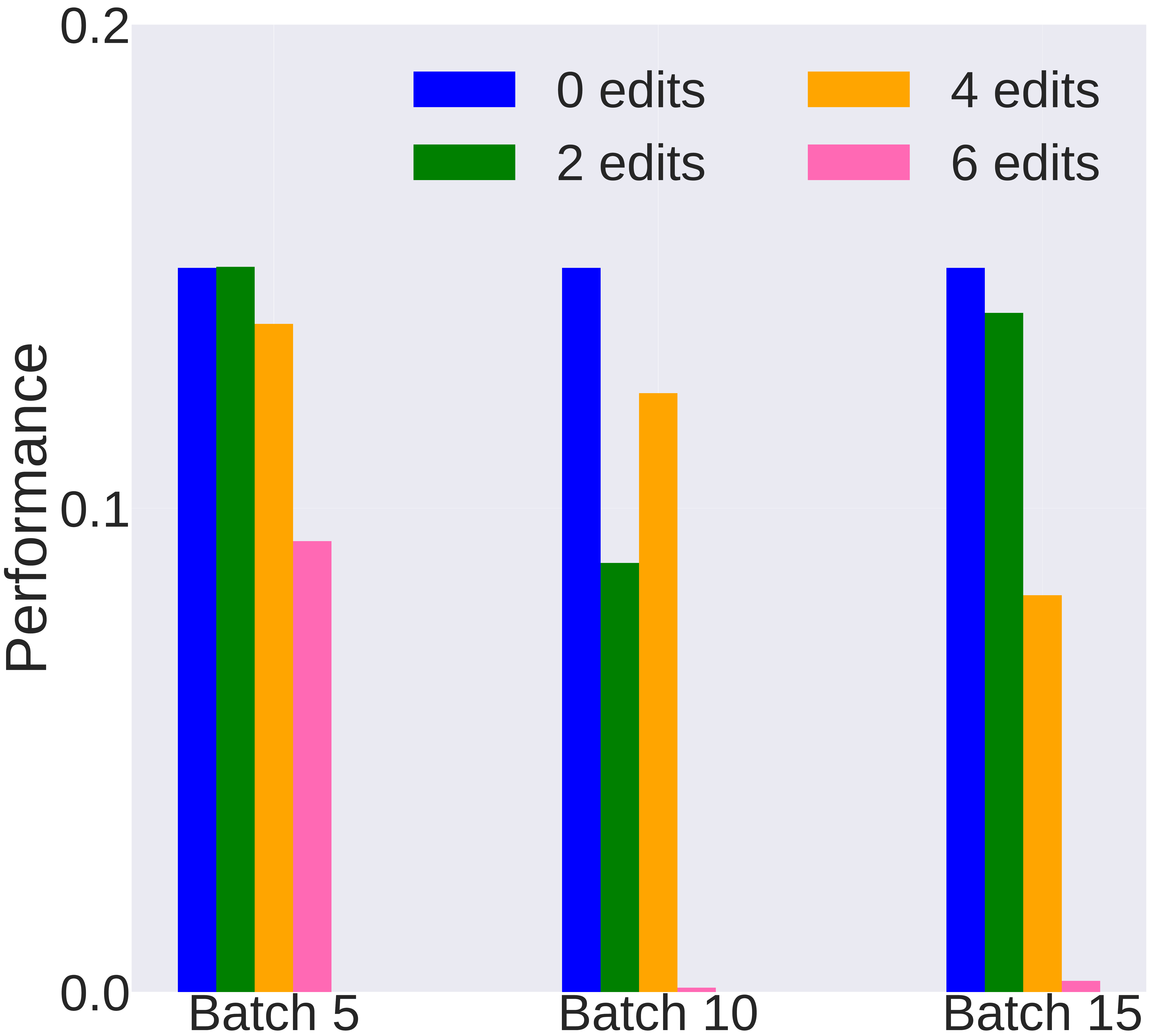}}
  \subfigure[NER]{
  \includegraphics[width=3.6cm]{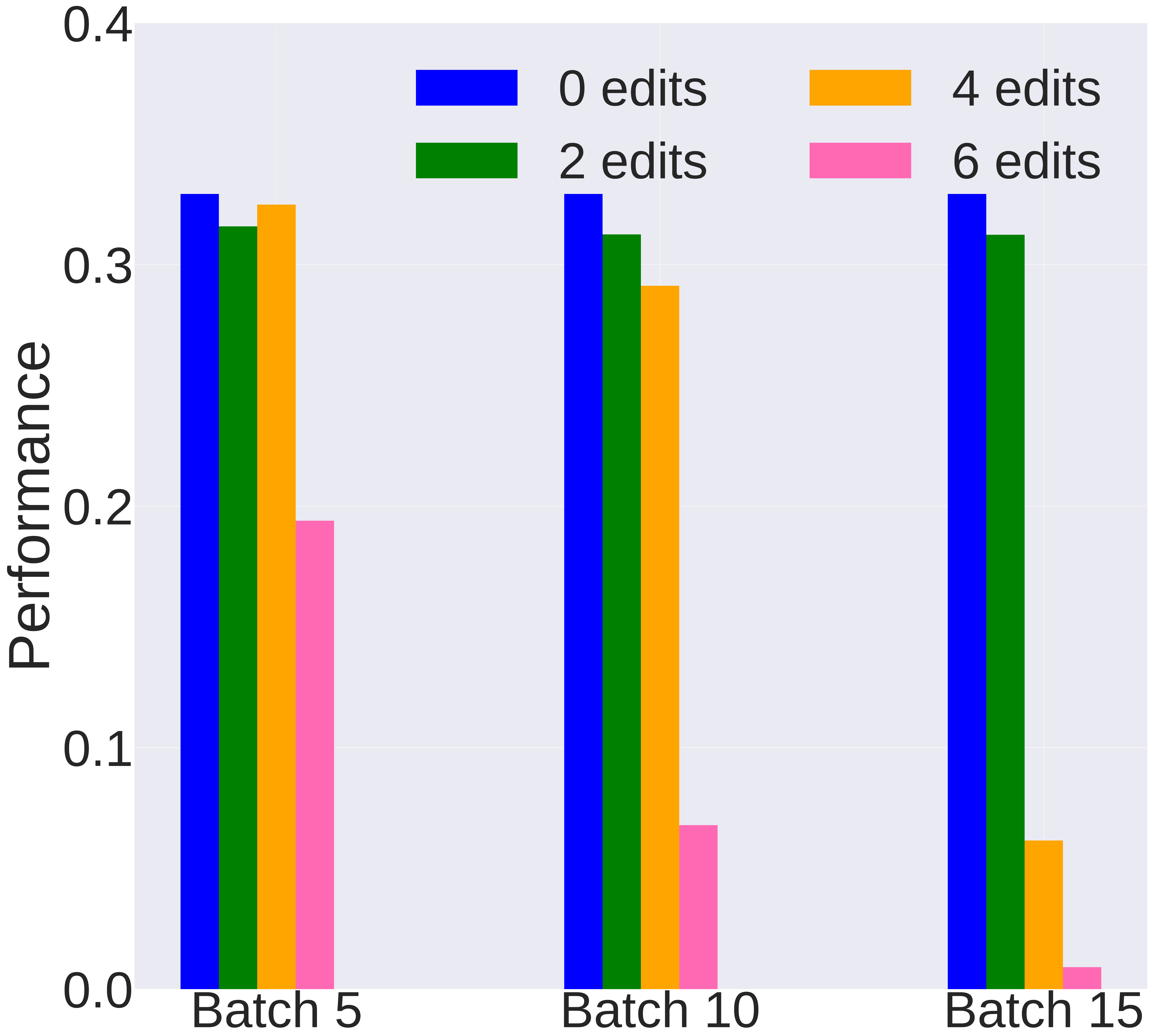}}
  \subfigure[Sentiment analysis]{
  \includegraphics[width=3.6cm]{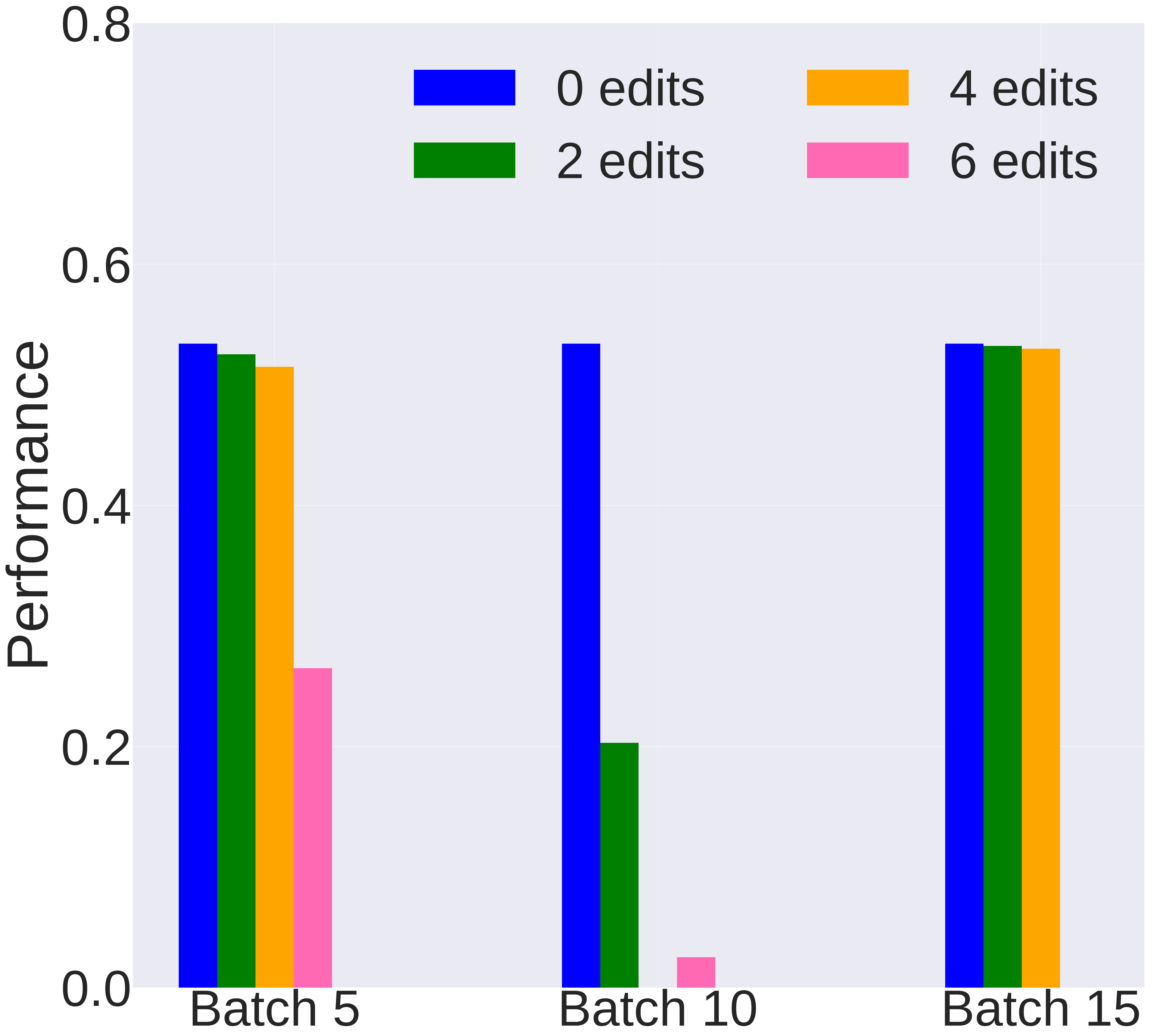}}
  \vspace{-4mm}
  \caption{Performance on general tasks of edited models using MEMIT to edit LLaMA-2 (7B) as the number of edits increases in \emph{batch- and sequential-editing}.}
  \vspace{-4mm}
  \label{fig-batch-sequential-memit-llama2-7b}
\end{figure*}

%%%%%%%%%%%%%%%%%%%%%%%%%%%%%%%%%%%%%%%%%%%%%%%%%%%%%%%%%%%%%%%%%%%%%%%%%%%%

\clearpage
\subsection{Results of \MODELNAME{}} \label{sec-appendix-result-rect}

\begin{figure}[!hbt]
  \centering
  \subfigure[ROME on LLaMA-2 (7B)]{
  \includegraphics[width=3.8cm]{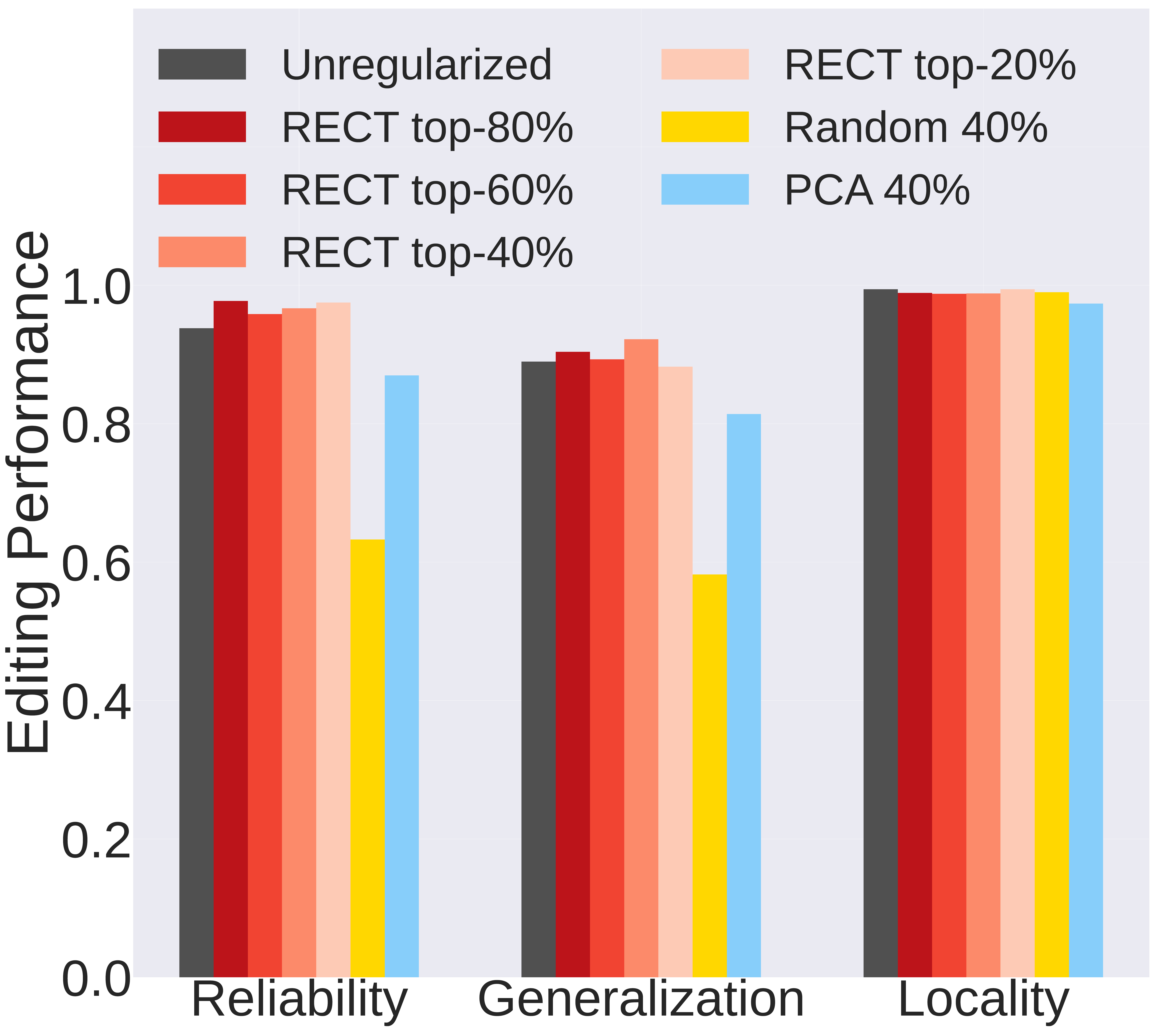}}
  \subfigure[MEMIT on LLaMA-2 (7B)]{
  \includegraphics[width=3.8cm]{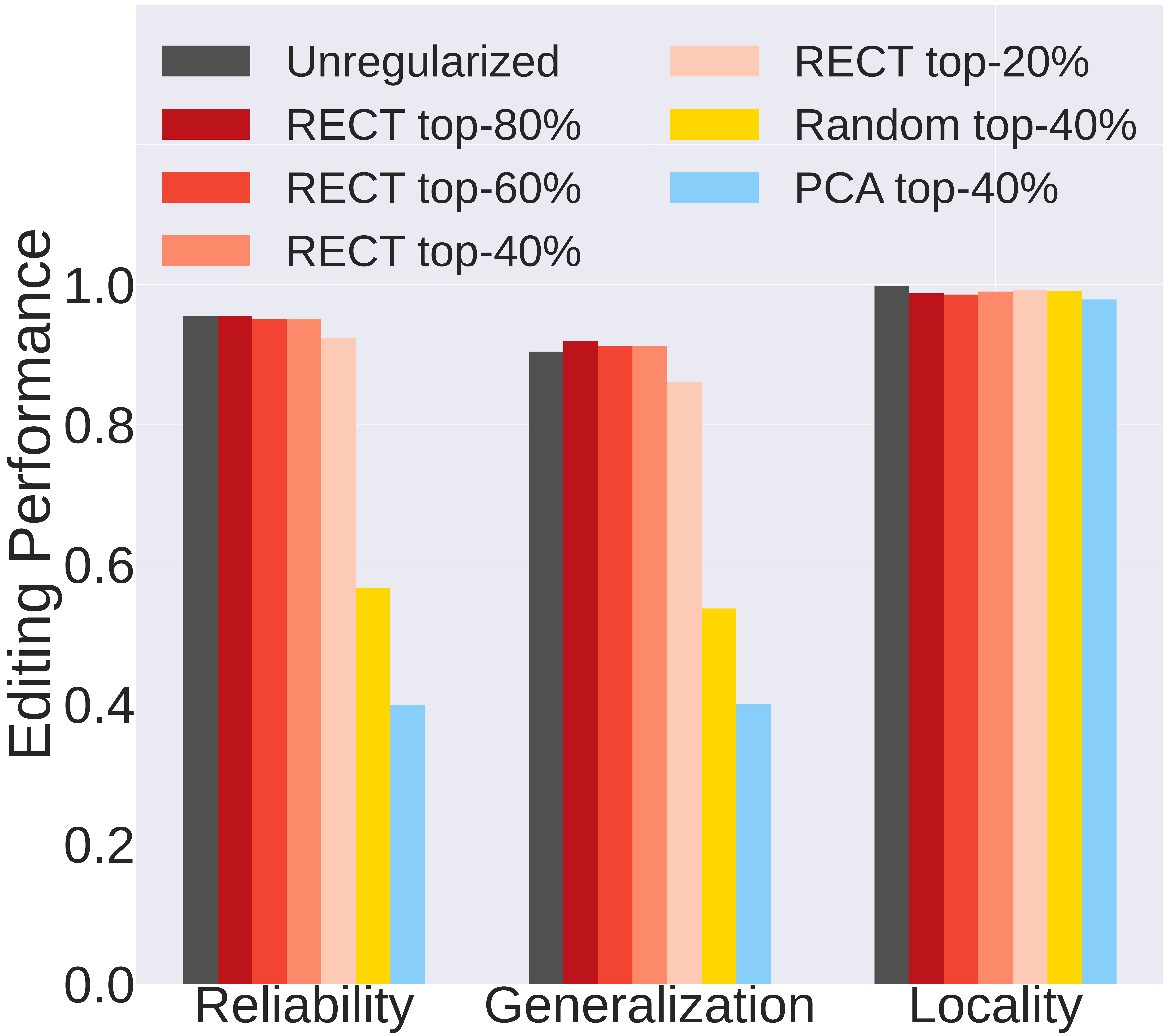}}
  \vspace{-2mm}
  \caption{Comparison of introducing various regularization methods and how the \emph{editing performance} change with respect to different top-\emph{k}\% for \MODELNAME{}.}
  \vspace{-4mm}
  \label{fig-rect-edit-2}
\end{figure}

%%%%%%%%%%%%%%%%%%%%%%%%%%%%%%%%%%%%%%%%%%%%%%%%%%%%%%%%%%%%%%%%%%%%%%%%%%%%

\begin{figure*}[!hbt]
  \centering
  \subfigure[Dialogue]{
  \includegraphics[width=3.8cm]{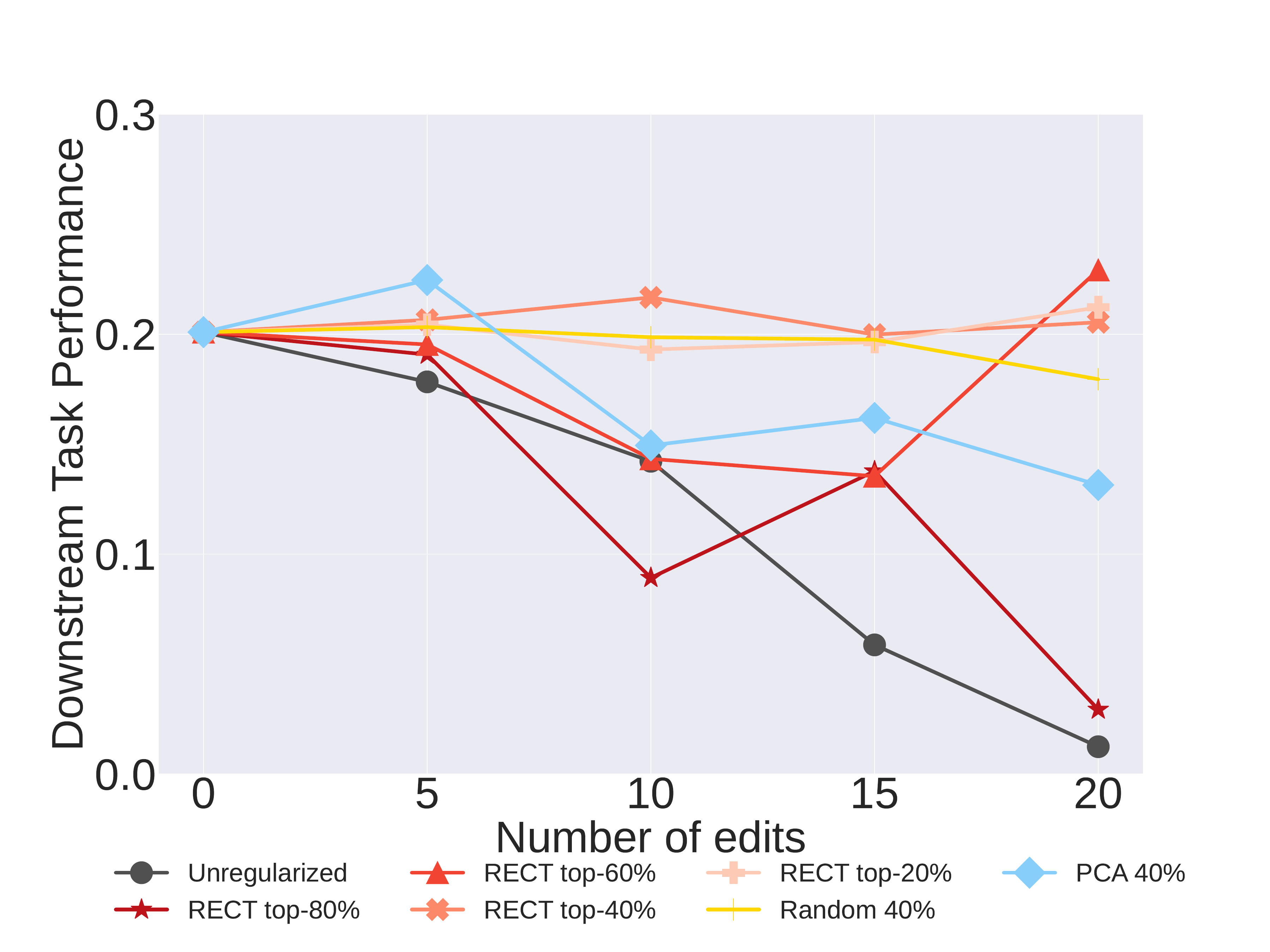}}
  \subfigure[NER]{
  \includegraphics[width=3.8cm]{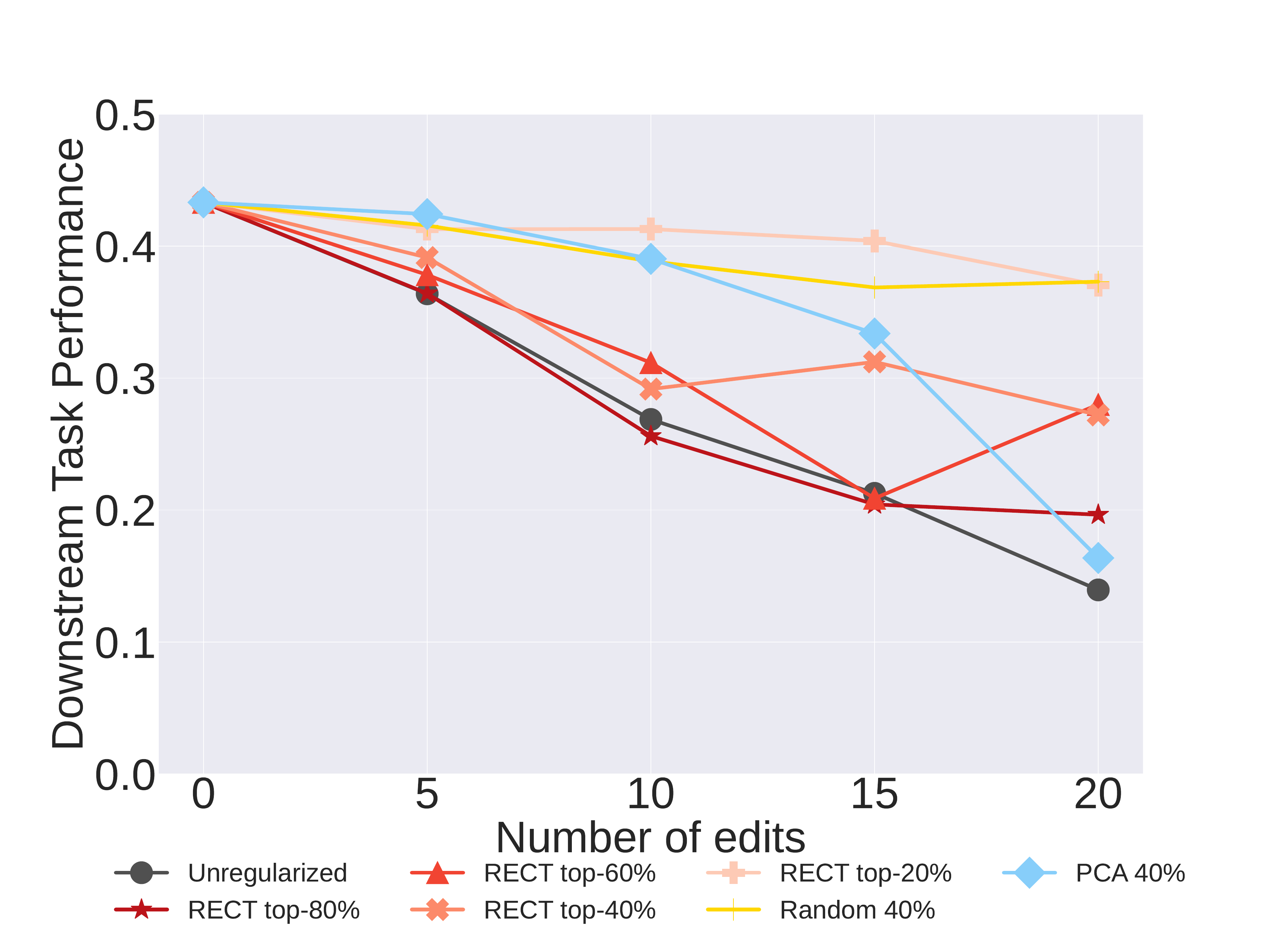}}
  \subfigure[NLI]{
  \includegraphics[width=3.8cm]{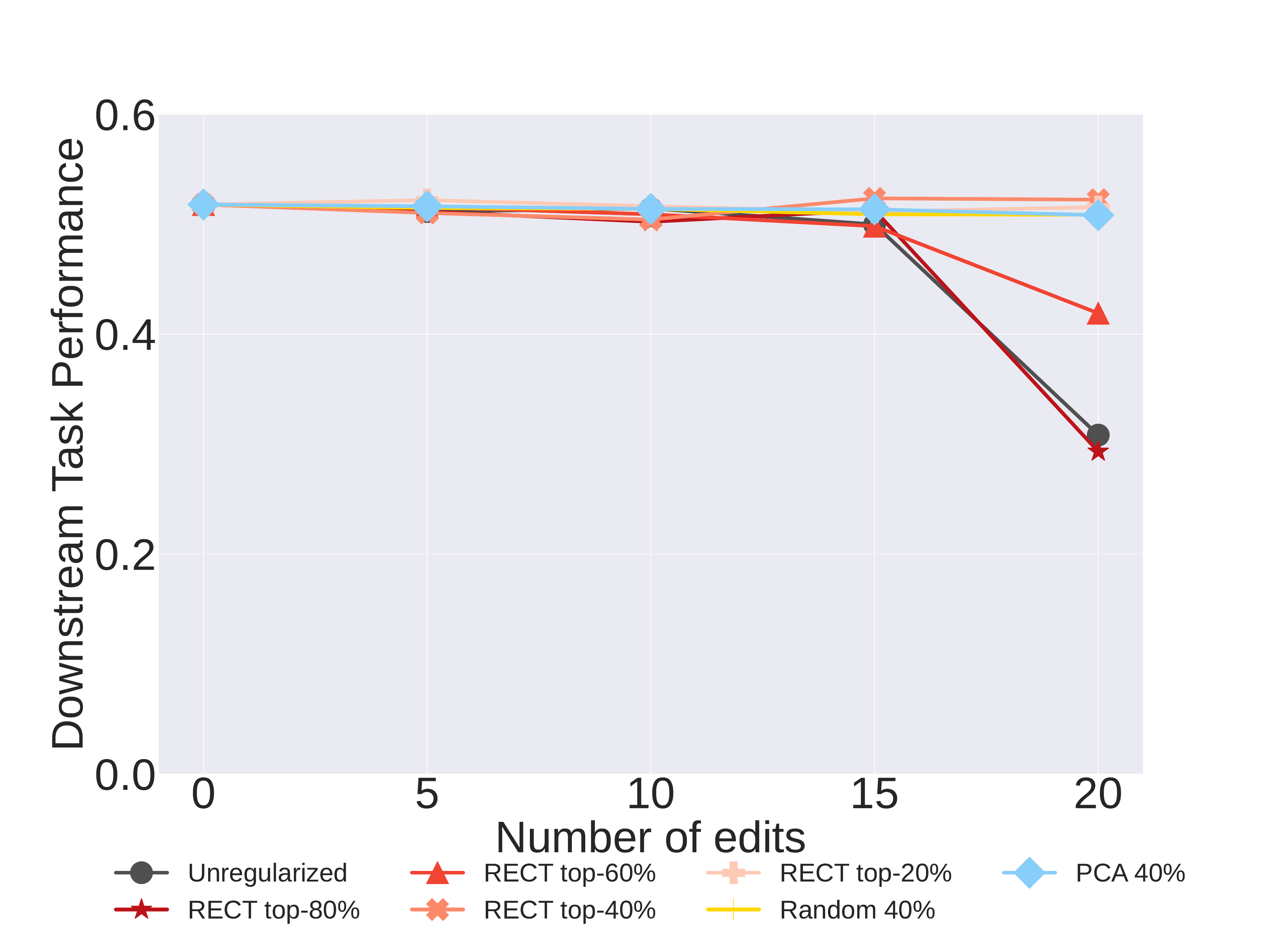}}
  \subfigure[Reasoning]{
  \includegraphics[width=3.8cm]{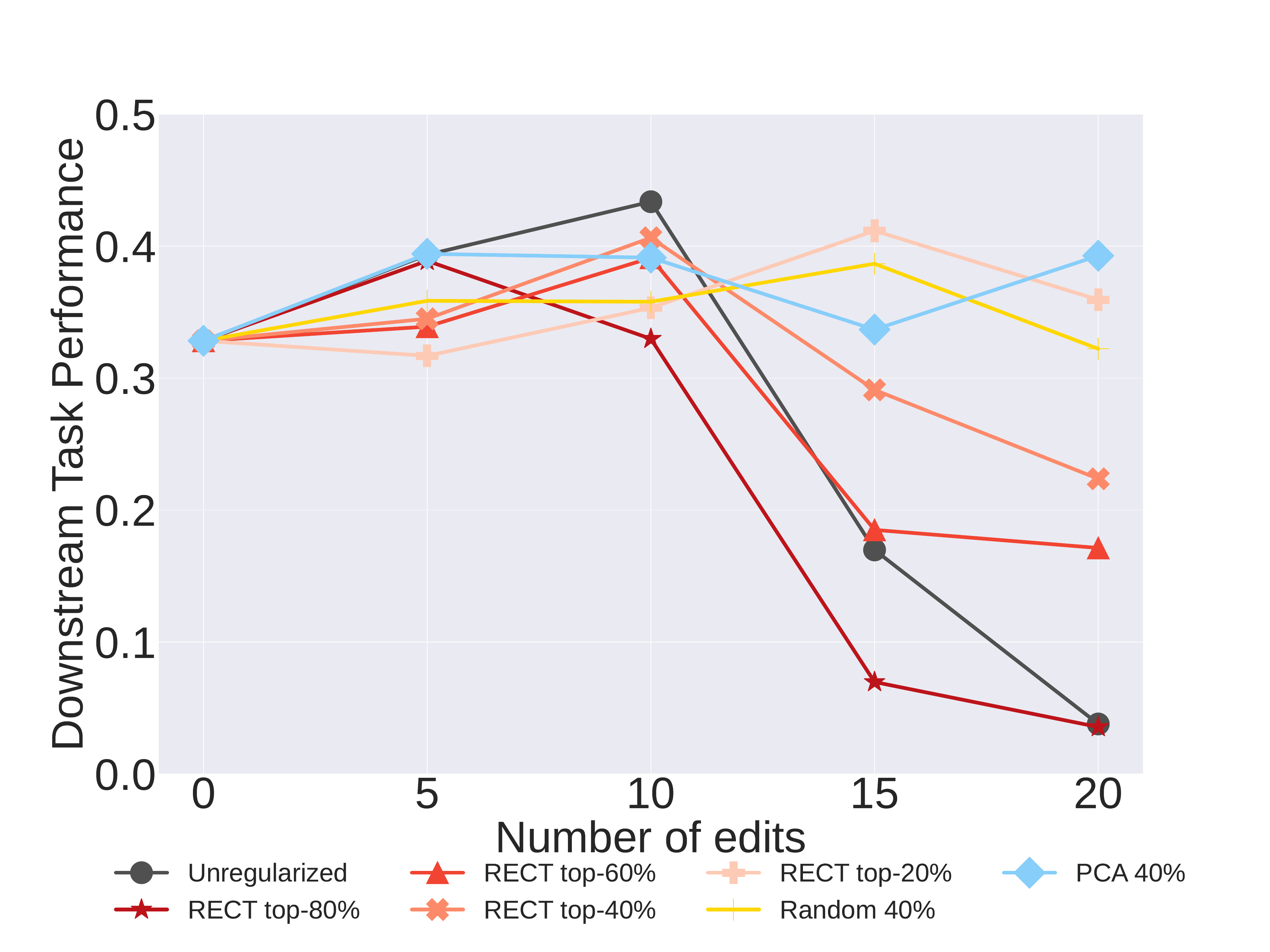}}
  \vspace{-4mm}
  \caption{Comparison of introducing various regularization methods for ROME and how the \emph{downstream task performance} change with respect to different top-\emph{k}\% for \MODELNAME{} based on GPT-2 XL.
  }
  \vspace{-4mm}
  \label{fig-rect-downstream-2}
\end{figure*}

%% file: acl_latex.bbl
\begin{thebibliography}{46}
\providecommand{\natexlab}[1]{#1}

\bibitem[{Brown et~al.(2020)Brown, Mann, Ryder, Subbiah, Kaplan et~al.}]{DBLP:conf/nips/BrownMRSKDNSSAA20}
Tom~B. Brown, Benjamin Mann, Nick Ryder, Melanie Subbiah, Jared Kaplan, et~al. 2020.
\newblock \href {https://proceedings.neurips.cc/paper/2020/hash/1457c0d6bfcb4967418bfb8ac142f64a-Abstract.html} {Language models are few-shot learners}.
\newblock In \emph{Advances in Neural Information Processing Systems 33: Annual Conference on Neural Information Processing Systems 2020, NeurIPS 2020, December 6-12, 2020, virtual}.

\bibitem[{Cao et~al.(2021)Cao, Aziz, and Titov}]{DBLP:conf/emnlp/CaoAT21}
Nicola~De Cao, Wilker Aziz, and Ivan Titov. 2021.
\newblock \href {https://doi.org/10.18653/v1/2021.emnlp-main.522} {Editing factual knowledge in language models}.
\newblock In \emph{Proceedings of the 2021 Conference on Empirical Methods in Natural Language Processing, {EMNLP} 2021, Virtual Event / Punta Cana, Dominican Republic, 7-11 November, 2021}, pages 6491--6506. Association for Computational Linguistics.

\bibitem[{Chen et~al.(2017)Chen, Fisch, Weston, and Bordes}]{DBLP:conf/acl/ChenFWB17}
Danqi Chen, Adam Fisch, Jason Weston, and Antoine Bordes. 2017.
\newblock \href {https://doi.org/10.18653/V1/P17-1171} {Reading wikipedia to answer open-domain questions}.
\newblock In \emph{Proceedings of the 55th Annual Meeting of the Association for Computational Linguistics, {ACL} 2017, Vancouver, Canada, July 30 - August 4, Volume 1: Long Papers}, pages 1870--1879. Association for Computational Linguistics.

\bibitem[{Cheng et~al.(2023)Cheng, Tian, Liu, Chen, Wang, Chen, and Zhang}]{DBLP:conf/emnlp/0008TL0WC023}
Siyuan Cheng, Bozhong Tian, Qingbin Liu, Xi~Chen, Yongheng Wang, Huajun Chen, and Ningyu Zhang. 2023.
\newblock \href {https://aclanthology.org/2023.emnlp-main.856} {Can we edit multimodal large language models?}
\newblock In \emph{Proceedings of the 2023 Conference on Empirical Methods in Natural Language Processing, {EMNLP} 2023, Singapore, December 6-10, 2023}, pages 13877--13888. Association for Computational Linguistics.

\bibitem[{Chowdhery et~al.(2023)Chowdhery, Narang, Devlin, Bosma, Mishra, Roberts, Barham, Chung, Sutton, Gehrmann et~al.}]{DBLP:journals/jmlr/ChowdheryNDBMRBCSGSSTMRBTSPRDHPBAI23}
Aakanksha Chowdhery, Sharan Narang, Jacob Devlin, Maarten Bosma, Gaurav Mishra, Adam Roberts, Paul Barham, Hyung~Won Chung, Charles Sutton, Sebastian Gehrmann, et~al. 2023.
\newblock \href {http://jmlr.org/papers/v24/22-1144.html} {Palm: Scaling language modeling with pathways}.
\newblock \emph{J. Mach. Learn. Res.}, 24:240:1--240:113.

\bibitem[{Clark et~al.(2019)Clark, Lee, Chang, Kwiatkowski, Collins, and Toutanova}]{DBLP:conf/naacl/ClarkLCK0T19}
Christopher Clark, Kenton Lee, Ming{-}Wei Chang, Tom Kwiatkowski, Michael Collins, and Kristina Toutanova. 2019.
\newblock \href {https://doi.org/10.18653/V1/N19-1300} {Boolq: Exploring the surprising difficulty of natural yes/no questions}.
\newblock In \emph{Proceedings of the 2019 Conference of the North American Chapter of the Association for Computational Linguistics: Human Language Technologies, {NAACL-HLT} 2019, Minneapolis, MN, USA, June 2-7, 2019, Volume 1 (Long and Short Papers)}, pages 2924--2936. Association for Computational Linguistics.

\bibitem[{Cobbe et~al.(2021)Cobbe, Kosaraju, Bavarian, Chen, Jun, Kaiser, Plappert, Tworek, Hilton, Nakano, Hesse, and Schulman}]{DBLP:journals/corr/abs-2110-14168}
Karl Cobbe, Vineet Kosaraju, Mohammad Bavarian, Mark Chen, Heewoo Jun, Lukasz Kaiser, Matthias Plappert, Jerry Tworek, Jacob Hilton, Reiichiro Nakano, Christopher Hesse, and John Schulman. 2021.
\newblock \href {https://arxiv.org/abs/2110.14168} {Training verifiers to solve math word problems}.
\newblock \emph{CoRR}, abs/2110.14168.

\bibitem[{Cohen et~al.(2024)Cohen, Biran, Yoran, Globerson, and Geva}]{DBLP:journals/tacl/CohenBYGG24}
Roi Cohen, Eden Biran, Ori Yoran, Amir Globerson, and Mor Geva. 2024.
\newblock \href {https://doi.org/10.1162/TACL\_A\_00644} {Evaluating the ripple effects of knowledge editing in language models}.
\newblock \emph{Trans. Assoc. Comput. Linguistics}, 12:283--298.

\bibitem[{Cui et~al.(2020)Cui, Wu, Liu, Zhang, and Zhou}]{DBLP:conf/acl/CuiWLZZ20}
Leyang Cui, Yu~Wu, Shujie Liu, Yue Zhang, and Ming Zhou. 2020.
\newblock \href {https://doi.org/10.18653/V1/2020.ACL-MAIN.130} {Mutual: {A} dataset for multi-turn dialogue reasoning}.
\newblock In \emph{Proceedings of the 58th Annual Meeting of the Association for Computational Linguistics, {ACL} 2020, Online, July 5-10, 2020}, pages 1406--1416. Association for Computational Linguistics.

\bibitem[{Dagan et~al.(2005)Dagan, Glickman, and Magnini}]{DBLP:conf/mlcw/DaganGM05}
Ido Dagan, Oren Glickman, and Bernardo Magnini. 2005.
\newblock \href {https://doi.org/10.1007/11736790\_9} {The {PASCAL} recognising textual entailment challenge}.
\newblock In \emph{Machine Learning Challenges, Evaluating Predictive Uncertainty, Visual Object Classification and Recognizing Textual Entailment, First {PASCAL} Machine Learning Challenges Workshop, {MLCW} 2005, Southampton, UK, April 11-13, 2005, Revised Selected Papers}, volume 3944 of \emph{Lecture Notes in Computer Science}, pages 177--190. Springer.

\bibitem[{Dai et~al.(2022)Dai, Dong, Hao, Sui, Chang, and Wei}]{DBLP:conf/acl/DaiDHSCW22}
Damai Dai, Li~Dong, Yaru Hao, Zhifang Sui, Baobao Chang, and Furu Wei. 2022.
\newblock \href {https://doi.org/10.18653/v1/2022.acl-long.581} {Knowledge neurons in pretrained transformers}.
\newblock In \emph{Proceedings of the 60th Annual Meeting of the Association for Computational Linguistics (Volume 1: Long Papers), {ACL} 2022, Dublin, Ireland, May 22-27, 2022}, pages 8493--8502. Association for Computational Linguistics.

\bibitem[{Gandikota et~al.(2023)Gandikota, Materzynska, Fiotto{-}Kaufman, and Bau}]{DBLP:conf/iccv/GandikotaMFB23}
Rohit Gandikota, Joanna Materzynska, Jaden Fiotto{-}Kaufman, and David Bau. 2023.
\newblock \href {https://doi.org/10.1109/ICCV51070.2023.00230} {Erasing concepts from diffusion models}.
\newblock In \emph{{IEEE/CVF} International Conference on Computer Vision, {ICCV} 2023, Paris, France, October 1-6, 2023}, pages 2426--2436. {IEEE}.

\bibitem[{Gliwa et~al.(2019)Gliwa, Mochol, Biesek, and Wawer}]{gliwa-etal-2019-samsum}
Bogdan Gliwa, Iwona Mochol, Maciej Biesek, and Aleksander Wawer. 2019.
\newblock \href {https://doi.org/10.18653/v1/D19-5409} {{SAMS}um corpus: A human-annotated dialogue dataset for abstractive summarization}.
\newblock In \emph{Proceedings of the 2nd Workshop on New Frontiers in Summarization}, pages 70--79, Hong Kong, China. Association for Computational Linguistics.

\bibitem[{Gupta et~al.(2024)Gupta, Rao, and Anumanchipalli}]{DBLP:conf/acl/GuptaRA24}
Akshat Gupta, Anurag Rao, and Gopala Anumanchipalli. 2024.
\newblock \href {https://aclanthology.org/2024.findings-acl.902} {Model editing at scale leads to gradual and catastrophic forgetting}.
\newblock In \emph{Findings of the Association for Computational Linguistics, {ACL} 2024, Bangkok, Thailand and virtual meeting, August 11-16, 2024}, pages 15202--15232. Association for Computational Linguistics.

\bibitem[{Hase et~al.(2023)Hase, Bansal, Kim, and Ghandeharioun}]{DBLP:conf/nips/HaseBKG23}
Peter Hase, Mohit Bansal, Been Kim, and Asma Ghandeharioun. 2023.
\newblock \href {http://papers.nips.cc/paper\_files/paper/2023/hash/3927bbdcf0e8d1fa8aa23c26f358a281-Abstract-Conference.html} {Does localization inform editing? surprising differences in causality-based localization vs. knowledge editing in language models}.
\newblock In \emph{Advances in Neural Information Processing Systems 36: Annual Conference on Neural Information Processing Systems 2023, NeurIPS 2023, New Orleans, LA, USA, December 10 - 16, 2023}.

\bibitem[{Huang et~al.(2023)Huang, Shen, Zhang, Zhou, Rong, and Xiong}]{DBLP:conf/iclr/HuangSZZR023}
Zeyu Huang, Yikang Shen, Xiaofeng Zhang, Jie Zhou, Wenge Rong, and Zhang Xiong. 2023.
\newblock \href {https://openreview.net/pdf?id=4oYUGeGBPm} {Transformer-patcher: One mistake worth one neuron}.
\newblock In \emph{The Eleventh International Conference on Learning Representations, {ICLR} 2023, Kigali, Rwanda, May 1-5, 2023}. OpenReview.net.

\bibitem[{Ji et~al.(2023)Ji, Lee, Frieske, Yu, Su, Xu, Ishii, Bang, Madotto, and Fung}]{DBLP:journals/csur/JiLFYSXIBMF23}
Ziwei Ji, Nayeon Lee, Rita Frieske, Tiezheng Yu, Dan Su, Yan Xu, Etsuko Ishii, Yejin Bang, Andrea Madotto, and Pascale Fung. 2023.
\newblock \href {https://doi.org/10.1145/3571730} {Survey of hallucination in natural language generation}.
\newblock \emph{{ACM} Comput. Surv.}, 55(12):248:1--248:38.

\bibitem[{Kwiatkowski et~al.(2019)Kwiatkowski, Palomaki, Redfield, Collins, Parikh, Alberti, Epstein, Polosukhin, Devlin, Lee, Toutanova, Jones, Kelcey, Chang, Dai, Uszkoreit, Le, and Petrov}]{DBLP:journals/tacl/KwiatkowskiPRCP19}
Tom Kwiatkowski, Jennimaria Palomaki, Olivia Redfield, Michael Collins, Ankur~P. Parikh, Chris Alberti, Danielle Epstein, Illia Polosukhin, Jacob Devlin, Kenton Lee, Kristina Toutanova, Llion Jones, Matthew Kelcey, Ming{-}Wei Chang, Andrew~M. Dai, Jakob Uszkoreit, Quoc Le, and Slav Petrov. 2019.
\newblock \href {https://doi.org/10.1162/TACL\_A\_00276} {Natural questions: a benchmark for question answering research}.
\newblock \emph{Trans. Assoc. Comput. Linguistics}, 7:452--466.

\bibitem[{Lee et~al.(2019)Lee, Chang, and Toutanova}]{DBLP:conf/acl/LeeCT19}
Kenton Lee, Ming{-}Wei Chang, and Kristina Toutanova. 2019.
\newblock \href {https://doi.org/10.18653/V1/P19-1612} {Latent retrieval for weakly supervised open domain question answering}.
\newblock In \emph{Proceedings of the 57th Conference of the Association for Computational Linguistics, {ACL} 2019, Florence, Italy, July 28- August 2, 2019, Volume 1: Long Papers}, pages 6086--6096. Association for Computational Linguistics.

\bibitem[{Levy et~al.(2017)Levy, Seo, Choi, and Zettlemoyer}]{DBLP:conf/conll/LevySCZ17}
Omer Levy, Minjoon Seo, Eunsol Choi, and Luke Zettlemoyer. 2017.
\newblock \href {https://doi.org/10.18653/v1/K17-1034} {Zero-shot relation extraction via reading comprehension}.
\newblock In \emph{Proceedings of the 21st Conference on Computational Natural Language Learning (CoNLL 2017), Vancouver, Canada, August 3-4, 2017}, pages 333--342. Association for Computational Linguistics.

\bibitem[{Li et~al.(2024)Li, Zhang, Yao, Wang, Chen, and Chen}]{DBLP:conf/iclr/Li0YW0C24}
Zhoubo Li, Ningyu Zhang, Yunzhi Yao, Mengru Wang, Xi~Chen, and Huajun Chen. 2024.
\newblock \href {https://openreview.net/forum?id=fNktD3ib16} {Unveiling the pitfalls of knowledge editing for large language models}.
\newblock In \emph{The Twelfth International Conference on Learning Representations, {ICLR} 2024, Vienna, Austria, May 7-11, 2024}. OpenReview.net.

\bibitem[{Lin(2004)}]{lin-2004-rouge}
Chin-Yew Lin. 2004.
\newblock \href {https://aclanthology.org/W04-1013} {{ROUGE}: A package for automatic evaluation of summaries}.
\newblock In \emph{Text Summarization Branches Out}, pages 74--81, Barcelona, Spain. Association for Computational Linguistics.

\bibitem[{Lowe et~al.(2015)Lowe, Pow, Serban, and Pineau}]{DBLP:conf/sigdial/LowePSP15}
Ryan Lowe, Nissan Pow, Iulian Serban, and Joelle Pineau. 2015.
\newblock \href {https://doi.org/10.18653/V1/W15-4640} {The ubuntu dialogue corpus: {A} large dataset for research in unstructured multi-turn dialogue systems}.
\newblock In \emph{Proceedings of the {SIGDIAL} 2015 Conference, The 16th Annual Meeting of the Special Interest Group on Discourse and Dialogue, 2-4 September 2015, Prague, Czech Republic}, pages 285--294. The Association for Computer Linguistics.

\bibitem[{Ma et~al.(2023)Ma, Gu, Ling, Liu, and Liu}]{DBLP:journals/corr/abs-2310-10322}
Jun{-}Yu Ma, Jia{-}Chen Gu, Zhen{-}Hua Ling, Quan Liu, and Cong Liu. 2023.
\newblock \href {https://doi.org/10.48550/ARXIV.2310.10322} {Untying the reversal curse via bidirectional language model editing}.
\newblock \emph{CoRR}, abs/2310.10322.

\bibitem[{Mao et~al.(2023)Mao, Zhang, Wang, Wang, Yao, Jiang, Xie, Huang, and Chen}]{DBLP:journals/corr/abs-2310-02168}
Shengyu Mao, Ningyu Zhang, Xiaohan Wang, Mengru Wang, Yunzhi Yao, Yong Jiang, Pengjun Xie, Fei Huang, and Huajun Chen. 2023.
\newblock \href {https://doi.org/10.48550/ARXIV.2310.02168} {Editing personality for llms}.
\newblock \emph{CoRR}, abs/2310.02168.

\bibitem[{Meng et~al.(2022)Meng, Bau, Andonian, and Belinkov}]{DBLP:conf/nips/MengBAB22}
Kevin Meng, David Bau, Alex Andonian, and Yonatan Belinkov. 2022.
\newblock \href {http://papers.nips.cc/paper\_files/paper/2022/hash/6f1d43d5a82a37e89b0665b33bf3a182-Abstract-Conference.html} {Locating and editing factual associations in {GPT}}.
\newblock In \emph{Advances in Neural Information Processing Systems 35: Annual Conference on Neural Information Processing Systems 2022, NeurIPS 2022, New Orleans, LA, USA, November 28 - December 9, 2022}.

\bibitem[{Meng et~al.(2023)Meng, Sharma, Andonian, Belinkov, and Bau}]{DBLP:conf/iclr/MengSABB23}
Kevin Meng, Arnab~Sen Sharma, Alex~J. Andonian, Yonatan Belinkov, and David Bau. 2023.
\newblock \href {https://openreview.net/pdf?id=MkbcAHIYgyS} {Mass-editing memory in a transformer}.
\newblock In \emph{The Eleventh International Conference on Learning Representations, {ICLR} 2023, Kigali, Rwanda, May 1-5, 2023}. OpenReview.net.

\bibitem[{Mitchell et~al.(2022{\natexlab{a}})Mitchell, Lin, Bosselut, Finn, and Manning}]{DBLP:conf/iclr/MitchellLBFM22}
Eric Mitchell, Charles Lin, Antoine Bosselut, Chelsea Finn, and Christopher~D. Manning. 2022{\natexlab{a}}.
\newblock \href {https://openreview.net/forum?id=0DcZxeWfOPt} {Fast model editing at scale}.
\newblock In \emph{The Tenth International Conference on Learning Representations, {ICLR} 2022, Virtual Event, April 25-29, 2022}. OpenReview.net.

\bibitem[{Mitchell et~al.(2022{\natexlab{b}})Mitchell, Lin, Bosselut, Manning, and Finn}]{DBLP:conf/icml/MitchellLBMF22}
Eric Mitchell, Charles Lin, Antoine Bosselut, Christopher~D. Manning, and Chelsea Finn. 2022{\natexlab{b}}.
\newblock \href {https://proceedings.mlr.press/v162/mitchell22a.html} {Memory-based model editing at scale}.
\newblock In \emph{International Conference on Machine Learning, {ICML} 2022, 17-23 July 2022, Baltimore, Maryland, {USA}}, volume 162 of \emph{Proceedings of Machine Learning Research}, pages 15817--15831. {PMLR}.

\bibitem[{Peng et~al.(2023)Peng, Galley, He, Cheng, Xie, Hu, Huang, Liden, Yu, Chen, and Gao}]{DBLP:journals/corr/abs-2302-12813}
Baolin Peng, Michel Galley, Pengcheng He, Hao Cheng, Yujia Xie, Yu~Hu, Qiuyuan Huang, Lars Liden, Zhou Yu, Weizhu Chen, and Jianfeng Gao. 2023.
\newblock \href {https://doi.org/10.48550/arXiv.2302.12813} {Check your facts and try again: Improving large language models with external knowledge and automated feedback}.
\newblock \emph{CoRR}, abs/2302.12813.

\bibitem[{Qin et~al.(2023)Qin, Zhang, Zhang, Chen, Yasunaga, and Yang}]{DBLP:conf/emnlp/QinZ0CYY23}
Chengwei Qin, Aston Zhang, Zhuosheng Zhang, Jiaao Chen, Michihiro Yasunaga, and Diyi Yang. 2023.
\newblock \href {https://aclanthology.org/2023.emnlp-main.85} {Is chatgpt a general-purpose natural language processing task solver?}
\newblock In \emph{Proceedings of the 2023 Conference on Empirical Methods in Natural Language Processing, {EMNLP} 2023, Singapore, December 6-10, 2023}, pages 1339--1384. Association for Computational Linguistics.

\bibitem[{Radford et~al.(2019)Radford, Wu, Child, Luan, Amodei, Sutskever et~al.}]{radford2019language}
Alec Radford, Jeffrey Wu, Rewon Child, David Luan, Dario Amodei, Ilya Sutskever, et~al. 2019.
\newblock Language models are unsupervised multitask learners.
\newblock \emph{OpenAI blog}, 1(8):9.

\bibitem[{Sang and Meulder(2003)}]{DBLP:conf/conll/SangM03}
Erik F. Tjong~Kim Sang and Fien~De Meulder. 2003.
\newblock \href {https://aclanthology.org/W03-0419/} {Introduction to the conll-2003 shared task: Language-independent named entity recognition}.
\newblock In \emph{Proceedings of the Seventh Conference on Natural Language Learning, CoNLL 2003, Held in cooperation with {HLT-NAACL} 2003, Edmonton, Canada, May 31 - June 1, 2003}, pages 142--147. {ACL}.

\bibitem[{Sinitsin et~al.(2020)Sinitsin, Plokhotnyuk, Pyrkin, Popov, and Babenko}]{DBLP:conf/iclr/SinitsinPPPB20}
Anton Sinitsin, Vsevolod Plokhotnyuk, Dmitry~V. Pyrkin, Sergei Popov, and Artem Babenko. 2020.
\newblock \href {https://openreview.net/forum?id=HJedXaEtvS} {Editable neural networks}.
\newblock In \emph{8th International Conference on Learning Representations, {ICLR} 2020, Addis Ababa, Ethiopia, April 26-30, 2020}. OpenReview.net.

\bibitem[{Socher et~al.(2013)Socher, Perelygin, Wu, Chuang, Manning, Ng, and Potts}]{DBLP:conf/emnlp/SocherPWCMNP13}
Richard Socher, Alex Perelygin, Jean Wu, Jason Chuang, Christopher~D. Manning, Andrew~Y. Ng, and Christopher Potts. 2013.
\newblock \href {https://aclanthology.org/D13-1170/} {Recursive deep models for semantic compositionality over a sentiment treebank}.
\newblock In \emph{Proceedings of the 2013 Conference on Empirical Methods in Natural Language Processing, {EMNLP} 2013, 18-21 October 2013, Grand Hyatt Seattle, Seattle, Washington, USA, {A} meeting of SIGDAT, a Special Interest Group of the {ACL}}, pages 1631--1642. {ACL}.

\bibitem[{Touvron et~al.(2023{\natexlab{a}})Touvron, Lavril, Izacard, Martinet, Lachaux, Lacroix, Rozi{\`{e}}re, Goyal, Hambro, Azhar, Rodriguez, Joulin, Grave, and Lample}]{DBLP:journals/corr/abs-2302-13971}
Hugo Touvron, Thibaut Lavril, Gautier Izacard, Xavier Martinet, Marie{-}Anne Lachaux, Timoth{\'{e}}e Lacroix, Baptiste Rozi{\`{e}}re, Naman Goyal, Eric Hambro, Faisal Azhar, Aur{\'{e}}lien Rodriguez, Armand Joulin, Edouard Grave, and Guillaume Lample. 2023{\natexlab{a}}.
\newblock \href {https://doi.org/10.48550/arXiv.2302.13971} {Llama: Open and efficient foundation language models}.
\newblock \emph{CoRR}, abs/2302.13971.

\bibitem[{Touvron et~al.(2023{\natexlab{b}})Touvron, Martin, Stone, Albert, Almahairi, Babaei, Bashlykov, Batra, Bhargava, Bhosale, Bikel, Blecher, Canton{-}Ferrer, Chen, Cucurull et~al.}]{DBLP:journals/corr/abs-2307-09288}
Hugo Touvron, Louis Martin, Kevin Stone, Peter Albert, Amjad Almahairi, Yasmine Babaei, Nikolay Bashlykov, Soumya Batra, Prajjwal Bhargava, Shruti Bhosale, Dan Bikel, Lukas Blecher, Cristian Canton{-}Ferrer, Moya Chen, Guillem Cucurull, et~al. 2023{\natexlab{b}}.
\newblock \href {https://doi.org/10.48550/arXiv.2307.09288} {Llama 2: Open foundation and fine-tuned chat models}.
\newblock \emph{CoRR}, abs/2307.09288.

\bibitem[{Wang et~al.(2024{\natexlab{a}})Wang, Zhang, Xie, Yao, Tian, Wang, Xi, Cheng, Liu, Zheng, and Chen}]{DBLP:journals/corr/abs-2308-07269}
Peng Wang, Ningyu Zhang, Xin Xie, Yunzhi Yao, Bozhong Tian, Mengru Wang, Zekun Xi, Siyuan Cheng, Kangwei Liu, Guozhou Zheng, and Huajun Chen. 2024{\natexlab{a}}.
\newblock \href {https://aclanthology.org/2024.acl-demos.9} {{E}asy{E}dit: An easy-to-use knowledge editing framework for large language models}.
\newblock In \emph{Proceedings of the 62nd Annual Meeting of the Association for Computational Linguistics (Volume 3: System Demonstrations)}, pages 82--93. Association for Computational Linguistics.

\bibitem[{Wang et~al.(2024{\natexlab{b}})Wang, Chen, Peng, and Chang}]{DBLP:journals/corr/abs-2401-10471}
Yiwei Wang, Muhao Chen, Nanyun Peng, and Kai{-}Wei Chang. 2024{\natexlab{b}}.
\newblock \href {https://doi.org/10.48550/ARXIV.2401.10471} {Deepedit: Knowledge editing as decoding with constraints}.
\newblock \emph{CoRR}, abs/2401.10471.

\bibitem[{Wei et~al.(2022)Wei, Bosma, Zhao, Guu, Yu, Lester, Du, Dai, and Le}]{DBLP:conf/iclr/WeiBZGYLDDL22}
Jason Wei, Maarten Bosma, Vincent~Y. Zhao, Kelvin Guu, Adams~Wei Yu, Brian Lester, Nan Du, Andrew~M. Dai, and Quoc~V. Le. 2022.
\newblock \href {https://openreview.net/forum?id=gEZrGCozdqR} {Finetuned language models are zero-shot learners}.
\newblock In \emph{The Tenth International Conference on Learning Representations, {ICLR} 2022, Virtual Event, April 25-29, 2022}. OpenReview.net.

\bibitem[{Wu et~al.(2023{\natexlab{a}})Wu, Peng, Chen, Su, and Sun}]{DBLP:journals/corr/abs-2308-09954}
Suhang Wu, Minlong Peng, Yue Chen, Jinsong Su, and Mingming Sun. 2023{\natexlab{a}}.
\newblock \href {https://doi.org/10.48550/ARXIV.2308.09954} {Eva-kellm: {A} new benchmark for evaluating knowledge editing of llms}.
\newblock \emph{CoRR}, abs/2308.09954.

\bibitem[{Wu et~al.(2023{\natexlab{b}})Wu, Li, Xu, Dong, Wu, Bian, and Xiong}]{DBLP:conf/emnlp/WuLXDW0X23}
Xinwei Wu, Junzhuo Li, Minghui Xu, Weilong Dong, Shuangzhi Wu, Chao Bian, and Deyi Xiong. 2023{\natexlab{b}}.
\newblock \href {https://aclanthology.org/2023.emnlp-main.174} {{DEPN:} detecting and editing privacy neurons in pretrained language models}.
\newblock In \emph{Proceedings of the 2023 Conference on Empirical Methods in Natural Language Processing, {EMNLP} 2023, Singapore, December 6-10, 2023}, pages 2875--2886. Association for Computational Linguistics.

\bibitem[{Yao et~al.(2023)Yao, Wang, Tian, Cheng, Li, Deng, Chen, and Zhang}]{DBLP:conf/emnlp/YaoWT0LDC023}
Yunzhi Yao, Peng Wang, Bozhong Tian, Siyuan Cheng, Zhoubo Li, Shumin Deng, Huajun Chen, and Ningyu Zhang. 2023.
\newblock \href {https://aclanthology.org/2023.emnlp-main.632} {Editing large language models: Problems, methods, and opportunities}.
\newblock In \emph{Proceedings of the 2023 Conference on Empirical Methods in Natural Language Processing, {EMNLP} 2023, Singapore, December 6-10, 2023}, pages 10222--10240. Association for Computational Linguistics.

\bibitem[{Zhang et~al.(2024)Zhang, Yao, Tian, Wang, Deng, Wang, Xi, Mao, Zhang, Ni, Cheng, Xu, Xu, Gu, Jiang, Xie, Huang, Liang, Zhang, Zhu, Zhou, and Chen}]{DBLP:journals/corr/abs-2401-01286}
Ningyu Zhang, Yunzhi Yao, Bozhong Tian, Peng Wang, Shumin Deng, Mengru Wang, Zekun Xi, Shengyu Mao, Jintian Zhang, Yuansheng Ni, Siyuan Cheng, Ziwen Xu, Xin Xu, Jia{-}Chen Gu, Yong Jiang, Pengjun Xie, Fei Huang, Lei Liang, Zhiqiang Zhang, Xiaowei Zhu, Jun Zhou, and Huajun Chen. 2024.
\newblock \href {https://doi.org/10.48550/ARXIV.2401.01286} {A comprehensive study of knowledge editing for large language models}.
\newblock \emph{CoRR}, abs/2401.01286.

\bibitem[{Zhang et~al.(2023)Zhang, Li, Cui, Cai, Liu, Fu, Huang, Zhao, Zhang, Chen, Wang, Luu, Bi, Shi, and Shi}]{DBLP:journals/corr/abs-2309-01219}
Yue Zhang, Yafu Li, Leyang Cui, Deng Cai, Lemao Liu, Tingchen Fu, Xinting Huang, Enbo Zhao, Yu~Zhang, Yulong Chen, Longyue Wang, Anh~Tuan Luu, Wei Bi, Freda Shi, and Shuming Shi. 2023.
\newblock \href {https://doi.org/10.48550/arXiv.2309.01219} {Siren's song in the {AI} ocean: {A} survey on hallucination in large language models}.
\newblock \emph{CoRR}, abs/2309.01219.

\bibitem[{Zhong et~al.(2023)Zhong, Wu, Manning, Potts, and Chen}]{DBLP:conf/emnlp/ZhongWMPC23}
Zexuan Zhong, Zhengxuan Wu, Christopher~D. Manning, Christopher Potts, and Danqi Chen. 2023.
\newblock \href {https://aclanthology.org/2023.emnlp-main.971} {Mquake: Assessing knowledge editing in language models via multi-hop questions}.
\newblock In \emph{Proceedings of the 2023 Conference on Empirical Methods in Natural Language Processing, {EMNLP} 2023, Singapore, December 6-10, 2023}, pages 15686--15702. Association for Computational Linguistics.

\end{thebibliography}
